\documentclass[sn-mathphys-ay]{sn-jnl}

\usepackage{graphicx}%
\usepackage{multirow}%
\usepackage{amsmath,amssymb,amsfonts}%
\usepackage{amsthm}%
\usepackage{mathrsfs}%
\usepackage[title]{appendix}%
\usepackage{xcolor}%
\usepackage{textcomp}%
\usepackage{manyfoot}%
\usepackage{booktabs}%
\usepackage{algorithm}%
\usepackage{algorithmicx}%
\usepackage{algpseudocode}%
\usepackage{listings}%
\usepackage{hyperref}

\usepackage{blindtext}
\usepackage{enumitem}
\setlist[itemize]{noitemsep, nolistsep}
\setlist[enumerate]{noitemsep, nolistsep}

\usepackage{amsmath}
\usepackage{amsfonts}
\usepackage{amssymb}
\usepackage{booktabs}
\usepackage{siunitx}
\usepackage{subcaption}

\DeclareUnicodeCharacter{2009}{\,}
\DeclareUnicodeCharacter{03C6}{$\phi$} 

\newcommand{\papertitle}{Empirical evaluation of normalizing flows in Markov Chain Monte Carlo\@\xspace}
\newcommand{\papertitleshort}{Empirical evaluation of normalizing flows in MCMC}
\newcommand{\diag}{\mathrm{diag}}

\newcommand{\eg}{e.g.,\@\xspace}
\newcommand{\ie}{i.e.,\@\xspace}
\newcommand{\bigo}[1]{\mathcal{O}( #1 )}
\newcommand{\pderiv}[2]{\ensuremath { \partial_{ #2 } #1 } }
\newcommand{\detjac}[2]{\ensuremath { \det \left( \partial_{ #2 } #1 \right) } }

\newcommand{\absdetjac}[2]{\ensuremath {\left| \detjac{ #1 }{ #2 } \right|}}
\newcommand{\logabsdetjac}[2]{\ensuremath{\log \absdetjac{ #1 }{ #2 }}}
\newcommand{\trace}[1]{\ensuremath \mathrm{Tr}\left( #1 \right)}

\newcommand{\ceil}[1]{\ensuremath \lceil #1 \rceil}
\newcommand{\logten}{\ensuremath \log_{10}}
\newcommand{\maxtwo}[2]{\ensuremath \max{\left( #1 , #2 \right)}}
\newcommand{\ndim}[1]{#1D\@\xspace}
\newcommand{\softmax}{\ensuremath \mathrm{softmax}}

\newcommand{\naftransformerdeep}{NN\textsubscript{deep}\@\xspace}
\newcommand{\naftransformerdense}{NN\textsubscript{dense}\@\xspace}
\newcommand{\naftransformerboth}{NN\textsubscript{both}\@\xspace}
\newcommand{\nafdeep}{NAF\textsubscript{deep}\@\xspace}
\newcommand{\nafdense}{NAF\textsubscript{dense}\@\xspace}
\newcommand{\nafboth}{NAF\textsubscript{both}\@\xspace}
\newcommand{\realnvp}{Real NVP\@\xspace}
\newcommand{\cnfeuler}{CNF\textsubscript{Euler}\@\xspace}
\newcommand{\cnfrk}{CNF\textsubscript{RK}\@\xspace}
\newcommand{\cnfrkreg}{CNF\textsubscript{RK(R)}\@\xspace}
\newcommand{\neutramcmc}{NeuTra MCMC\@\xspace}
\newcommand{\neutrahmc}{NeuTra HMC\@\xspace}
\newcommand{\neutramh}{NeuTra MH\@\xspace}

\newcommand{\tablefontsize}{\scriptsize}

\begin{document}

\title[\papertitleshort]{\papertitle}

\author*[1]{\fnm{David} \sur{Nabergoj}}\email{david.nabergoj@fri.uni-lj.si}
\author[1]{\fnm{Erik} \sur{\v{S}trumbelj}}\email{erik.strumbelj@fri.uni-lj.si}

\affil*[1]{\orgname{University of Ljubljana}, \orgdiv{Faculty of Computer and Information Science}, \orgaddress{\street{ Ve\v{c}na pot 113}, \postcode{1000} \city{Ljubljana}, \country{Slovenia}}}

\abstract{Recent advances in MCMC use normalizing flows to precondition target distributions and enable jumps to distant regions.
However, there is currently no systematic comparison of different normalizing flow architectures for MCMC.
As such, many works choose simple flow architectures that are readily available and do not consider other models.
Guidelines for choosing an appropriate architecture would reduce analysis time for practitioners and motivate researchers to take the recommended models as foundations to be improved.
We provide the first such guideline by extensively evaluating many normalizing flow architectures on various flow-based MCMC methods and target distributions.
When the target density gradient is available, we show that flow-based MCMC outperforms classic MCMC for suitable NF architecture choices with minor hyperparameter tuning.
When the gradient is unavailable, flow-based MCMC wins with off-the-shelf architectures.
We find contractive residual flows to be the best general-purpose models with relatively low sensitivity to hyperparameter choice.
We also provide various insights into normalizing flow behavior within MCMC when varying their hyperparameters, properties of target distributions, and the overall computational budget.}


\pacs[MSC Classification]{62-08\\ \\
{\footnotesize © The Author(s) 2025. This is the author’s accepted manuscript of an article accepted for publication in \textit{Machine Learning} (Springer). The final authenticated version will be available at \url{https://link.springer.com/} once published.}}

\maketitle

\section{Introduction}
In recent years, many works have used normalizing flows (NF) within Markov Chain Monte Carlo (MCMC) and Bayesian inference to accelerate distribution sampling in lattice field theory~\citep{del_debbio_efficient_2021, matthews_continual_2022, abbott_aspects_2023}, molecular dynamics~\citep{wu_stochastic_2020}, gravitational wave analyses~\citep{karamanis_accelerating_2022, williams_nested_2021}, general Bayesian model posteriors~\citep{hoffman_neutra-lizing_2019, grumitt_deterministic_2022, grumitt_flow_2024}, and other fields. 
NF-based MCMC (NFMC) approaches typically either transform a geometrically complex target distribution into a simple one that is more amenable to MCMC sampling~\citep{hoffman_neutra-lizing_2019} or replace some MCMC steps with independent sampling from an NF, which facilitates transitions to distant parts of the sampling space~\citep{samsonov_local-global_2024}.
These enhancements allowed researchers to obtain satisfactory analysis results compared to classic MCMC. Still, many papers use specific combinations of MCMC samplers and NF architectures for their target distribution.
A recent paper discusses some practical differences between NFMC samplers~\citep{grenioux_sampling_2023}.
In Section~\ref{sec:related-work}, we list several NFMC works and highlight key observations regarding NF properties within NFMC.
Many propose NFMC and NF combinations for their specific problem, but do not thoroughly evaluate existing methods or compare them to other NFMC and MCMC methods.
The lack of systematic evaluation and comparison hinders the adoption of NFMC for sampling and leaves some key questions unanswered:
\begin{enumerate}
    \item When does NFMC accelerate sampling over MCMC?
    \item When NFMC is the better choice, which combination of sampler and NF architecture is best?
\end{enumerate}
The first question is crucial to understanding when to invest time in configuring the more complex NFMC samplers.
The second question deals with choosing a suitable initial sampler-NF combination and saving time otherwise spent trying different, possibly unsuitable methods.
Answering these questions would substantially speed up research and analysis workflows that rely on NFMC while introducing the field to a broader audience that does not necessarily have the expertise required to select appropriate sampler-NF combinations.

We choose to answer these questions empirically for several reasons.
First, it is time-consuming to develop a reasonable theory for an NF model, a family of target distributions, or an NFMC sampler. Actionable findings can be obtained empirically and may also motivate different approaches to theoretical research. An empirical evaluation still compares the methods with useful metrics, while the comparison framework can also be used for future NF methods to quickly assess their practicality in MCMC. Lastly, we acknowledge that theoretical insights exist for training NFs in multimodal scenarios~\citep{cornish_relaxing_2020} and that they have been connected to NFMC~\citep{grenioux_sampling_2023}. Specifically, the difficulty of fitting an NF with a unimodal latent distribution increases with increasing distance between target modes. This may result in an ill-posed NF fit, where a pathological strand connects two modes in the NF distribution due to a topological mismatch between the target and the latent~\citep{cornish_relaxing_2020}.
However, the probability of sampling points on the strand may be very small compared to the modes. In applications with global NF proposals, it may only cause a small number of additional rejections while still allowing efficient jumps between modes.
Related phenomena may exist and be unaddressed by existing NFMC theory, further motivating an empirical approach.

We answer the first question by evaluating different NF architectures and NFMC samplers on various target distributions and comparing results to MCMC.
We answer the second question by narrowing our analysis to contexts where NFMC is superior to MCMC and then performing a detailed comparison of NFs within NFMC.
We state which NFMC-NF combination is the best choice in general and for families of similar distributions, as well as what NF properties are correlated with good sampling performance.

We investigate architectures in the family of autoregressive, residual, and continuous NFs, including popular models like \realnvp~\citep{dinh_density_2017}, MAF~\citep{papamakarios_masked_2017} and IAF~\citep{kingma_improved_2016}, neural spline flows~\citep[NSF]{durkan_neural_2019}, invertible ResNets~\citep[i-ResNet]{behrmann_invertible_2019}, continuous NFs~\citep{grathwohl_ffjord_2018, salman_deep_2018}, and others.
We first consider Metropolis-Hastings (MH) and Hamiltonian Monte Carlo (HMC) as classic MCMC baselines when the target log density gradients are unavailable and available, respectively.
We then augment each in two separate ways: by preconditioning the target according to \neutramcmc~\citep{hoffman_neutra-lizing_2019} and by replacing some MCMC steps with independent NF jumps according to Local-Global MCMC~\citep{gabrie_adaptive_2022, samsonov_local-global_2024}, which we term \textit{Jump MCMC}, abbreviating the MH version as Jump MH and HMC version as Jump HMC.
As a special case of Jump MH, we consider the independent Metropolis-Hastings sampler~\citep[IMH]{samsonov_local-global_2024, brofos_adaptation_2022}, which proposes the new state at each step as a sample from an NF.
We compare NFMC and NF methods on a custom benchmark consisting of four target distribution families: synthetic Gaussians, synthetic unimodal non-Gaussians, synthetic multimodal targets, and real-world distributions consisting of commonly analyzed Bayesian model posteriors.
In Section~\ref{sec:methods}, we describe in detail all considered NF architectures, samplers, target distributions, and our comparison methodology.

\section{Related work}\label{sec:related-work}
The most relevant work is by~\cite{grenioux_sampling_2023}, who analyze \neutramcmc, Jump MCMC, and NF-based importance sampling.
They find that NeuTra MCMC tends to fail for multimodal targets, but is competitive in the unimodal case, with both claims supported by empirical and theoretical evidence.
They also state that NFMC performance drops with increasing target dimensionality, with Jump MCMC being the most affected.
However, they draw their conclusions based on only one or two synthetic targets per experiment and only four autoregressive NF architectures, which limits generalization.
We build on their work by increasing the number of NF architectures with representatives from different NF families, as well as the suite of both synthetic and real-world target distributions.

A recent analysis finds that IMH outperforms the classic Metropolis-adjusted Langevin algorithm (MALA) on a 50D Gaussian process, a 2D multimodal target, and a multimodal lattice field theory target~\citep{brofos_adaptation_2022}.
Other works also find that IMH improves upon MCMC in field theory analyses~\citep{albergo_flow-based_2019, del_debbio_efficient_2021}, with 
\cite{abbott_aspects_2023} noting that scaling and scalability are affected by the choice of sampler hyperparameters, NF architecture, and NF hyperparameters.
However, numerical experiments in these works mainly consider field theory examples and use a few NF architectures at most.
Furthermore, they focus on multimodal distributions, whereas IMH could also be applied to other targets.
We note that~\cite{grenioux_sampling_2023} derive mixing time bounds for IMH with an isotropic Gaussian proposal, which suggest worse performance in sampling log-concave (a subset of unimodal) distributions than MCMC without NFs.
We complement these findings by empirically investigating IMH performance with NFs, which do not generally attain an isotropic Gaussian fit, thus also making our experiments on unimodal targets relevant.
\cite{hoffman_neutra-lizing_2019} state that sampling quality in \neutramcmc depends on the NF fit and that poor fits could lead to slow mixing in the tails, but only consider the IAF architecture without investigating sensitivity to hyperparameters or other architectures.
\cite{samsonov_local-global_2024} show that interleaving MALA with \realnvp jumps explores a 50D multimodal target and a 128D latent space of a generative adversarial network target better than either sampler alone.
Their approach outperforms the no-U-turn sampler~\citep[NUTS]{hoffman_no-u-turn_2011} on the funnel and Rosenbrock distributions in low-to-moderate dimensions but performs worse in high dimensions.

There are a number of findings for general NF-based sampling methods, including particle transport methods.
\cite{karamanis_accelerating_2022} state that their method is useful for computationally expensive targets, targets with highly correlated dimensions, and multimodal targets. However, they only consider the MAF architecture in their experiments.
\cite{grumitt_deterministic_2022} also apply their method to computationally expensive targets and find that small learning rates increase NF robustness for targets with complicated geometries.
\cite{grumitt_flow_2024} suggest using NF architectures with good inductive biases for common target geometries, \eg in hierarchical Bayesian models, but do not state specific architectures or empirically explore this idea.
\cite{wu_stochastic_2020} find NSF to be comparable or more expressive than the commonly used \realnvp for particle transport, which suggests that NFMC methods would benefit from more careful architecture choices. 
They also find that their NFMC method improves upon MCMC for the double well and alanine dipeptide multimodal test cases.
\cite{arbel_annealed_2021} and~\cite{matthews_continual_2022} build on this work, but both only consider the \realnvp architecture.
Similar to Jump MCMC,~\cite{cabezas_markovian_2024} recently proposed an adaptive MCMC algorithm using continuous NFs as global proposals. They train continuous NFs via a flow matching objective, which minimizes the distance between a continuous NF time-dependent vector field and a time-dependent vector field that corresponds to the target distribution.
While promising, the approach relies on a temperature annealing scheme that is directly tied to NF training.
This prevents a direct comparison with classic MCMC, making it difficult to isolate the quality of preconditioning or jumps, and to produce an accurate architecture comparison.

In summary, the field lacks clarity with respect to choosing an appropriate NF architecture and lacks an answer as to when NFMC is even a suitable alternative to MCMC.
Rough guidelines exist for the latter (low-to-moderate dimensional, multimodal, non-Gaussian, or computationally expensive targets), but they are not verified with a thorough empirical evaluation across many targets and NF architectures.

\section{Methods}\label{sec:methods}
In this section, we describe our notation, then list the analyzed MCMC methods and NF architectures.
We also describe the target distributions in the benchmark and our comparison methodology.

This section provides an overview of many methods whose conventional notation sometimes overlaps.
We keep the notation similar to the referenced papers when describing each method.
If two methods use the same symbol to represent different objects, we redefine the symbol in each description.
This helps avoid an overwhelming number of globally defined symbols.
All points, distributions, and probability density functions are $D$-dimensional unless noted otherwise.
All points and samples are in $\mathbb{R}^D$, and all probability density functions are defined on $\mathbb{R}^D$
We use $X$ and $Q$ to denote the target and NF distributions.
Similarly, we use $p_X$ and $q$ to denote the target and NF densities.
We denote the partial derivative of $f$ with respect to $x$ as $\pderiv{f(x)}{x}$.
If $x$ is a vector, then $\pderiv{f(x)}{x}$ is the Jacobian matrix of $f$.
Unless noted otherwise, if $f$ is a bijection mapping between a \textit{target space} and a \textit{latent space}, it is understood that the forward map $f$ maps a target point to a latent point, and the inverse map $f^{-1}$ maps a latent point to a target point.

\subsection{Samplers}
We first aim to show that adding NFs to MCMC can improve performance through independent jumps or preconditioning.
To eliminate some sources of variance within our experiments, we first limit ourselves to MCMC samplers with established and stable kernel tuning procedures.
We focus on \neutramcmc and Jump MCMC as they are direct extensions of MCMC without additional sampler components. 
They let us assess the preconditioning and jump performance of NFs in a controlled manner.
Methods like NUTS are very successful in practice. However, NUTS presently has no \neutramcmc or Jump MCMC extensions, and theoretically developing these is beyond the scope of our paper.
Furthermore, it has already been compared to NFMC in previous works~\citep{samsonov_local-global_2024, grumitt_deterministic_2022}.
Due to the large number of NF architectures and target distributions in the benchmark, evaluating many MCMC samplers would also lead to an even greater combinatorial explosion in the number of experiments.

We thus limit ourselves to MH as the gradient-free MCMC representative and HMC as the gradient-based one.
Besides being used in various practical analyses, their \neutramcmc and Jump MCMC extensions have also been theoretically analyzed~\citep{brofos_adaptation_2022, samsonov_local-global_2024, hoffman_neutra-lizing_2019}, which facilitates the understanding and discussion of experiments in our work.
Our results thus enrich the past assessments of these methods with a broader range of NF architectures.
We also provide many new empirical results for Jump MH and \neutramh, which are not commonly used but fit exactly into the frameworks defined by~\cite{samsonov_local-global_2024} and~\cite{hoffman_neutra-lizing_2019}.
Our results may encourage their use in practical gradient-free analyses.

All samplers we investigate are Metropolis methods.
Such methods start with an initial state for each chain, then iteratively propose new states and apply the Metropolis accept/reject rule to generate samples and compute distribution moments.
In the rest of this section, we list the investigated NF-based samplers for preconditioning and global exploration, as well as MH and HMC samplers that underlie these NFMC methods.
We further describe the investigated samplers in Appendix~\ref{app:sampler-definitions}.

\subsubsection{NeuTra preconditioning}
For preconditioning, we consider the NeuTra method~\citep{hoffman_neutra-lizing_2019}.
Instead of sampling from the target density $p_X$, \neutramcmc adjusts $p_X$ with a bijection $f$ and samples from the adjusted density.
The adjusted log density is defined as:
\begin{align}
    \log \widetilde{p}(z) = \log p_X(f^{-1}(z)) + \logabsdetjac{f^{-1}(z)}{z}. \label{eqn:neutra}
\end{align}
After MCMC, we transform the sampled points $z$ with $f^{-1}(z) = x$, yielding the samples from the target density $p_X$.
The bijection $f$ is associated with an NF, which is fit to $\log p_X$ with stochastic variational inference~\citep[SVI; see][for application with NFs]{rezende_variational_2015}.
The NF remains fixed after SVI, and we sample from the adjusted density.
Given a fixed computational budget, the challenge is training NFs well enough to outperform MCMC despite being unable to adjust $f$ during sampling.
\cite{schar_parallel_2024} recently explored a similar approach based on affine transformations, consisting of a shift operation and matrix multiplication using the general linear group.
While not directly tied to NF preconditioning, the work provides a framework for adaptive tuning of the preconditioner with ergodicity guarantees.

\subsubsection{Local MCMC and global NF proposals}
Another approach to using NFs within MCMC is replacing some MCMC proposals with independent sampling from an NF~\citep{gabrie_adaptive_2022}.
We term this Jump MCMC, as independent sampling with a global NF proposal amounts to \textit{jumping} to different parts of space to continue MCMC exploration.
The most straightforward approach is to replace the MCMC proposal with an independent sample from an NF $Q$ every $K$-th iteration.
Let $x_t$ denote the chain state at iteration $t$, divisible by $K$.
The proposed chain state and log acceptance probability are:
\begin{align}
    x^\prime_{t+1} &\sim Q, \label{eqn:jump-mcmc-x-prime} \\
    \log \alpha_{t} &= \log p_X(x^\prime_{t+1}) - \log p_X(x_t) + \log q(x_t) - \log q(x^\prime_{t+1}). \label{eqn:jump-mcmc-log-alpha}
\end{align}
When $K = 1$, we use an independent NF proposal at every iteration, which corresponds to the IMH sampler.
Whereas Jump MCMC combines global NF samples with local MCMC exploration, IMH uses solely NF-based jumps.
This can be beneficial if the NF approximates the target density well and the local MCMC fails to explore the space adequately.
An example where IMH is the better choice is finding mode weights of a multimodal distribution with strongly separated peaks.
Since our primary interest is finding how many particles fall into a particular mode, we do not care how well the modes are explored, so standard MCMC steps do not benefit us.
The challenge is again to find an NF that approximates the target density well enough to jump between regions of space more quickly than MCMC trajectories.

When investigating NFs in these aspects, we consider two scenarios that govern which underlying MCMC samplers are sensible choices.
First, we consider the case where the target log density gradient is unavailable. Such cases are found in, e.g., cosmology~\citep{karamanis_accelerating_2022} or simulations of physical systems~\citep{grumitt_flow_2024}, as many existing codes for dynamical system simulations are not differentiable despite recent efforts to change this~\citep{schoenholz_jax_2021}.
Second, we consider cases where the target log density has an available gradient.
This covers many different Bayesian model posteriors~\citep{agrawal_disentangling_2024, hoffman_neutra-lizing_2019} and recent work in lattice field theory~\citep{albergo_flow-based_2019}.

\subsubsection{Other NFMC samplers and related methods}
We acknowledge NFMC samplers such as Deterministic Langevin Monte Carlo~\citep[DLMC]{grumitt_deterministic_2022} and Transport elliptical slice sampling~\citep[TESS]{cabezas_transport_2023}. However, we choose to exclude them from our experiments.
We justify this as follows:
\begin{itemize}
    \item MH and HMC are already suitable for a fair comparison of NF architectures and have a simpler, extensively tested kernel-tuning procedure that is less prone to errors.
    \item Comparing MH/HMC performance to their NFMC analogs lets us determine when NFMC is better than MCMC while observing changes after adding preconditioning and NF jumps without otherwise affecting underlying MCMC dynamics. However, no such analog exists for DLMC. DLMC also mixes jump proposals with preconditioning, hindering our investigation of which approach is more efficient.
    \item We omit TESS as a gradient-free representative, as the underlying dynamics of elliptical slice sampling are more complex and challenging to analyze than simple MH dynamics.
\end{itemize}

Several approaches utilize NFs for preconditioning and jumps, but do not explicitly form an MCMC method.
We acknowledge nested sampling with NFs~\citep{williams_nested_2021}, which performs marginal likelihood estimation by combining independent NF sampling with rejection sampling.
We exclude the method from our experiments, as its primary application is marginal likelihood estimation instead of target sampling. Moreover, selecting its user-defined hyperparameters requires domain expertise and additional experimentation time to avoid high variability of results, which is beyond the scope of our analyses.
We also acknowledge methods with roots in sequential Monte Carlo with NFMC mutation kernels~\citep{karamanis_accelerating_2022, wu_stochastic_2020, arbel_annealed_2021, matthews_continual_2022} and NF-based importance sampling~\citep{midgley_flow_2023}. However, these rely on a more extensive set of hyperparameters, including careful target-dependent temperature scheduling, which would make an accurate comparison difficult.

\subsection{Normalizing flow architectures}\label{subsec:nf-architectures}
An NF is a distribution $Q$ defined as a transformation of a simple distribution $Z$ with a bijection $f$.
$f$ is typically parameterized by deep neural networks.
$Z$ is typically a multivariate standard normal distribution in referenced works.
We also use $Z = N(0, I)$ in this paper.
Sampling $x \sim Q$ is equivalent to sampling $z \sim Z$ and transforming the sample with $x = f^{-1}(z)$.
The log density $\log q$ of an NF $Q$ is computed as:
\begin{align}
    \log q(x) = \log p_Z(f(x)) + \logabsdetjac{f(x)}{x}, \label{eqn:nf-log-density}
\end{align}
where $p_Z$ is the density of $Z$.
This expression is similar to Equation~\ref{eqn:neutra}.
The difference is in our base distribution and intended use: Equation~\ref{eqn:neutra} transforms the target log density into a density that is easier to sample.
Equation~\ref{eqn:nf-log-density} transforms a simple distribution into a complex one that acts as a global proposal distribution or whose bijection $f$ preconditions a target density.
\cite{papamakarios_normalizing_2022} reviewed a large number of NF architectures, all defined using Equation~\ref{eqn:nf-log-density}, and identified three main NF families with different approaches to constructing $f$.
We describe these families in the following subsections and list the architectures we investigate.

\subsubsection{Autoregressive NFs}
The first family consists of autoregressive architectures, where $f$ is a composition of invertible deep bijections $f_i$ and the Jacobian of $f$ is triangular.
The functions $f_i$ are typically either coupling bijections or masked autoregressive (MA) bijections.

Coupling bijections receive an input $x$ and partition it into disjoint inputs $(x_A, x_B)$.
The output $f_i(x) = y$ is partitioned into $(y_A, y_B)$ on the same dimensions.
Part B stays constant ($y_B = x_A$) while $y_A$ is computed as $y_A = \tau(x_A; \phi(x_B))$, where $\tau$ is a \textit{transformer} -- a bijection, parameterized with $\phi(x_B)$.
The function $\phi$ is a \textit{conditioner}, which takes one of the vectors as input and predicts the parameters for $\tau$.
Coupling bijections are autoregressive as each dimension of $y$ is a function of the corresponding preceding dimensions in $x$.
$y_A$ trivially stays constant, while $y_B$ is transformed using the preceding $x_A$.
For the same reason, their Jacobian is also triangular and its determinant can be computed efficiently.
The function $\phi$ need not be bijective, so we make it a deep neural network, which makes $f_i$ expressive.
The composition $f$ thus accurately models complex distributions provided $\phi$ is sufficiently complex and the number of bijections $f_i$ is sufficiently large~\citep{, draxler_universality_2024, lee_universal_2021}.
We compose coupling bijections with permutations (also bijections) to avoid repeatedly using the same dimensions in parts A and B. This also means the dimension order can be arbitrary when analyzing the Jacobian.

MA bijections explicitly compute each output dimension as a function of \textit{all} preceding input dimensions. This also holds for inverse autoregressive (IA) bijections, which are the inverses of MA bijections.
This is in contrast to coupling bijections, which retain the autoregressive property by partitioning the input and thus cleverly ignoring specific dimensions.
MA bijections achieve this with a Masked Autoencoder for Distribution Estimation~\citep[MADE]{germain_made_2015} as the conditioner.
MADE maps a $D$-dimensional input to a $D$-dimensional output with an autoencoder whose weights are masked to retain the autoregressive property between layers and thus in the entire neural network.
All coupling bijection transformers are compatible with MA bijection transformers and vice versa.
The drawback of MA bijections is that their inverse pass requires $\bigo{D}$ operations.
This results in poor scaling with data dimensionality when we need to perform both the forward and inverse passes.

Each coupling and MA bijection consists of a transformer and a conditioner. 
For thoroughness, we investigate all combinations of the following:
\begin{itemize}
    \item For transformers, we consider shift~\citep[used in NICE]{dinh_nice_2015}, affine map~\citep[used in \realnvp]{dinh_density_2017}, linear rational spline~\citep[LRS]{dolatabadi_invertible_2020}, rational quadratic spline~\citep[RQS]{durkan_neural_2019}, and invertible neural networks~\citep[used in NAF]{huang_neural_2018}: \naftransformerdeep, \naftransformerdense, and \naftransformerboth, corresponding to neural networks that are (1) deep and thin, (2) shallow and dense, and (3) deep and dense.
    \item For conditioners, we use (1) a coupling conditioner that splits an input tensor in half according to its first dimension and uses a feed-forward neural network to predict transformer parameters and (2) a MADE conditioner.
\end{itemize}
We denote all coupling architectures with the ``C-'' prefix and all IA architectures with the ``IA-'' prefix.
We investigate IA architectures instead of MA architectures as the former have efficient inverses, which is necessary for \neutramcmc.
We do not investigate MA architectures, as they have efficient forward but inefficient inverse passes.
This property is not useful for any investigated NFMC sampler.
We provide further details on conditioner and transformer hyperparameters in Appendix~\ref{app:sec:nf-hyperparameters}.

\subsubsection{Residual NFs}
In residual architectures, $f$ is a composition of residual bijections $f_i$.
These map an input according to $f_i(x) = x + g_i(x)$, where $g$ outputs a residual value.
A sufficiently long composition ensures that $x$ can gradually be transformed into a desired data point using small residual values.
We place the investigated residual NFs into two categories: architectures based on the matrix determinant lemma and contractive residual architectures that incrementally transform data with contractive maps.

The former contain bijections $f_i$, designed to have the Jacobian determinant equal to $\det \left( A + VW^\top \right)$, where $A \in \mathbb{R}^{D \times D}$ is invertible and $V, W \in \mathbb{R}^{D \times M}$.
If computing $\det A$ and $A^{-1}$ is tractable and $M \ll D$, the determinant can be computed efficiently via the matrix determinant lemma:
\begin{align*}
    \detjac{f_i(x)}{x} = \det \left( A + VW^\top \right) = \det \left( I + W^\top A^{-1} V \right) \det A.
\end{align*}
We investigate three architectures following this lemma: planar flows~\citep{rezende_variational_2015}, Sylvester flows~\citep{berg_sylvester_2018}, and radial flows~\citep{tabak_family_2013, rezende_variational_2015}.
Let $g_i^{(p)}, g_i^{(s)}, g_i^{(r)}$ denote the residual functions for Planar, Sylvester, and radial flows, respectively.
Residual function definitions and the resulting Jacobian determinants for Planar, Sylvester, and radial flows are, respectively:
\begin{alignat*}{3}
    g_i^{(p)}(x) &= v\sigma(w^\top x + b), && \detjac{f_i^{(p)}(x)}{x} = 1 + \sigma^{\prime}(w^\top x+b) w^\top v, \\
    g_i^{(s)}(x) &= V\sigma(W^\top x + b), && \detjac{f_i^{(s)}(x)}{x} = \det \left( I + S(x)W^\top V \right), \\
    g_i^{(r)}(x) &= \frac{\beta(x-x_0)}{\alpha + r(x)}, && \detjac{f_i^{(r)}(x)}{x} = \left( 1 + \frac{\alpha\beta}{(\alpha + r(x))^2} \right) \left( 1 + \frac{\beta}{\alpha + r(x)} \right)^{D-1},
\end{alignat*}
where $x, x_0, v, w \in \mathbb{R}^D; \alpha, \beta, b \in \mathbb{R}; \alpha > 0; r(x) = ||x-x_0||_2$; $\sigma$ is a differentiable elementwise activation function; $s(x) = \sigma^\prime(w^\top x + b); S(x) = \diag \left(\sigma^\prime(W^\top x + b)\right)$.
In this paper, we set $\sigma$ to be the sigmoid function.
Note that the Sylvester determinant further simplifies if W and V are specified with an orthonormal set of vectors~\citep{berg_sylvester_2018}.

Contractive residual architectures are compositions of residual bijections, which use contractive maps as functions $g_i$. A map is contractive with respect to a distance function $\delta$ if there exists a constant $L < 1$ such that $\delta \left( f_i(x), f_i(y) \right) \leq L \delta(x, y)$ for any $x, y \in \mathbb{R}^d$.
By the Banach fixed point theorem~\citep{behrmann_invertible_2019}, any contractive map has one fixed point $x_* = f_i(x_*)$, which we obtain by starting with an arbitrary $x_1$ and repeatedly applying $x_{k+1} = f_i(x_k)$.
This lets us compute the inverse of $f_i$ as well.
The update $x_{k+1} = x^\prime - g_i(x_k)$ is guaranteed to converge to $x_* = f_i^{-1}(x^\prime)$ for any starting point $x_1$.
We can compute an unbiased estimate of the log determinant with the Hutchinson trace estimator~\citep{hutchinson_stochastic_1989} within a power series~\citep{behrmann_invertible_2019}:
\begin{align*}
    \logabsdetjac{f_i(z)}{z} = \sum_{i=1}^\infty \frac{(-1)^{k+1}}{k} \trace{\pderiv{g_i(z)}{z}} \approx \sum_{j = 1}^n \frac{(-1)^{k+1}}{k} w_j^\top \pderiv{g_i(z)}{z} w_j,
\end{align*}
where $ w_j \sim \mathcal{N}(0, I)$.
We alternatively compute the series with the Russian roulette estimator~\citep{chen_residual_2019}:
\begin{align*}
    \sum_{i=1}^\infty \frac{(-1)^{k+1}}{k} \trace{ \pderiv{g_i(z)}{z} } \approx \mathbb{E}_{n, w} \left[ \sum_{k=1}^n \frac{(-1)^{k+1}}{k} w_j^\top \left( \pderiv{g_i(z)}{z} \right)^k w_j / P(N \geq k) \right],
\end{align*}
where $N$ is a positive random variable and $n \sim N$. We use $N \sim \mathrm{Geom}(0.5)$ as in the original paper.
We note that the Hutchinson trace estimator is also valid for $w_j^{(k)} \sim_{\mathrm{iid}} \mathrm{Rademacher}$, however, an analysis by~\cite{chen_residual_2019} found the differences to be fairly small and even in favor of Gaussian random variables under a certain parametrization.
We investigate the difference between the two by evaluating both i-ResNet that uses the power series estimator and the residual flow~\citep[ResFlow]{chen_residual_2019} that uses the Roulette estimator.
Both construct $g_i$ as neural networks with spectral regularization to ensure $L < 1$, which makes the maps contractive.
We provide neural network parameterization details in Appendix~\ref{app:sec:nf-hyperparameters}.

\subsubsection{Continuous NFs}
Lastly, we consider continuous NF architectures.
Unlike autoregressive and residual architectures, which are compositions of a finite number of layers, continuous NFs transform points between latent and target spaces by simulating an ordinary differential equation (ODE).
This continuously maps a latent point $z = z_0 \sim p_Z$ from time $t_0$ to a target data point $x = z_1 = f^{-1}(z_0)$ at time $t_1$.
We compute the target point and log determinant as:
\begin{align}
	z_1 = z_0 + \int_{t_0}^{t_1} g_\phi (t, z_t), \;
	\logabsdetjac{f^{-1}(z_0)}{z_0} = -\int_{t_0}^{t_1} \trace{\pderiv{g_\phi (t, z_t)}{z_t}}, \label{eqn:continuous-nf}
\end{align}
where the neural network $g_\phi: \mathbb{R^d} \rightarrow \mathbb{R^d}$ determines the ODE: $g_\phi(t, z_t) = \pderiv{z_t}{t}$.
To compute the latent point, we subtract the integral in Equation~\ref{eqn:continuous-nf} (left) from both sides.
Similarly, we omit the minus sign in Equation~\ref{eqn:continuous-nf} (right) to obtain the log determinant of $f(z_1)$ with respect to $z_1$.
We compute the integrals using numerical solvers.
The integrals for point and log determinant computation can be computed jointly by combining the two ODEs in $\mathbb{R}^d$ into a single one in $\mathbb{R}^{2d}$.

We investigate three kinds of continuous NFs with different ways of solving the integral in Equation~\ref{eqn:continuous-nf} and different specifications of neural networks $g_\phi$:
\begin{itemize}
    \item \cnfeuler, which approximates the integral with the Euler-Maruyama solver in 150 steps~\citep{salman_deep_2018} and parameterizes $g_\phi$ as a time-independent feed-forward network.
    \item \cnfrk, which approximates the integral with the adaptive Runge-Kutta 4(5) solver \citep{grathwohl_ffjord_2018, finlay_how_2020} and parameterizes $g_\phi$ with a time-dependent neural network.
    \item \cnfrkreg, which is the same as \cnfrk, but regularizes $g_\phi$ by the squared norm of the Jacobian of its transformation~\citep{grathwohl_ffjord_2018}.
\end{itemize}
All three methods use the Hutchinson trace estimator to compute an unbiased estimate of the trace and avoid costly deterministic computations:
\begin{align*}
	\trace{\pderiv{f(x)}{x}} \approx \frac{1}{n} \sum_{i=1}^n w_i^\top \pderiv{f(x)}{x} w_i, w_i \sim \mathcal{N}(0, I),
\end{align*}
where the sum is computed efficiently using Jacobian-vector products.
\cnfeuler solves the ODE in a finite number of steps with the Euler-Maruyama method, which can create significant errors compared to more advanced ODE solvers but is very fast to evaluate.
Conversely, the adaptive Runge-Kutta 4(5) solver provides more accurate log density estimates and precise samples but can be expensive to evaluate.
We provide neural network parameterization details in Appendix~\ref{app:sec:nf-hyperparameters}.

\subsection{Benchmark target distributions}
We consider synthetic and real-world target distributions in our benchmark, which we describe in the following sections.
Our benchmark includes various commonly analyzed targets and partly overlaps with the recently described NF benchmark for SVI~\citep{agrawal_disentangling_2024}.

\subsubsection{Synthetic targets}\label{subsubsec:synthetic-targets}
Synthetic targets let us evaluate NFs in scenarios that mimic regions of real-world distributions.

If our target distribution is approximately Gaussian, it is beneficial to focus on NFMC and NF methods with good performance on actual synthetic Gaussians.
We include four such 100D distributions in our benchmark: standard Gaussian, diagonal Gaussian, full-rank Gaussian, and ill-conditioned full-rank Gaussian.
The mean of all targets is zero in each dimension.
Eigenvalues for diagonal and full-rank Gaussians are linearly spaced between 1 and 10. 
Reciprocals of eigenvalues for the ill-conditioned full-rank Gaussian are sampled from $\mathrm{Gamma}(0.5, 1)$, making the condition number of the covariance matrix high and causing sampling from this target to be difficult.
We provide details on full-rank Gaussians in Appendix~\ref{app:gaussian-targets}.

Hierarchical Bayesian models are common ways of modeling real-world phenomena and often use priors with spatially varying curvature~\citep{grumitt_deterministic_2022, grumitt_flow_2024}.
Sampling from such priors is complex, which leads to long MCMC runs when we have few data points for the likelihood.
We include the 100D funnel and 100D Rosenbrock distributions in our benchmark to facilitate choosing sampling methods in this scenario.
The funnel distribution and the Rosenbrock log density are respectively defined as:
\begin{align*}
    x_1 &\sim \mathcal{N}(0, 3); x_i | x_1 \sim N(0, \exp(x_1 / 2)) \\
    \log p_X(x) &= -\sum_{d=1}^{D/2} s (x_{2d-1}^2 - x_{2d})^2 + (x_{2d-1} - 1)^2 - C,
\end{align*}
where $D$ is even, $s > 0$ is a fixed scale parameter, and $C$ is the log of the normalization constant.

Many NFMC and NF methods were proposed to sample from multimodal distributions.
To evaluate NFMC methods and NFs in such cases, we include the following targets in our benchmark:
\begin{itemize}
    \item 100D Gaussian mixture with three components, equal weights.
    \item 100D Gaussian mixture with 20 components, random weights.
    \item 10D double well distribution with $2^{10}$ modes.
    \item 100D double well distribution with $2^{100}$ modes.
\end{itemize}
The double well density is defined as $\log p_X(x) = -\sum_{d=1}^D (x^2 - 4)^2 - C$, where $C$ is the log of the normalization constant.
We provide details on target definitions in Appendix~\ref{app:multimodal-targets}.
We view these distributions as increasingly complex due to their growing number of components and modes.
A successful sampling method will retain good performance regardless of the number of modes.

\subsubsection{Real-world Bayesian model posteriors}
We include diverse Bayesian model posterior distributions that describe real-world phenomena.
These are also commonly used for MCMC benchmarking~\citep{magnusson_posteriordb_2024}:
\begin{itemize}
    \item 10D Eight schools target, which models the effectiveness of coaching programs for standardized college admission tests based on scores from eight schools.
    \item 25D German credit and 51D sparse German credit targets, which model credit risk.
    \item Two 89D targets and one 175D target, which model the concentration of radon in Minnesotan households.
    \item 501D synthetic item response theory target, which models the process of students answering questions.
    \item 3003D stochastic volatility target, which models the evaluation of derivative securities, such as options.
\end{itemize}
We provide precise distribution definitions in Appendix~\ref{app:real-world-targets}.

\subsection{Evaluation methodology}
We compare samplers and NFs in estimating the second moment of the target with a given computational budget.
We use the squared bias of the second moment as the comparison metric, shortened as $b^2$.
It measures the difference between the true second moment and the second moment as estimated using MCMC samples.
We provide a detailed definition and discuss its relation to the bias-variance decomposition of mean squared error in Appendix~\ref{app:sec:comparison-metrics}.
Using $b^2$ relates our experiments to other works in NFMC, as it is a commonly used metric in the field~\citep[see \eg][]{hoffman_neutra-lizing_2019, grumitt_deterministic_2022}.

When comparing methods across different targets, $b^2$ cannot be naively compared due to different true second moments.
We thus opt for a rank standardization approach~\citep{urbano_new_2019}.
We rank different methods from minimum to maximum $b^2$ on each target.
We compute standardized ranks (SR) on each target, then observe a method's empirical average rank $\overline{r}$ and standard error of the mean $\hat{\sigma}$ as an uncertainty estimate.
This ensures that all methods are comparable across all targets.
We choose not to transform SR to the $[0, 1]$ interval as in ~\citep{urbano_new_2019} because it would result in skewed uncertainties that are difficult to interpret.
We provide a detailed definition of SR in Appendix~\ref{app:sec:comparison-metrics}.

Each NFMC experiment consists of a target distribution, NFMC sampler, NF architecture, and the corresponding set of NF hyperparameters.
When comparing NFs, we consider two main cases when analyzing each experiment:
\begin{itemize}
    \item We observe $b^2$ when using the default NF hyperparameter set for the used architecture.
    \item We observe $b^2$ when using the NF hyperparameter set that yields the smallest $b^2$ among all hyperparameter sets on that experiment.
\end{itemize}
We choose the default hyperparameters as follows: we run all experiments with six different hyperparameter sets.
We then count the number of times each hyperparameter set attains the smallest $b^2$ across all experiments.
The set with the highest count is marked as the default.
By analyzing the results of experiments with default hyperparameters, we obtain estimates of how architectures will behave in new experiments without any hyperparameter tuning.
By analyzing results that pertain to the smallest $b^2$, we obtain best-case performance estimates for different architectures.

\section{Results}\label{sec:results}
In this section, we show our main sampler and NF comparison results.
For every experiment, we warm up the sampler for 3 hours and sample for 8 hours to ensure each NF has enough time and data to attain a good fit.
In Appendix~\ref{app:additional-results}, we provide additional results regarding NF operation speed and differences between autoregressive NF components, as well as experiments with short NFMC runs.
We provide all experiment configuration details in Appendix~\ref{app:sec:experiment-configuration}.
We also repeat some analyses using kernelized Stein discrepancy~\citep{liu_kernelized_2016} as the comparison metric in Appendix~\ref{app:subsec:ksd}, which describes other properties of empirical MCMC sample distributions.
In Appendix~\ref{app:subsec:isir}, we also evaluate Jump MCMC with an iterated sampling importance resampling kernel for global proposals~\citep{samsonov_local-global_2024}, which uses multiple proposals that further improve Jump MCMC efficiency.

\subsection{MCMC vs NFMC}\label{subsec:mcmc-vs-nfmc}
We show that NFMC can outperform MCMC despite using NF models with many trainable parameters.
We provide a short summary of our findings at the end of this section.

\subsubsection{Sampler comparison across all targets and NFs}
\begin{figure}[ht]
    \begin{subfigure}[t]{0.55\textwidth}
        \centering
        \includegraphics[width=\textwidth, trim={0 0 230 0}, clip]{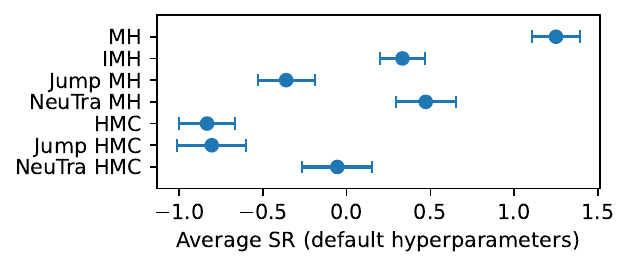}
        \caption{$\overline{r} \pm \hat{\sigma}$ across all targets and NFs for each sampler, using $b^2$ limited to experiments with default NF hyperparameters.}
        \label{subfig:nfmc-all-benchmarks-left}
    \end{subfigure}
    \hfill
    \begin{subfigure}[t]{0.421\textwidth}
        \centering
        \includegraphics[width=\textwidth, trim={300 0 0 0}, clip]{nfmc-all-benchmarks.pdf}
        \caption{$\overline{r} \pm \hat{\sigma}$ across all targets and NFs for each sampler, values estimated with minimum $b^2$ across all NF hyperparameter sets.}
        \label{subfig:nfmc-all-benchmarks-right}
    \end{subfigure}
    \caption{Numerical comparison of NFMC methods on the entire benchmark.}
\end{figure}

In Figure~\ref{subfig:nfmc-all-benchmarks-left}, we compare MCMC with NFMC in terms of SR based on experiments with default NF hyperparameters.

In the gradient-based setting, the differences between HMC and Jump HMC are negligible.
In the worst case, most jumps are either rejected, or states visited in this way do not contribute to moment estimates any more than HMC dynamics.
\neutrahmc performs more poorly than HMC and Jump HMC, suggesting that HMC exploration is generally hindered by NF preconditioning when NF hyperparameters are not tuned.

In the gradient-free setting, IMH, Jump MH, and \neutramh all outrank MH.
This suggests that preconditioning and jumps both improve MH dynamics.
Finding a balance between jumps and MH transitions is preferable to independent jumps or standard MH, as indicated by Jump MH outperforming both IMH and MH.
A pure global proposal strategy appears to outperform classic MCMC when the underlying transition kernel dynamics are not very expressive.
This can explain IMH ranking better than MH, but worse than HMC.
Indeed,~\cite{brofos_adaptation_2022} similarly found IMH to yield a better effective sample size and Kolmogorov-Smirnov statistic values than the less-expressive MALA sampler on synthetic and real-world examples.

Both Jump HMC and Jump MH outrank their NeuTra counterparts, with Jump MH also being better than \neutrahmc.
These results thus suggest that, on average, independent NF jumps enable better exploration than NF-based preconditioning.
This is consistent with a field theory analysis~\citep{grenioux_sampling_2023}, where a purely global NF sampler outperforms NeuTra MCMC with the local MALA and elliptical slice sampler kernels.
In some other experiments, the authors investigated NeuTra MCMC, Jump MCMC, and the combination thereof, and found examples where each performs the best.
Our experiments on this benchmark clarify that the greatest benefits stem from independent NF jumps.

In Figure~\ref{subfig:nfmc-all-benchmarks-right}, we compare samplers by their best attainable sampling performance.
Jump HMC outranks all other samplers.
Each jump-based sampler again outranks its MCMC and \neutramcmc counterparts, which agrees with our previous findings that jumps are preferable to preconditioning.
Even with a suitable NF architecture and hyperparameters, \neutrahmc only matches the performance of HMC and does not improve on it.
Despite being gradient-free samplers, IMH and Jump MH also potentially outrank or at least match HMC, which further demonstrates the benefits of independent jumps.

We note that SR compares the investigated samplers relative to each other and is not an absolute measure of sample quality.
For example, HMC achieves a worse SR in Figure~\ref{subfig:nfmc-all-benchmarks-right} than in Figure~\ref{subfig:nfmc-all-benchmarks-left} because of differently parametrized NFMC samplers.
Its $b^2$ is the same in both plots, but the NFMC $b^2$ values are different, which results in different SR values.

\subsubsection{Best-case analysis for specific target families}
We compare samplers when applied to different families of target distributions in Table~\ref{tab:nfmc-per-family}.

\begin{table}[t]

            \renewrobustcmd{\bfseries}{\fontseries{b}\selectfont}
            \renewrobustcmd{\boldmath}{}
            \sisetup{%
                table-align-uncertainty=true,
                detect-all,
                separate-uncertainty=true,
                mode=text,
                round-mode=uncertainty,
                round-precision=2,
                table-format = 2.2(2),
            }
            \tablefontsize
\begin{tabular}{l
S
S
S
S
S}
\toprule
{Sampler} & {Gaussian} & {Non-Gaussian} & {Multimodal} & {Real-world} & {All} \\
\midrule
{MH} & 1.5 & 1.5 & 1.5 & 1.38(0.12) & 1.44(0.06) \\
{IMH} & 0.38(0.47) & 0.00(0.50) & 0.25(0.25) & -0.69(0.09) & -0.17(0.17) \\
{Jump MH} & -0.50(0.29) & \bfseries -1.25(0.25) & -0.75(0.32) & 0.31(0.31) & -0.28(0.21) \\
{NeuTra MH} & 0.75(0.14) & 1.0 & 0.50(0.50) & 0.25(0.27) & 0.50(0.17) \\
{HMC} & -0.62(0.12) & 0.00(0.50) & -0.12(0.43) & 0.06(0.27) & -0.14(0.17) \\
{Jump HMC} & \bfseries -1.5 & \bfseries -1.25(0.25) & \bfseries -1.38(0.12) & \bfseries -0.94(0.39) & \bfseries -1.19(0.18) \\
{NeuTra HMC} & 0.00(0.20) & 0.0 & 0.00(0.20) & -0.38(0.26) & -0.17(0.13) \\
\bottomrule
\end{tabular}
\caption{$\overline{r} \pm \hat{\sigma}$ for all samplers and target families. Samplers with the best $\overline{r}$ are shown in bold for each target family. We estimate $\overline{r} \pm \hat{\sigma}$ with the minimum $b^2$ across all NFs for each target within a family. Entries without $\hat{\sigma}$ always attain the same $\overline{r}$. Ranks are computed separately for each target family.}
\label{tab:nfmc-per-family}
\end{table}

Jump HMC ranks best on each family and is tied with Jump MH on non-Gaussian unimodal targets.
In gradient-free sampling, Jump MH performs best on synthetic targets, while IMH is better on real-world targets.
This is consistent with a sparse logistic regression analysis by~\cite{grenioux_sampling_2023}, which found that adding global NF proposals to HMC outperforms classic HMC.
They similarly found IMH to be better on a multimodal target compared to a sampler that cannot transition between modes.
In our multimodal analyses, HMC ranks better than IMH on average, which could be due to the broad initialization of many chains and expressive HMC dynamics allowing chains to cross low target density barriers during the step size tuning phase.

\cite{hoffman_neutra-lizing_2019} show that \neutrahmc explores geometrically complex targets more efficiently than HMC. 
However, we find their ranks to be the same on non-Gaussian targets.
Due to the high uncertainty in HMC, it is plausible that some NFs allow \neutrahmc to rank better, though this is not the case in general.
Moreover, \neutrahmc is actually worse for Gaussian targets and does not improve results in general (see Figures~\ref{subfig:nfmc-all-benchmarks-left} and~\ref{subfig:nfmc-all-benchmarks-right}).
This finding is complementary to the multivariate Gaussian target analysis by~\cite{grenioux_sampling_2023}. The authors found that NeuTra MCMC performs worse when the Gaussian NF is close to the target, but outperforms Jump MCMC if the NF fit is poor.
This is consistent with the Gaussian targets in this benchmark being relatively simple to model, especially considering long NF training times and fairly large amounts of training data.

\subsubsection{Short summary}
Among the investigated samplers, we find no benefit to using gradient-based NFMC without tuning NF hyperparameters.
Jump MH is better than all other gradient-free samplers, even without hyperparameter tuning.
\neutrahmc ranks worse than Jump HMC and is on par with HMC when tuned properly.
\neutramh is also worse than Jump MH but better than MH when tuned properly.
Having found contexts where NFMC is better than MCMC, we now evaluate different NF architectures to see which one yields the best NFMC performance.

\subsection{NF architecture evaluation}
In this section, we empirically compare NF architectures in various contexts and provide model choice guidelines.
We give NF recommendations based on our findings at the end of this section.

\subsubsection{Jump performance on different target families}
We compare the performance of different NFs in Jump MCMC for different target families.
We focus on experiments where NFs use their default hyperparameters.
We show the results in Table~\ref{tab:nf-per-family}.

\begin{table}

            \renewrobustcmd{\bfseries}{\fontseries{b}\selectfont}
            \renewrobustcmd{\boldmath}{}
            \sisetup{%
                table-align-uncertainty=true,
                detect-all,
                separate-uncertainty=true,
                mode=text,
                round-mode=uncertainty,
                round-precision=2,
                table-format = 2.2(2),
            }
            \tablefontsize
\begin{tabular}{l
S
S
S
S
S}
\toprule
{NF} & {Gaussian} & {Non-Gaussian} & {Multimodal} & {Real-world} & {All} \\
\midrule
{NICE} & -0.43(0.31) & 1.45(0.14) & \bfseries -0.80(0.25) & 0.00(0.32) & -0.11(0.22) \\
{\realnvp} & \bfseries -0.80(0.25) & \bfseries -1.16(0.43) & 0.29(0.36) & 0.29(0.32) & -0.11(0.22) \\
{C-LR-NSF} & -0.14(0.72) & 0.72(0.58) & \bfseries -0.51(0.30) & -0.40(0.19) & -0.24(0.20) \\
{C-RQ-NSF} & \bfseries -0.87(0.19) & 1.01 & \bfseries -0.51(0.38) & 0.43(0.39) & 0.00(0.24) \\
{C-\nafdeep} & 0.58(0.19) & -0.14(0.87) & 0.72(0.35) & 0.80(0.25) & 0.63(0.17) \\
{C-\nafdense} & 0.94(0.46) & 0.29(0.14) & 0.58(0.65) & 1.09(0.15) & 0.85(0.18) \\
{C-\nafboth} & 1.30(0.20) & 0.29(0.43) & 0.72(0.78) & 0.76(0.29) & 0.82(0.22) \\
\midrule
{i-ResNet} & \bfseries -0.80(0.36) & -0.14(0.58) & -0.07(0.60) & \bfseries -0.98(0.25) & \bfseries -0.64(0.20) \\
{ResFlow} & 0.22(0.62) & -0.29(0.14) & 0.80(0.18) & -0.18(0.33) & 0.11(0.21) \\
\midrule
{\cnfeuler} & 0.14(0.35) & \bfseries -0.87(0.43) & -0.36(0.62) & \bfseries -0.54(0.31) & \bfseries -0.39(0.21) \\
{\cnfrk} & 0.29(0.65) & 0.00(1.59) & -0.07(0.48) & \bfseries -0.72(0.27) & -0.27(0.26) \\
{\cnfrkreg} & -0.43(0.31) & \bfseries -1.16(0.14) & \bfseries -0.80(0.27) & \bfseries -0.54(0.32) & \bfseries -0.64(0.17) \\
\bottomrule
\end{tabular}
\caption{$\overline{r} \pm \hat{\sigma}$ for all NFs and target families in IMH, Jump MH, and Jump HMC. NFs in the top 20th percentile are shown in bold for each target family. We estimate $\overline{r} \pm \hat{\sigma}$ with $b^2$ from runs with default hyperparameters. Entries without $\hat{\sigma}$ always attain the same $\overline{r}$. Ranks are computed separately for each target family.}
\label{tab:nf-per-family}
\end{table}

All four affine and spline-based autoregressive NFs achieve $\overline{r} < 0$ on Gaussian targets.
This is reasonable because NAF transformers have many more trainable parameters than are needed to model Gaussian distributions.
C-RQ-NSF achieves the lowest $\overline{r}$ of all NFs, making it the obvious default choice for approximately Gaussian targets.
However, the conceptually similar C-LR-NSF ranks worse with greater uncertainty.
We found that C-RQ-NSF yields $\overline{r} < -0.43$ on each Gaussian target, while results vary greatly for C-LR-NSF.
It yields $\overline{r} = -1.59$ on the ill-conditioned full-rank Gaussian target and $\overline{r} = 1.59$ on the full-rank Gaussian target.
The major difference is in the spline definition, which suggests that the LRS transformer is less stable than RQS, as both transformers use nearly identical spline parameterizations otherwise.
Among residual NFs, i-ResNet is tied for the second-best NF on Gaussians, while ResFlow ranks noticeably worse.
We further investigated the difference by narrowing the comparison to standard and diagonal Gaussian targets.
We found i-ResNet to always achieve lower $b^2$ on both targets across all jump MCMC samplers.
ResFlow wins in four of six full-rank Gaussian experiments.
This does not necessarily imply that one estimator is more suited for Gaussians than the other because i-ResNet parameterizes $g$ with two hidden layers of size 10 by default, and ResFlow parameterizes it with five hidden layers of size 100 by default.
However, i-ResNet is a better off-the-shelf architecture for diagonal Gaussians and a decent option for full-rank ones.
\cnfrkreg is the only continuous NF with $\overline{r} < 0$ on Gaussians. However, all continuous architectures exhibit a relatively high uncertainty.
Their unrestricted Jacobian allows very expressive transformations. 
However, this is unnecessary on Gaussians, which only require an appropriate scale, rotation, and shift of the base standard Gaussian distribution.

\cnfrkreg, \realnvp, and \cnfeuler rank best for non-Gaussian sampling.
\cnfrkreg is the preferred default choice due to having lower uncertainty than \realnvp.
Furthermore, all NFs except \cnfrkreg either attain a poor $\overline{r}$ or a high uncertainty, making most methods ineffective with default hyperparameters.
After checking specific experiment results, we found \realnvp to achieve the lowest $b^2$ across all three Jump MCMC samplers on the funnel target when using default hyperparameters.
This is sensible because an exact transformation from standard Gaussian to the funnel involves scaling dimensions by multiplying them with the exponential function of the first dimension.
The \realnvp architecture easily learns this transformation provided the first transformer is conditioned on the first dimension.
NICE ranks the worst of all autoregressive NFs as it only transforms with a shift and only uses two bijective layers in its default hyperparameter set.
Note that each coupling NF contained at least one coupling layer where the first dimension was passed to the conditioner.
All coupling NFs except NICE could thus find an exact solution for the funnel.
\cnfrkreg, NICE, and both NSF models achieve the best $\overline{r}$ on multimodal targets.
All other methods again exhibit poor $\overline{r}$ or high uncertainty.
As NF families, residual and continuous NFs outperform autoregressive NFs on real-world Bayesian model posterior targets.
i-ResNet ranks the best, followed by the three continuous architectures.

\begin{figure}
    \centering
    \includegraphics[width=0.75\linewidth]{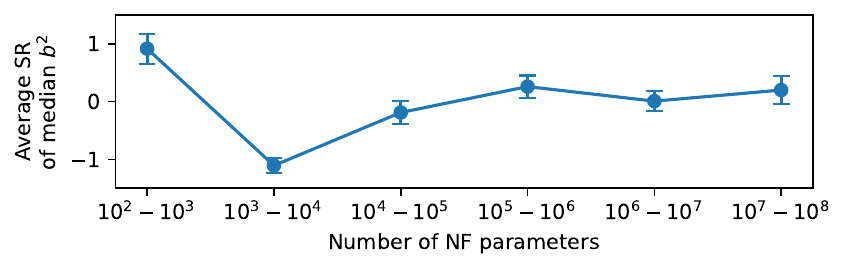}
    \caption{$\overline{r} \pm \hat{\sigma}$ for groups of NFs, defined using $\logten$ of the trainable NF parameter count. Groups were ranked for each target individually and then averaged across all targets.}
    \label{fig:nf-capacity}
\end{figure}

The best-performing architectures on the entire benchmark are \cnfrkreg, i-ResNet, and \cnfeuler.
All obtain a good $\overline{r}$ on each family except \cnfeuler on Gaussians.
\cnfrkreg achieves better overall $\overline{r}$ than \cnfrk, suggesting that regularization is beneficial for continuous NF models.
NAF models rank the worst, which could be due to having a very high parameter count.
A high parameter count increases the space of possible solutions, which can result in slower training.
Additionally, NAF models use neural network transformers, which are more difficult to optimize than affine or spline maps.
Residual NFs and continuous NFs have fewer restrictions on the form of their bijections than autoregressive NFs.
This could explain their better performance on real-world targets, whose complex geometry requires expressive bijections to be modeled.
Having observed how the bijection form can contribute to sampling quality, we also investigate the effect of the NF parameter count.
In Figure~\ref{fig:nf-capacity}, we plot the trainable parameter count against the SR of the NF.

We find a noticeable improvement in performance with $10^3$ to $10^4$ parameters where we observe a dip in SR.
The results verify our claim that one must choose hyperparameters that create a suitably expressive architecture.
Furthermore, they suggest that NF parameter counts in this range are a suitable default choice.
For SVI with \realnvp, \cite{agrawal_disentangling_2024} generally recommend using 10 or more layers for targets with complex geometry, as well as many hidden units for high-dimensional targets.
Our analysis shows that these recommendations do not directly translate into NFMC, as the best results are generally achieved by NFs with few parameters.

\begin{figure}
    \centering
    \includegraphics[width=\linewidth]{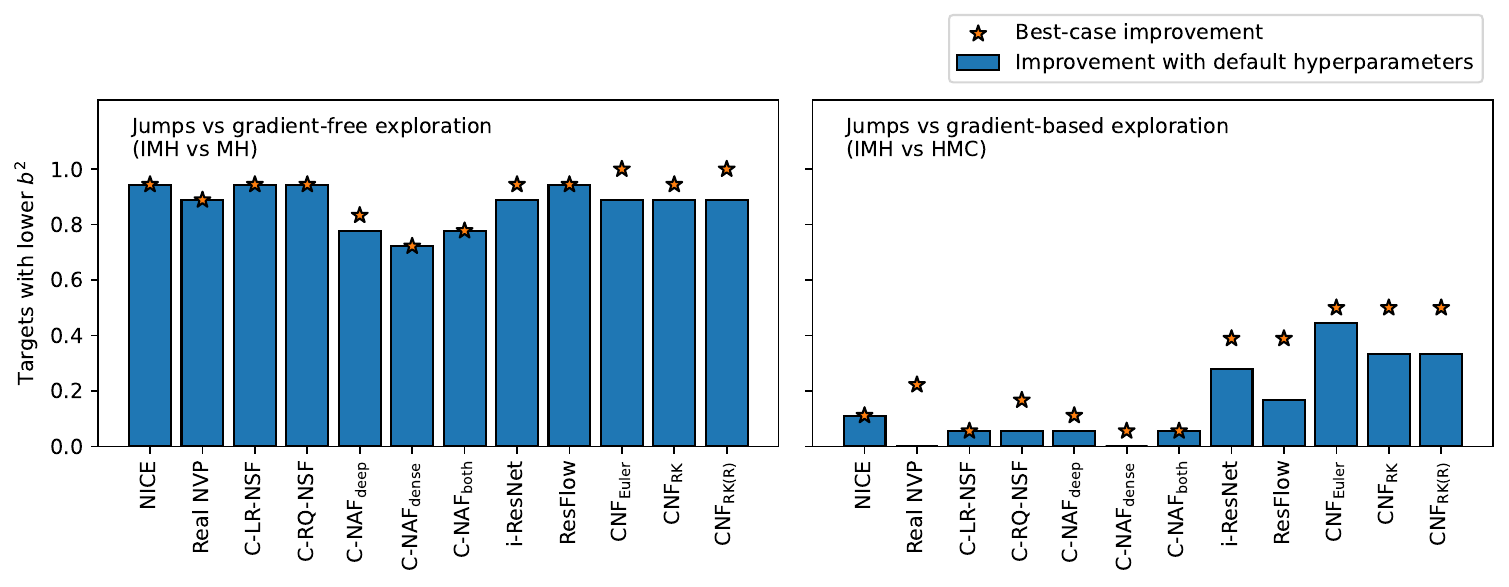}
    \caption{%
    Jump efficiency for NF architectures with and without target log density gradients, measured by the fraction of benchmark targets where IMH yields smaller $b^2$ than MH (left) and HMC (right). Bars denote IMH performance with default NF hyperparameters and stars denote IMH performance, corresponding to the minimum $b^2$ across all hyperparameter sets for an experiment.}
    \label{fig:jump-efficiency}
\end{figure}

Lastly, we provide a measure of jump efficiency for a chosen NF, independent of other architectures: we check the number of targets in our benchmark where IMH with a particular NF proposal achieves lower $b^2$ than MH and HMC.
The former lets us measure jump efficiency in the gradient-free setting, while the latter measures gradient-based performance.
For each target where IMH achieves lower $b^2$, we know that independent jumps with NFs are a more efficient exploration strategy than MCMC.
We show the results in Figure~\ref{fig:jump-efficiency}.

Regardless of which NF we choose, IMH is better than MH in at least 60\% of all cases.
The improvement is clear even when only considering the default NF hyperparameters.
Independent jumps prove to be a much weaker exploration strategy compared to gradient-based HMC exploration.
Continuous NFs beat HMC on 50\% of all targets in the best case and are slightly worse when using default hyperparameters.
We observe a similar pattern with residual NFs.
The superior performance of continuous and residual NFs is consistent with their better ranks in Table~\ref{tab:nf-per-family}.
While independent jumps are a suitable strategy for general gradient-free sampling, they are clearly inefficient for gradient-based targets if we have no prior knowledge.
These results are also consistent with Table~\ref{tab:nfmc-per-family}, where Jump MH outperforms MH and IMH, and Jump HMC outperforms HMC and IMH, further emphasizing the importance of mixing independent NF proposals with local MCMC exploration.

\subsubsection{Preconditioning quality on different target families}

We compare NF architectures by their performance in \neutramcmc.
Before interpreting the results, we note that \neutrahmc performed much worse than regular HMC.
Furthermore, while \neutramh improved upon regular MH, it was still vastly inferior to Jump MH.
This means that \neutramcmc may not be a very efficient preconditioning method, and the best-performing NFs could favor those whose transformation is close to the identity map.
We list the results in Table~\ref{tab:nf-per-family-neutra}.

\begin{table}[t]

            \renewrobustcmd{\bfseries}{\fontseries{b}\selectfont}
            \renewrobustcmd{\boldmath}{}
            \sisetup{%
                table-align-uncertainty=true,
                detect-all,
                separate-uncertainty=true,
                mode=text,
                round-mode=uncertainty,
                round-precision=2,
                table-format = 2.2(2),
            }
            \tablefontsize
\begin{tabular}{l
S
S
S
S
S}
\toprule
{NF} & {Gaussian} & {Non-Gaussian} & {Multimodal} & {Real-world} & {All} \\
\midrule
{NICE} & 0.21(0.42) & 1.07(0.25) & 0.62(0.38) & -0.31(0.27) & 0.17(0.20) \\
{\realnvp} & 0.17(0.32) & 1.57(0.08) & 0.83(0.29) & 0.52(0.37) & 0.62(0.21) \\
{C-LR-NSF} & -0.62(0.42) & 0.33(0.33) & -0.37(0.49) & 0.20(0.32) & -0.10(0.21) \\
{C-RQ-NSF} & \bfseries -0.87(0.33) & 0.33(0.17) & -0.04(0.25) & \bfseries -0.45(0.33) & -0.36(0.18) \\
{C-\nafdeep} & 1.07(0.08) & 0.50(0.17) & 1.40(0.11) & 0.62(0.41) & 0.88(0.20) \\
{C-\nafdense} & 0.33(0.45) & -0.91(0.58) & 1.11(0.23) & 0.56(0.30) & 0.47(0.22) \\
{C-\nafboth} & 0.04(0.32) & \bfseries -0.99(0.33) & 0.25(0.29) & 0.08(0.37) & -0.02(0.20) \\
\midrule
{IAF} & 0.54(0.44) & 1.16(0.33) & -0.04(0.60) & 0.39(0.27) & 0.41(0.21) \\
{IA-LR-NSF} & -0.25(0.55) & -0.25(0.08) & 0.37(0.62) & -0.07(0.41) & -0.03(0.25) \\
{IA-RQ-NSF} & -0.62(0.44) & -0.50(0.33) & \bfseries -0.87(0.31) & -0.30(0.28) & \bfseries -0.52(0.17) \\
{IA-\nafdeep} & \bfseries -0.83(0.24) & \bfseries -1.16(0.33) & -0.37(0.40) & \bfseries -0.45(0.38) & \bfseries -0.59(0.20) \\
{IA-\nafdense} & -0.54(0.27) & -0.91(0.41) & \bfseries -0.87(0.37) & -0.26(0.34) & \bfseries -0.53(0.19) \\
{IA-\nafboth} & 0.29(0.25) & -0.58(0.08) & 0.12(0.46) & -0.25(0.38) & -0.07(0.20) \\
\midrule
{i-ResNet} & -0.58(0.27) & -0.58(0.58) & \bfseries -0.70(0.18) & -0.39(0.21) & \bfseries -0.52(0.13) \\
{ResFlow} & \bfseries -0.95(0.39) & \bfseries -1.07(0.08) & -0.62(0.24) & \bfseries -0.43(0.22) & \bfseries -0.66(0.14) \\
\midrule
{Planar} & 0.29(0.55) & 0.25(0.08) & -0.04(0.21) & -0.11(0.31) & 0.04(0.18) \\
{Radial} & \bfseries -1.16(0.39) & \bfseries -1.32(0.33) & \bfseries -1.57(0.08) & \bfseries -1.07(0.21) & \bfseries -1.23(0.13) \\
{Sylvester} & -0.74(0.20) & -0.58(1.07) & 0.00(0.24) & \bfseries -0.67(0.34) & -0.47(0.19) \\
\midrule
{\cnfeuler} & 1.49 & 0.99 & -0.00(0.58) & 0.58(0.39) & 0.70(0.24) \\
{\cnfrk} & 1.32(0.23) & 1.49(0.17) & 0.08(0.86) & 0.93(0.38) & 0.89(0.27) \\
{\cnfrkreg} & 1.40(0.08) & 1.16 & 0.70(0.57) & 0.37(0.43) & 0.76(0.24) \\
\bottomrule
\end{tabular}
\caption{$\overline{r} \pm \hat{\sigma}$ for all NFs and target families in NeuTra MH and NeuTra HMC. NFs in the top 20th percentile are shown in bold for each target family. We estimate $\overline{r} \pm \hat{\sigma}$ with $b^2$ from runs with default hyperparameters. Entries without $\hat{\sigma}$ always attain the same $\overline{r}$. Ranks are computed separately for each target family.}
\label{tab:nf-per-family-neutra}
\end{table}

As groups, both IA and residual NFs perform the best across all targets (last column).
This is consistent with IA and radial flows having been designed to improve variational inference~\citep{tabak_family_2013, rezende_variational_2015, kingma_improved_2016}, which forms the crucial NF warm-up phase in \neutramcmc.
Interestingly, the radial flow performs better than IA within \neutramcmc despite the fact that IAF has been proposed to address the limitations of the former.
\neutramcmc can benefit from simpler preconditioning of the radial flow, which does not use neural networks, compared to more expressive bijections.
The form of the bijection could also play a role, as the simple planar NF ranks worse than radial and Sylvester flows.

The radial flow reaches the top 20\% of all NFs for every target family.
The next best are ResFlow and IA-\nafdeep, which rank among the best in all but multimodal sampling.
The radial flow appears somewhat more stable on real-world targets, while ResFlow is decisively stabler on synthetic targets. However, both attain the same uncertainty when evaluated across all targets.
We also notice good performance in some architectures that do not fit this pattern, namely coupling NFs.
While continuous NFs generally achieve better-than-average ranks on Jump MCMC, all of them perform poorly on \neutramcmc.
We again notice that \cnfrkreg achieves lower $\overline{r}$ than \cnfrk, which is consistent with our findings in Jump MCMC and further shows the benefits of using regularization in continuous NFs.

Our results also shed light on~\cite{hoffman_neutra-lizing_2019}, who precondition HMC with the IAF bijection, which improves moment estimates over HMC on the funnel target.
However, the second column of Table~\ref{tab:nf-per-family-neutra} shows that IAF is one of the worst-performing NFs for the funnel and Rosenbrock distributions.
The discrepancy could be due to differences in implementation, namely the number of chains: they use 16384 parallel chains on the GPU, whereas we use 100 on the CPU.
We cannot afford such a big number of GPU chains, as it would require a prohibitive amount of computational resources for a fair comparison with other, usually slower NFs.
\cite{hoffman_neutra-lizing_2019} also use a bigger SVI batch size compared to our single-sample unbiased loss estimator, which could contribute to good results.
We consider their IAF results complementary to ours, where we work with moderate computational resources, and they consider the case of higher resource and power consumption.
We also comment on batch size choices in Appendix~\ref{app:subsec:nf-training-details}.

\begin{figure}
    \centering
    \includegraphics[width=\linewidth]{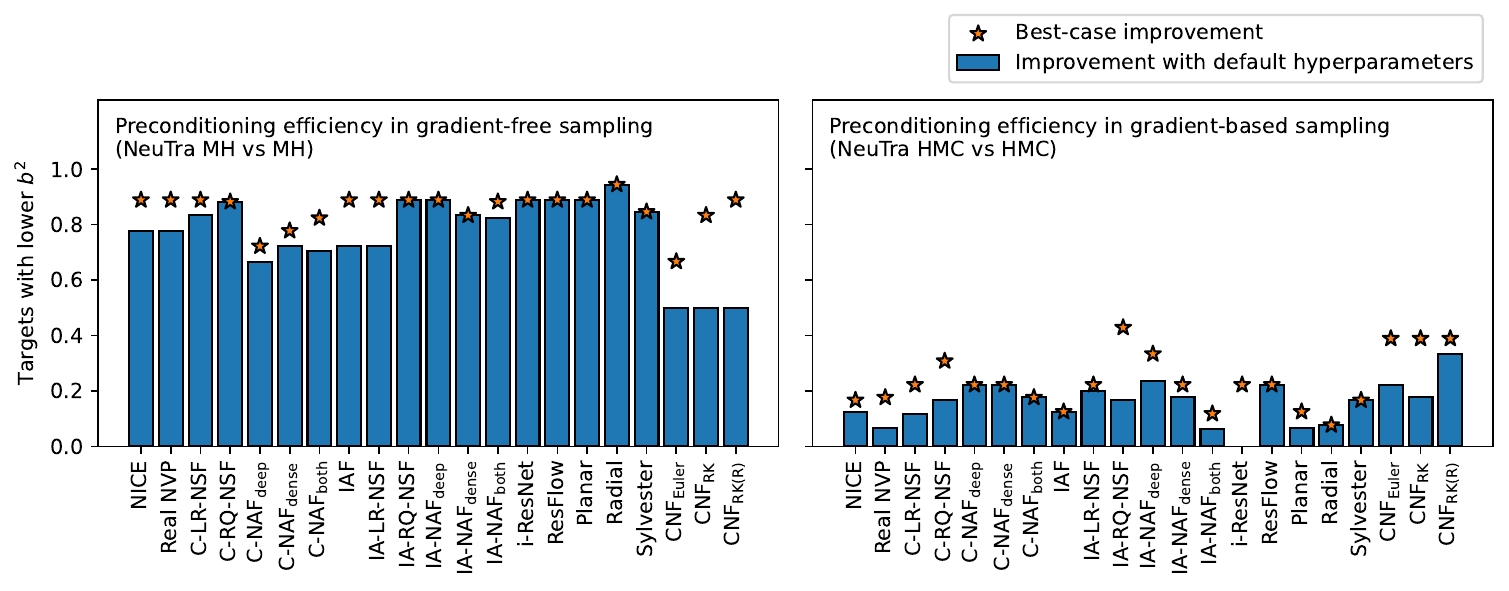}
    \caption{%
    Preconditioning efficiency for investigated NF architectures with and without target log density gradients, measured by the fraction of benchmark targets where \neutramcmc yields smaller $b^2$ than MCMC.  Bars denote \neutramcmc performance with default NF hyperparameters, and stars denote \neutramcmc performance, corresponding to the minimum $b^2$ across all hyperparameter sets for an experiment.}
    \label{fig:prec-efficiency}
\end{figure}

As before, we provide a measure of preconditioning efficiency for a particular NF, independent of other architectures.
For gradient-free sampling, we observe the percentage of targets where \neutramh achieves lower $b^2$ than MH.
We compare \neutrahmc to HMC in the same way for gradient-based sampling.
We show the results in Figure~\ref{fig:prec-efficiency}.

\neutramcmc shows the biggest improvement in the gradient-free case.
Continuous NFs have the worst gradient-free preconditioning performance with default hyperparameters, which is consistent with Table~\ref{tab:nf-per-family-neutra}.
However, they roughly match the performance of other NFs if tuned properly.
\neutramcmc is inefficient across the board in the gradient-based case, and improvements remain relatively small after hyperparameter tuning.
This matches with the best-case results in Table~\ref{tab:nfmc-per-family}, where \neutrahmc performs worse than Jump HMC and is similar to HMC.
\cnfrkreg performs better in \neutrahmc relative to other architectures than in \neutramh, despite attaining a poor average SR in Table~\ref{tab:nf-per-family-neutra}.
Conversely, the radial flow improves on fewer targets relative to other architectures in gradient-based sampling despite ranking the best in \neutramcmc.
This suggests that the choice of sampler plays a role in preconditioning efficiency.
However, gradient-based NF preconditioning is still an ineffective sampling strategy for our benchmark.

\subsubsection{Varying properties of synthetic targets}
In this section, we investigate the performance of NF architectures when different properties of synthetic distributions vary: dimensionality, curvature strength, number of multimodal components, and weights of multimodal components.
On the one hand, this lets us compare NFs on challenging target distributions with high dimensionality, high multimodality, and strong curvature.
On the other, we attain a measure of NF scalability in terms of these properties.

For scalability with dimensionality, we consider a diagonal Gaussian target with 2, 10, 100, 1000, and 10.000 dimensions.
For increasing curvature strength, we consider a \ndim{100} funnel target with first dimension scales equal to 0.01, 0.1, 1, 10, and 100.
For uneven multimodal weight tests, we consider the 20-component \ndim{100} Gaussian mixture with scales $\lambda \in \{0,1,2,3,4,5\}$ and weight of component $i$ equal to $w_i = \softmax(\lambda u)_i$, where $u_i \sim_{iid} N(0, 1)$. Scale $\lambda = 0$ results in equal weights.
For the increasing number of components, we consider both mixtures from Section~\ref{subsubsec:synthetic-targets} with 2, 8, 32, 128, and 512 components.
For each experiment, we first compute the median $b^2$ across NFMC methods and varied experiment values for each NF, then $\overline{r}$ and $\hat{\sigma}$ according to these medians.
We first report the results for Jump MCMC experiments in Table~\ref{tab:target-variations-jump-mcmc}.

\begin{table}

            \renewrobustcmd{\bfseries}{\fontseries{b}\selectfont}
            \renewrobustcmd{\boldmath}{}
            \sisetup{%
                table-align-uncertainty=true,
                detect-all,
                separate-uncertainty=true,
                mode=text,
                round-mode=uncertainty,
                round-precision=2,
                table-format = 2.2(2),
            }
            \tablefontsize
\begin{tabular}{l
S
S
S
S
S}
\toprule
{NF} & {Dimensionality} & {Curvature} & {Mode weight} & {Components} & {All} \\
\midrule
{NICE} & \bfseries -0.68(0.24) & \bfseries -0.48(0.27) & -0.26(0.19) & \bfseries -0.34(0.23) & \bfseries -0.43(0.11) \\
{\realnvp} & -0.13(0.24) & \bfseries -0.41(0.28) & -0.06(0.17) & -0.18(0.23) & -0.18(0.11) \\
{C-LR-NSF} & -0.27(0.17) & \bfseries -1.01(0.14) & \bfseries -0.45(0.27) & \bfseries -0.34(0.19) & \bfseries -0.49(0.11) \\
{C-RQ-NSF} & \bfseries -0.68(0.21) & -0.39(0.17) & -0.31(0.18) & \bfseries -0.49(0.21) & \bfseries -0.46(0.10) \\
{C-\nafdeep} & 0.55(0.23) & 0.29(0.31) & 0.50(0.21) & 0.43(0.26) & 0.45(0.12) \\
{C-\nafdense} & 0.64(0.25) & 0.68(0.25) & 0.43(0.16) & 0.40(0.19) & 0.52(0.10) \\
{C-\nafboth} & 0.70(0.26) & 0.80(0.32) & 1.24(0.13) & 0.74(0.28) & 0.90(0.12) \\
\midrule
{i-ResNet} & -0.21(0.19) & 0.10(0.15) & -0.06(0.18) & 0.20(0.27) & -0.00(0.10) \\
{ResFlow} & \bfseries -0.37(0.21) & 0.19(0.23) & -0.18(0.18) & 0.07(0.25) & -0.09(0.11) \\
\midrule
{\cnfeuler} & 0.30(0.28) & 0.36(0.27) & \bfseries -0.39(0.32) & -0.09(0.27) & 0.01(0.15) \\
{\cnfrk} & 0.33(0.29) & -0.00(0.26) & -0.03(0.23) & -0.03(0.25) & 0.07(0.13) \\
{\cnfrkreg} & 0.02(0.29) & -0.12(0.38) & \bfseries -0.43(0.27) & \bfseries -0.38(0.32) & -0.24(0.15) \\
\bottomrule
\end{tabular}
\caption{$\overline{r} \pm \hat{\sigma}$ for NFs in Jump MCMC given NF scalability scores when varying target properties: dimensionality, curvature strength, variance of mode weights, number of modes. NFs with $\overline{r}$ in the 20th percentile are shown in bold. Ranks computed separately for each target property.}
\label{tab:target-variations-jump-mcmc}
\end{table}

Aside from NAF, we find that autoregressive architectures generally outperform both residual and continuous NFs, always achieving $\overline{r} < 0$.
NICE and the two NSF architectures rank the best overall, followed by \cnfrkreg.
This puts into perspective the previous results from Table~\ref{tab:nf-per-family}: continuous NFs performed much better on the basic benchmark, but they rank worse when increasing the complexity of synthetic targets.

We also report experiment results for \neutramcmc in Table~\ref{tab:target-variations-neutra}.
C-LR-NSF handles complex synthetic targets the best, followed by inverse autoregressive NFs and contractive residual NFs.
C-LR-NSF, IA-\nafdeep, i-ResNet, and ResFlow always attain $\overline{r} < 0$, 
IA-NAF models perform better than their coupling counterparts despite both having a large number of transformer parameters.
This suggests that MADE conditioners are better suited for NAF transformers in \neutramcmc, which is also consistent with Jump MCMC results from Table~\ref{tab:nf-per-family-neutra}.
Despite its good performance on the regular benchmark, the radial flow is the worst overall for these synthetic target variations.
The poor ranks of continuous NF models are consistent with their poor \neutramcmc results on the regular benchmark.

\begin{table}[t]

            \renewrobustcmd{\bfseries}{\fontseries{b}\selectfont}
            \renewrobustcmd{\boldmath}{}
            \sisetup{%
                table-align-uncertainty=true,
                detect-all,
                separate-uncertainty=true,
                mode=text,
                round-mode=uncertainty,
                round-precision=2,
                table-format = 2.2(2),
            }
            \tablefontsize
\begin{tabular}{l
S
S
S
S
S}
\toprule
{NF} & {Dimensionality} & {Curvature} & {Mode weight} & {Components} & {All} \\
\midrule
{NICE} & 0.01(0.14) & -0.38(0.33) & \bfseries -0.93(0.15) & -0.42(0.31) & -0.46(0.13) \\
{\realnvp} & 0.05(0.22) & \bfseries -0.85(0.26) & -0.62(0.21) & 0.10(0.22) & -0.32(0.13) \\
{C-LR-NSF} & \bfseries -0.60(0.14) & -0.44(0.38) & \bfseries -1.16(0.10) & -0.47(0.30) & \bfseries -0.70(0.12) \\
{C-RQ-NSF} & 0.29(0.17) & \bfseries -0.74(0.53) & \bfseries -1.27(0.11) & -0.39(0.30) & \bfseries -0.53(0.15) \\
{C-\nafdeep} & 0.38(0.43) & 0.87(0.25) & 1.12(0.07) & 0.50(0.35) & 0.73(0.15) \\
{C-\nafdense} & 0.33(0.51) & 0.60(0.27) & 0.59(0.08) & 0.19(0.31) & 0.43(0.14) \\
{C-\nafboth} & 0.47(0.50) & 0.47(0.33) & 0.51(0.07) & 0.24(0.26) & 0.42(0.13) \\
\midrule
{IAF} & 0.50(0.25) & 0.03(0.30) & \bfseries -1.11(0.10) & -0.44(0.31) & -0.31(0.15) \\
{IA-LR-NSF} & -0.06(0.19) & 0.10(0.14) & -0.65(0.13) & \bfseries -0.88(0.24) & -0.41(0.11) \\
{IA-RQ-NSF} & -0.06(0.28) & 0.00(0.24) & \bfseries -0.93(0.11) & \bfseries -0.67(0.32) & \bfseries -0.51(0.14) \\
{IA-\nafdeep} & -0.16(0.20) & -0.05(0.44) & -0.83(0.18) & \bfseries -0.84(0.20) & \bfseries -0.51(0.13) \\
{IA-\nafdense} & \bfseries -0.53(0.34) & -0.08(0.26) & 0.34(0.22) & 0.48(0.18) & 0.09(0.14) \\
{IA-\nafboth} & \bfseries -0.51(0.40) & -0.05(0.25) & 0.80(0.20) & 0.53(0.20) & 0.27(0.15) \\
\midrule
{i-ResNet} & -0.30(0.40) & \bfseries -0.83(0.41) & -0.33(0.12) & \bfseries -0.59(0.22) & -0.47(0.16) \\
{ResFlow} & -0.38(0.23) & \bfseries -1.32(0.28) & -0.33(0.15) & -0.03(0.26) & -0.44(0.14) \\
\midrule
{Planar} & 0.43(0.19) & -0.64(0.19) & -0.02(0.14) & -0.19(0.22) & -0.07(0.10) \\
{Radial} & -0.11(0.51) & 0.89(0.46) & 1.65 & 1.65 & 1.12(0.18) \\
{Sylvester} & -0.40(0.15) & -0.31(0.44) & -0.16(0.08) & -0.32(0.23) & -0.29(0.11) \\
\midrule
{\cnfeuler} & 0.28(0.45) & 0.65(0.33) & 1.19(0.09) & 0.27(0.39) & 0.62(0.17) \\
{\cnfrk} & 0.98(0.30) & 0.70(0.28) & 0.90(0.24) & 0.72(0.35) & 0.83(0.14) \\
{\cnfrkreg} & \bfseries -0.82(0.35) & -0.07(0.30) & 0.92(0.05) & 0.24(0.21) & 0.12(0.15) \\
\bottomrule
\end{tabular}
\caption{$\overline{r} \pm \hat{\sigma}$ for NFs in NeuTra MCMC given NF scalability scores when varying target properties: dimensionality, curvature strength, variance of mode weights, number of modes. NFs with $\overline{r}$ in the 20th percentile are shown in bold. Ranks computed separately for each target property.}
\label{tab:target-variations-neutra}
\end{table}

\subsubsection{Measuring NF stability via hyperparameter sensitivity}\label{subsubsec:stability}
\begin{figure}
    \centering
    \includegraphics[width=1.0\linewidth]{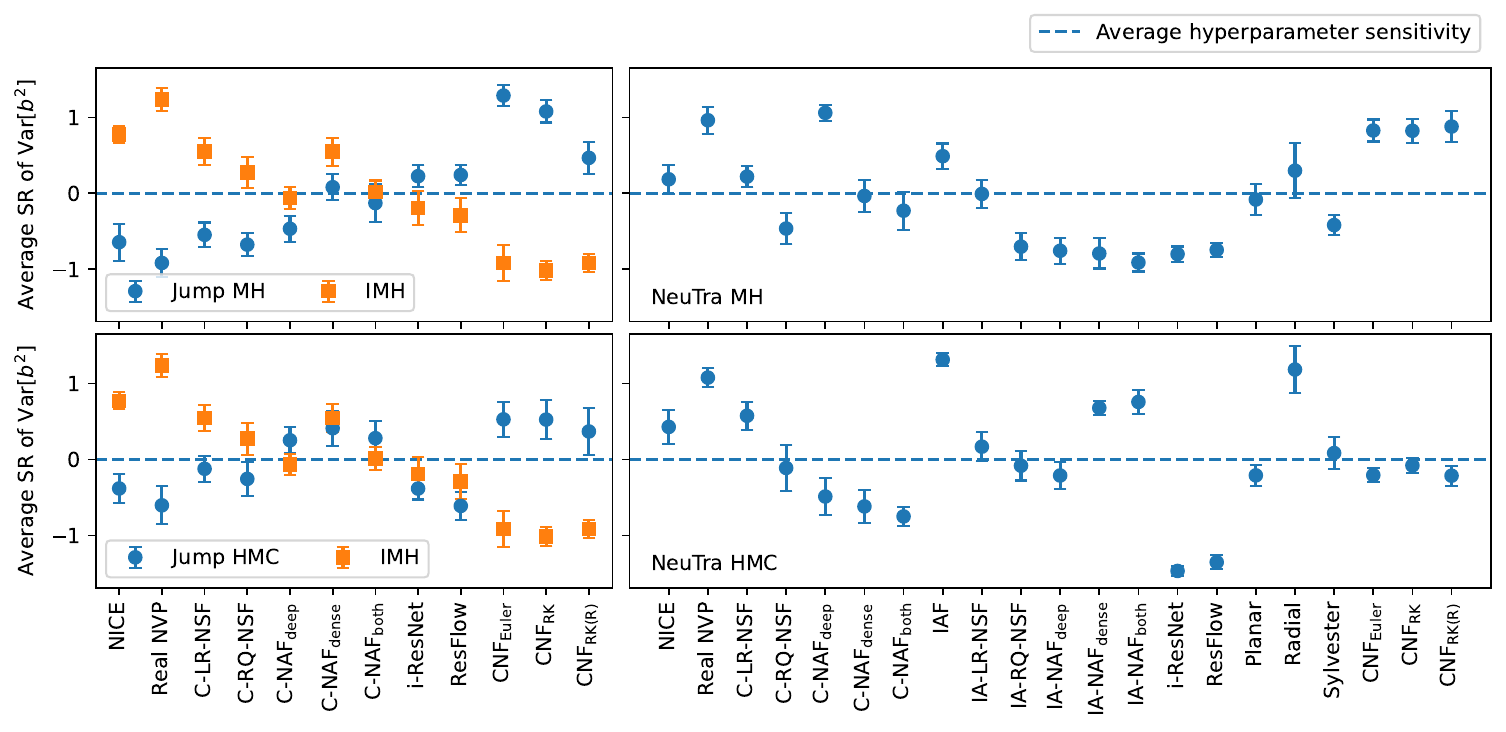}
    \caption{Hyperparameter sensitivity for different NFs across all benchmark targets for Jump MCMC (left) and \neutramcmc (right), described as $\overline{r} \pm \hat{\sigma}$ where NFs are ranked according to $\mathrm{Var}[b^2]$ across all hyperparameter configurations.}
    \label{fig:nf-sensitivity}
\end{figure}

We have observed several times that certain NF architectures exhibit a high standard error $\hat{\sigma}$ when estimating $\overline{r}$.
This partly depends on the variance of $b^2$ when testing the same architecture with multiple hyperparameter sets or different samplers.
If the variance is high, we expect the architecture to be sensitive to hyperparameter choice and thus require more time for hyperparameter tuning.
While our previous tests measured NF performance on our regular benchmark, we now further test NF stability when varying properties of synthetic distributions to make sampling more challenging.
We expect stable NFs to exhibit a low variance of $b^2$ across multiple different hyperparameter sets and samplers.
We separately rank NFs across all benchmark targets in terms of $\mathrm{Var}[b^2]$ in Figure~\ref{fig:nf-sensitivity}.

We find NF stability to be roughly the same in Jump MH and Jump HMC experiments.
Coupling NFs are the most stable architectures in Jump MH.
They rank somewhat worse in Jump HMC, where residual NFs perform better.
Continuous NFs are the least stable in both Jump MH and Jump HMC.
However, they are the best option for IMH, suggesting that they gain stability if we remove local MCMC transitions.
Conversely, coupling NFs lose stability if we add local transitions.

IA architectures rank the best in \neutramh, closely followed by i-ResNet and ResFlow.
The latter two are substantially better in \neutrahmc, which is also consistent with their improved stability after switching from Jump MH to Jump HMC.
The most stable \neutrahmc preconditioners are C-NAF architectures, i-ResNet, and ResFlow.
The radial flow is among the least stable ones.
Combined with our findings in Table~\ref{tab:nf-per-family-neutra}, its good performance is thus highly dependent on the choice of hyperparameters.
IAF is the least stable preconditioner for \neutrahmc, further suggesting that the original \neutrahmc formulation in~\citep{hoffman_neutra-lizing_2019} can easily be improved for general-purpose sampling with a different architecture like i-ResNet or ResFlow.

\subsubsection{NF recommendations for NFMC}
Practitioners interested in performing statistical analyses of models or distributions of their parameters are usually focused on obtaining the best quality samples and parameter estimates.
Considering our classification of distributions into four target families, we state the following:
\begin{itemize}
	\item If the practitioner has no knowledge of the target family and this knowledge is unattainable, we recommend Jump HMC with i-ResNet due to its good sampling performance and hyperparameter stability. One may also use other continuous or residual architectures. If the target has high dimensionality, curvature, or multimodality, we recommend coupling NFs instead (excluding NAF models). We also recommend such NFs over continuous NFs to reduce tuning time, as coupling NFs are less sensitive to the choice of hyperparameters in Jump MCMC.
	\item If the practitioner has knowledge of the target family, we recommend Jump HMC with C-RQ-NSF for approximately Gaussian targets, \cnfrkreg or \realnvp for unimodal non-Gaussian targets, and either \cnfrkreg or NICE for multimodal targets. If dealing with a general (real-world) Bayesian model posterior, we suggest i-ResNet or a continuous NF architecture.
\end{itemize}

Researchers can also be interested in NF performance to guide the design and development of new NFMC samplers:
\begin{itemize}
	\item If using NFs for jumps as independent global proposals, we suggest continuous NFs in both gradient-free and gradient-based cases as they offer the best gradient-based performance with or without hyperparameter tuning. However, we again note that continuous NFs may prove to be sensitive across hyperparameters depending on the underlying MCMC dynamics. A stabler option is coupling NFs (excluding NAF).
	\item If using NFs for preconditioning, we suggest the radial flow with default hyperparameters, contractive residual NFs, and IA architectures. If the target distribution has high dimensionality, strong curvature, many modes, or heavily unequal mode weights, one may opt for coupling NFs (excluding NAF).
    For a lower hyperparameter sensitivity, we suggest IA architectures in the gradient-free setting and contractive residual NFs in the gradient-based setting.
\end{itemize}

For target distributions similar to the ones in our benchmark, we suggest keeping NF parameter counts between roughly $10^3$ and $10^4$, especially if moment estimates with bigger or lower trainable parameter counts are poor.

\section{Conclusion}\label{sec:conclusion}

In this paper, we compared different NFMC methods and NF architectures in sampling from distributions.
We focused on the quality of second-moment estimation and tested many NFMC-NF combinations on various targets across four distribution families.
We focused on variations of HMC and MH as gradient-based and gradient-free MCMC representatives, extending them with independent NF jumps and NF-based preconditioning.
When comparing MCMC to NFMC with off-the-shelf hyperparameters, we found HMC to perform better than \neutrahmc and comparable to Jump HMC.
IMH, Jump MH, and \neutramh all outperformed MH. 
When picking the best NF architecture and hyperparameters for a particular target, we found Jump HMC to outperform all other samplers.
IMH, Jump MH, and \neutramh all outperformed MH.
In summary, we found jumps beneficial in all cases and preconditioning useful for gradient-free sampling.
We found i-ResNet to generally be the best architecture for NFMC on our benchmark, followed by other residual NFs.
NFMC samplers should consider such NFs as candidates, especially since many previously proposed samplers perform NF jumps or preconditioning with \realnvp and NSF models.
Aside from NAF, we found coupling NFs to attain relatively average results.
However, they were the most robust when the geometric complexity of the target was increased.
We found the radial flow to rank best in \neutramcmc with default hyperparameters. 
However, it was very sensitive to hyperparameter choice and performed poorly on complex synthetic targets.

Given our findings, it would be practical to re-evaluate and compare other NFMC samplers~\citep{grumitt_deterministic_2022, cabezas_transport_2023}, as well as transport methods~\citep{karamanis_accelerating_2022, wu_stochastic_2020, arbel_annealed_2021, matthews_continual_2022} and other sampling methods~\citep{grumitt_flow_2024} with the best architectures for the corresponding target families.
We similarly suggest using the highlighted NFs for the development of future samplers or as methods to be extended into NFMC-specific NF architectures.
Our results also suggest evaluating current and future NFMC samplers across various targets and, importantly, comparing them with classic MCMC methods to assess their practical uses.

\subsection{Limitations}
Our focus was on evaluating and comparing NF architectures within NFMC, which we performed with extensions of MH and HMC.
Due to a combinatorial explosion of the number of possible experiments, we did not analyze other MCMC and NFMC samplers. 
Doing so could let us exhaustively compare different NFMC samplers, which would be highly relevant to the development of the field.
We opted to first compare NFs, so we leave an exhaustive comparison of NFMC methods as future work.
In our results, we sometimes estimate uncertainties based on a small number of distributions belonging to a target family. By adding more targets, we could arrive at better rank estimates and smaller standard errors.

Our results are based on over 10 thousand experiments. We paid careful attention to successfully execute each experiment.
However, some experiments with specific combinations of samplers, NFs, and targets were not completed successfully.
While rare, this was mostly due to the slow optimization of NFs with many trainable parameters, which occurred in high dimensional targets where we adaptively increased the NF parameter count to enable expressive modeling.
Some experiments failed due to numerical instabilities in sampling from ill-posed targets or those stemming from sampler and NF definitions.
We mitigated these issues by performing over 10 thousand automated tests for the numerical stability of samplers and NFs (forward and inverse passes, log probability computation, NF and MCMC sampling, and autodifferentiation).

We performed our experiments in PyTorch~\citep{paszke_pytorch_2019}. Several works show that using packages with just-in-time compilation, like Jax~\citep{bradbury_jax_2018}, can vastly speed up program execution.
We opted for PyTorch as its object-oriented development paradigm allowed us to modularly implement and test each model.
With Jax, managing such a large code base would require substantially more engineering effort.
However, its faster execution speed could somewhat change the relative ranks of NFs and NFMC samplers in our results.
Nevertheless, our results show that some combinations of NFs and NFMC methods consistently yield better results than others.
Having identified these combinations, future research in NFMC can take them as initial models, refine and tune them according to their own target distribution needs, and finally implement them in Jax for maximum performance.

\subsection{Data availability}
Data for likelihood functions in real-world experiments is available at~\url{https://github.com/davidnabergoj/posteriordb}.
This is a copy of the main repository, available at~\url{https://github.com/stan-dev/posteriordb}.
Both addresses were accessed on January 16, 2025.

\subsection{Code availability}
All code is publicly available  (accessed: October 8, 2025):
\begin{itemize}
    \item NF implementations: \url{https://github.com/davidnabergoj/torchflows}.
    \item Sampler implementations: \url{https://github.com/davidnabergoj/nfmc}.
    \item Target distribution benchmark: \url{https://github.com/davidnabergoj/potentials}.
    \item Evaluation scripts: \url{https://github.com/davidnabergoj/nfmc-nf-evaluation}.
\end{itemize}

\backmatter

\bmhead{Acknowledgements}
This work was supported by the Slovenian Research and Innovation Agency (ARIS) grant P2-0442.

\newpage
\bibliography{sn-bibliography}


\begin{thebibliography}{52}
\ifx \bisbn   \undefined \def \bisbn  #1{ISBN #1}\fi
\ifx \binits  \undefined \def \binits#1{#1}\fi
\ifx \bauthor  \undefined \def \bauthor#1{#1}\fi
\ifx \batitle  \undefined \def \batitle#1{#1}\fi
\ifx \bjtitle  \undefined \def \bjtitle#1{#1}\fi
\ifx \bvolume  \undefined \def \bvolume#1{\textbf{#1}}\fi
\ifx \byear  \undefined \def \byear#1{#1}\fi
\ifx \bissue  \undefined \def \bissue#1{#1}\fi
\ifx \bfpage  \undefined \def \bfpage#1{#1}\fi
\ifx \blpage  \undefined \def \blpage #1{#1}\fi
\ifx \burl  \undefined \def \burl#1{\textsf{#1}}\fi
\ifx \doiurl  \undefined \def \doiurl#1{\url{https://doi.org/#1}}\fi
\ifx \betal  \undefined \def \betal{\textit{et al.}}\fi
\ifx \binstitute  \undefined \def \binstitute#1{#1}\fi
\ifx \binstitutionaled  \undefined \def \binstitutionaled#1{#1}\fi
\ifx \bctitle  \undefined \def \bctitle#1{#1}\fi
\ifx \beditor  \undefined \def \beditor#1{#1}\fi
\ifx \bpublisher  \undefined \def \bpublisher#1{#1}\fi
\ifx \bbtitle  \undefined \def \bbtitle#1{#1}\fi
\ifx \bedition  \undefined \def \bedition#1{#1}\fi
\ifx \bseriesno  \undefined \def \bseriesno#1{#1}\fi
\ifx \blocation  \undefined \def \blocation#1{#1}\fi
\ifx \bsertitle  \undefined \def \bsertitle#1{#1}\fi
\ifx \bsnm \undefined \def \bsnm#1{#1}\fi
\ifx \bsuffix \undefined \def \bsuffix#1{#1}\fi
\ifx \bparticle \undefined \def \bparticle#1{#1}\fi
\ifx \barticle \undefined \def \barticle#1{#1}\fi
\bibcommenthead
\ifx \bconfdate \undefined \def \bconfdate #1{#1}\fi
\ifx \botherref \undefined \def \botherref #1{#1}\fi
\ifx \url \undefined \def \url#1{\textsf{#1}}\fi
\ifx \bchapter \undefined \def \bchapter#1{#1}\fi
\ifx \bbook \undefined \def \bbook#1{#1}\fi
\ifx \bcomment \undefined \def \bcomment#1{#1}\fi
\ifx \oauthor \undefined \def \oauthor#1{#1}\fi
\ifx \citeauthoryear \undefined \def \citeauthoryear#1{#1}\fi
\ifx \endbibitem  \undefined \def \endbibitem {}\fi
\ifx \bconflocation  \undefined \def \bconflocation#1{#1}\fi
\ifx \arxivurl  \undefined \def \arxivurl#1{\textsf{#1}}\fi
\csname PreBibitemsHook\endcsname

\bibitem[\protect\citeauthoryear{Abbott et~al.}{2023}]{abbott_aspects_2023}
\begin{barticle}
\bauthor{\bsnm{Abbott}, \binits{R.}},
\bauthor{\bsnm{Albergo}, \binits{M.S.}},
\bauthor{\bsnm{Botev}, \binits{A.}},
\bauthor{\bsnm{Boyda}, \binits{D.}},
\bauthor{\bsnm{Cranmer}, \binits{K.}},
\bauthor{\bsnm{Hackett}, \binits{D.C.}},
\bauthor{\bsnm{Matthews}, \binits{A.G.D.G.}},
\bauthor{\bsnm{Racani\`{e}re}, \binits{S.}},
\bauthor{\bsnm{Razavi}, \binits{A.}},
\bauthor{\bsnm{Rezende}, \binits{D.J.}},
\bauthor{\bsnm{Romero-L\'{o}pez}, \binits{F.}},
\bauthor{\bsnm{Shanahan}, \binits{P.E.}},
\bauthor{\bsnm{Urban}, \binits{J.M.}}:
\batitle{Aspects of scaling and scalability for flow-based sampling of lattice {QCD}}.
\bjtitle{The European Physical Journal A}
\bvolume{59}(\bissue{11}),
\bfpage{257}
(\byear{2023})
\doiurl{10.1140/epja/s10050-023-01154-w}
\end{barticle}
\endbibitem

\bibitem[\protect\citeauthoryear{Agrawal and Domke}{2024}]{agrawal_disentangling_2024}
\begin{botherref}
\oauthor{\bsnm{Agrawal}, \binits{A.}},
\oauthor{\bsnm{Domke}, \binits{J.}}:
Disentangling impact of capacity, objective, batchsize, estimators, and step-size on flow {VI}.
arXiv.
arXiv:2412.08824
(2024).
\doiurl{10.48550/arXiv.2412.08824}
\end{botherref}
\endbibitem

\bibitem[\protect\citeauthoryear{Alquier et~al.}{2016}]{alquier_noisy_2016}
\begin{barticle}
\bauthor{\bsnm{Alquier}, \binits{P.}},
\bauthor{\bsnm{Friel}, \binits{N.}},
\bauthor{\bsnm{Everitt}, \binits{R.}},
\bauthor{\bsnm{Boland}, \binits{A.}}:
\batitle{Noisy {Monte Carlo}: Convergence of {Markov} chains with approximate transition kernels}.
\bjtitle{Statistics and Computing}
\bvolume{26}(\bissue{1}),
\bfpage{29}--\blpage{47}
(\byear{2016})
\doiurl{10.1007/s11222-014-9521-x}
\end{barticle}
\endbibitem

\bibitem[\protect\citeauthoryear{Albergo et~al.}{2019}]{albergo_flow-based_2019}
\begin{barticle}
\bauthor{\bsnm{Albergo}, \binits{M.S.}},
\bauthor{\bsnm{Kanwar}, \binits{G.}},
\bauthor{\bsnm{Shanahan}, \binits{P.E.}}:
\batitle{Flow-based generative models for {Markov} chain {Monte} {Carlo} in lattice field theory}.
\bjtitle{Physical Review D}
\bvolume{100}(\bissue{3}),
\bfpage{034515}
(\byear{2019})
\doiurl{10.1103/PhysRevD.100.034515}
\end{barticle}
\endbibitem

\bibitem[\protect\citeauthoryear{Arbel et~al.}{2021}]{arbel_annealed_2021}
\begin{bchapter}
\bauthor{\bsnm{Arbel}, \binits{M.}},
\bauthor{\bsnm{Matthews}, \binits{A.}},
\bauthor{\bsnm{Doucet}, \binits{A.}}:
\bctitle{Annealed {Flow} {Transport} {Monte} {Carlo}}.
In: \bbtitle{Proceedings of the 38th {International} {Conference} on {Machine} {Learning}}.
\bsertitle{Proceedings of Machine Learning Research},
vol. \bseriesno{139},
pp. \bfpage{318}--\blpage{330}.
\bpublisher{PMLR},
\blocation{Vienna, Austria (virtual conference)}
(\byear{2021})
\end{bchapter}
\endbibitem

\bibitem[\protect\citeauthoryear{Bradbury et~al.}{2018}]{bradbury_jax_2018}
\begin{botherref}
\oauthor{\bsnm{Bradbury}, \binits{J.}},
\oauthor{\bsnm{Frostig}, \binits{R.}},
\oauthor{\bsnm{Hawkins}, \binits{P.}},
\oauthor{\bsnm{Johnson}, \binits{M.J.}},
\oauthor{\bsnm{Leary}, \binits{C.}},
\oauthor{\bsnm{Maclaurin}, \binits{D.}},
\oauthor{\bsnm{Necula}, \binits{G.}},
\oauthor{\bsnm{Paszke}, \binits{A.}},
\oauthor{\bsnm{Vander{P}las}, \binits{J.}},
\oauthor{\bsnm{Wanderman-{M}ilne}, \binits{S.}},
\oauthor{\bsnm{Zhang}, \binits{Q.}}:
{JAX}: composable transformations of {P}ython+{N}um{P}y programs
(2018).
\url{http://github.com/jax-ml/jax}
\end{botherref}
\endbibitem

\bibitem[\protect\citeauthoryear{Brofos et~al.}{2022}]{brofos_adaptation_2022}
\begin{bchapter}
\bauthor{\bsnm{Brofos}, \binits{J.}},
\bauthor{\bsnm{Gabri\'{e}}, \binits{M.}},
\bauthor{\bsnm{Brubaker}, \binits{M.A.}},
\bauthor{\bsnm{Lederman}, \binits{R.R.}}:
\bctitle{Adaptation of the {Independent} {Metropolis}-{Hastings} {Sampler} with {Normalizing} {Flow} {Proposals}}.
In: \bbtitle{Proceedings of {The} 25th {International} {Conference} on {Artificial} {Intelligence} And {Statistics}}.
\bsertitle{Proceedings of Machine Learning Research},
vol. \bseriesno{151},
pp. \bfpage{5949}--\blpage{5986}.
\bpublisher{PMLR},
\blocation{Virtual conference}
(\byear{2022})
\end{bchapter}
\endbibitem

\bibitem[\protect\citeauthoryear{Behrmann et~al.}{2019}]{behrmann_invertible_2019}
\begin{bchapter}
\bauthor{\bsnm{Behrmann}, \binits{J.}},
\bauthor{\bsnm{Grathwohl}, \binits{W.}},
\bauthor{\bsnm{Chen}, \binits{R.T.Q.}},
\bauthor{\bsnm{Duvenaud}, \binits{D.}},
\bauthor{\bsnm{Jacobsen}, \binits{J.-H.}}:
\bctitle{Invertible {Residual} {Networks}}.
In: \bbtitle{Proceedings of the 36th {International} {Conference} on {Machine} {Learning}}.
\bsertitle{Proceedings of Machine Learning Research},
vol. \bseriesno{97},
pp. \bfpage{573}--\blpage{582}.
\bpublisher{PMLR},
\blocation{Los Angeles, United States}
(\byear{2019})
\end{bchapter}
\endbibitem

\bibitem[\protect\citeauthoryear{Berg et~al.}{2018}]{berg_sylvester_2018}
\begin{bchapter}
\bauthor{\bsnm{Berg}, \binits{R.v.d.}},
\bauthor{\bsnm{Hasenclever}, \binits{L.}},
\bauthor{\bsnm{Tomczak}, \binits{J.M.}},
\bauthor{\bsnm{Welling}, \binits{M.}}:
\bctitle{Sylvester {Normalizing} {Flows} for {Variational} {Inference}}.
In: \bbtitle{Proceedings of the 34th {Conference} on {Uncertainty} in {Artificial} {Intelligence}},
pp. \bfpage{393}--\blpage{402}.
\bpublisher{AUAI Press},
\blocation{Monterey, United States}
(\byear{2018})
\end{bchapter}
\endbibitem

\bibitem[\protect\citeauthoryear{Chen et~al.}{2019}]{chen_residual_2019}
\begin{bchapter}
\bauthor{\bsnm{Chen}, \binits{R.T.Q.}},
\bauthor{\bsnm{Behrmann}, \binits{J.}},
\bauthor{\bsnm{Duvenaud}, \binits{D.}},
\bauthor{\bsnm{Jacobsen}, \binits{J.-H.}}:
\bctitle{Residual {Flows} for {Invertible} {Generative} {Modeling}}.
In: \bbtitle{Advances in {Neural} {Information} {Processing} {Systems}},
vol. \bseriesno{32}.
\bpublisher{Curran Associates, Inc.},
\blocation{Vancouver, Canada}
(\byear{2019})
\end{bchapter}
\endbibitem

\bibitem[\protect\citeauthoryear{Cornish et~al.}{2020}]{cornish_relaxing_2020}
\begin{bchapter}
\bauthor{\bsnm{Cornish}, \binits{R.}},
\bauthor{\bsnm{Caterini}, \binits{A.}},
\bauthor{\bsnm{Deligiannidis}, \binits{G.}},
\bauthor{\bsnm{Doucet}, \binits{A.}}:
\bctitle{Relaxing {Bijectivity} {Constraints} with {Continuously} {Indexed} {Normalising} {Flows}}.
In: \bbtitle{Proceedings of the 37th {International} {Conference} on {Machine} {Learning}}.
\bsertitle{Proceedings of Machine Learning Research},
vol. \bseriesno{119},
pp. \bfpage{2133}--\blpage{2143}.
\bpublisher{PMLR},
\blocation{Vienna, Austria (virtual conference)}
(\byear{2020})
\end{bchapter}
\endbibitem

\bibitem[\protect\citeauthoryear{Chan and Geyer}{1994}]{chan_discussion_1994}
\begin{barticle}
\bauthor{\bsnm{Chan}, \binits{K.S.}},
\bauthor{\bsnm{Geyer}, \binits{C.J.}}:
\batitle{Discussion: {Markov} {Chains} for {Exploring} {Posterior} {Distributions}}.
\bjtitle{The Annals of Statistics}
\bvolume{22}(\bissue{4}),
\bfpage{1747}--\blpage{1758}
(\byear{1994})
\doiurl{10.1214/aos/1176325754}
\end{barticle}
\endbibitem

\bibitem[\protect\citeauthoryear{Cabezas and Nemeth}{2023}]{cabezas_transport_2023}
\begin{bchapter}
\bauthor{\bsnm{Cabezas}, \binits{A.}},
\bauthor{\bsnm{Nemeth}, \binits{C.}}:
\bctitle{Transport {Elliptical} {Slice} {Sampling}}.
In: \bbtitle{Proceedings of {The} 26th {International} {Conference} on {Artificial} {Intelligence} And {Statistics}}.
\bsertitle{Proceedings of Machine Learning Research},
vol. \bseriesno{206},
pp. \bfpage{3664}--\blpage{3676}.
\bpublisher{PMLR},
\blocation{Valencia, Spain}
(\byear{2023})
\end{bchapter}
\endbibitem

\bibitem[\protect\citeauthoryear{Cabezas et~al.}{2024}]{cabezas_markovian_2024}
\begin{bchapter}
\bauthor{\bsnm{Cabezas}, \binits{A.}},
\bauthor{\bsnm{Sharrock}, \binits{L.}},
\bauthor{\bsnm{Nemeth}, \binits{C.}}:
\bctitle{Markovian {Flow} {Matching}: {Accelerating} {MCMC} with {Continuous} {Normalizing} {Flows}}.
In: \bbtitle{Advances in {Neural} {Information} {Processing} {Systems}},
vol. \bseriesno{37},
pp. \bfpage{104383}--\blpage{104411}.
\bpublisher{Curran Associates, Inc.},
\blocation{Vancouver, Canada}
(\byear{2024})
\end{bchapter}
\endbibitem

\bibitem[\protect\citeauthoryear{Durkan et~al.}{2019}]{durkan_neural_2019}
\begin{bchapter}
\bauthor{\bsnm{Durkan}, \binits{C.}},
\bauthor{\bsnm{Bekasov}, \binits{A.}},
\bauthor{\bsnm{Murray}, \binits{I.}},
\bauthor{\bsnm{Papamakarios}, \binits{G.}}:
\bctitle{Neural {Spline} {Flows}}.
In: \bbtitle{Advances in {Neural} {Information} {Processing} {Systems}},
vol. \bseriesno{32}.
\bpublisher{Curran Associates, Inc.},
\blocation{Vancouver, Canada}
(\byear{2019})
\end{bchapter}
\endbibitem

\bibitem[\protect\citeauthoryear{Del~Debbio et~al.}{2021}]{del_debbio_efficient_2021}
\begin{barticle}
\bauthor{\bsnm{Del~Debbio}, \binits{L.}},
\bauthor{\bsnm{Marsh~Rossney}, \binits{J.}},
\bauthor{\bsnm{Wilson}, \binits{M.}}:
\batitle{Efficient {Modelling} of {Trivializing} {Maps} for {Lattice} {$\phi^4$} {Theory} {Using} {Normalizing} {Flows}: {A} {First} {Look} at {Scalability}}.
\bjtitle{Physical Review D}
\bvolume{104}(\bissue{9}),
\bfpage{094507}
(\byear{2021})
\doiurl{10.22323/1.396.0059}
\end{barticle}
\endbibitem

\bibitem[\protect\citeauthoryear{Dolatabadi et~al.}{2020}]{dolatabadi_invertible_2020}
\begin{bchapter}
\bauthor{\bsnm{Dolatabadi}, \binits{H.M.}},
\bauthor{\bsnm{Erfani}, \binits{S.}},
\bauthor{\bsnm{Leckie}, \binits{C.}}:
\bctitle{Invertible {Generative} {Modeling} using {Linear} {Rational} {Splines}}.
In: \bbtitle{Proceedings of {The} 23rd {International} {Conference} on {Artificial} {Intelligence} And {Statistics}}.
\bsertitle{Proceedings of Machine Learning Research},
vol. \bseriesno{108},
pp. \bfpage{4236}--\blpage{4246}.
\bpublisher{PMLR},
\blocation{Virtual conference}
(\byear{2020})
\end{bchapter}
\endbibitem

\bibitem[\protect\citeauthoryear{Dinh et~al.}{2015}]{dinh_nice_2015}
\begin{botherref}
\oauthor{\bsnm{Dinh}, \binits{L.}},
\oauthor{\bsnm{Krueger}, \binits{D.}},
\oauthor{\bsnm{Bengio}, \binits{Y.}}:
{NICE}: {Non}-linear {Independent} {Components} {Estimation}.
arXiv.
arXiv:1410.8516
(2015).
\doiurl{10.48550/arXiv.1410.8516}
\end{botherref}
\endbibitem

\bibitem[\protect\citeauthoryear{Dinh et~al.}{2017}]{dinh_density_2017}
\begin{botherref}
\oauthor{\bsnm{Dinh}, \binits{L.}},
\oauthor{\bsnm{Sohl-Dickstein}, \binits{J.}},
\oauthor{\bsnm{Bengio}, \binits{S.}}:
Density estimation using {Real} {NVP}.
arXiv.
arXiv:1605.08803
(2017).
\doiurl{10.48550/arXiv.1605.08803}
\end{botherref}
\endbibitem

\bibitem[\protect\citeauthoryear{Draxler et~al.}{2024}]{draxler_universality_2024}
\begin{bchapter}
\bauthor{\bsnm{Draxler}, \binits{F.}},
\bauthor{\bsnm{Wahl}, \binits{S.}},
\bauthor{\bsnm{Schn\"{o}rr}, \binits{C.}},
\bauthor{\bsnm{K\"{o}the}, \binits{U.}}:
\bctitle{On the {Universality} of {Volume}-{Preserving} and {Coupling}-{Based} {Normalizing} {Flows}}.
In: \bbtitle{Proceedings of the 41st {International} {Conference} on {Machine} {Learning}}.
\bsertitle{Proceedings of Machine Learning Research},
vol. \bseriesno{235},
pp. \bfpage{11613}--\blpage{11641}.
\bpublisher{PMLR},
\blocation{Vienna, Austria}
(\byear{2024})
\end{bchapter}
\endbibitem

\bibitem[\protect\citeauthoryear{Finlay et~al.}{2020}]{finlay_how_2020}
\begin{bchapter}
\bauthor{\bsnm{Finlay}, \binits{C.}},
\bauthor{\bsnm{Jacobsen}, \binits{J.-H.}},
\bauthor{\bsnm{Nurbekyan}, \binits{L.}},
\bauthor{\bsnm{Oberman}, \binits{A.}}:
\bctitle{How to {Train} {Your} {Neural} {ODE}: the {World} of {Jacobian} and {Kinetic} {Regularization}}.
In: \bbtitle{Proceedings of the 37th {International} {Conference} on {Machine} {Learning}}.
\bsertitle{Proceedings of Machine Learning Research},
vol. \bseriesno{119},
pp. \bfpage{3154}--\blpage{3164}.
\bpublisher{PMLR},
\blocation{Vienna, Austria (virtual conference)}
(\byear{2020})
\end{bchapter}
\endbibitem

\bibitem[\protect\citeauthoryear{Grathwohl et~al.}{2018}]{grathwohl_ffjord_2018}
\begin{botherref}
\oauthor{\bsnm{Grathwohl}, \binits{W.}},
\oauthor{\bsnm{Chen}, \binits{R.T.Q.}},
\oauthor{\bsnm{Bettencourt}, \binits{J.}},
\oauthor{\bsnm{Sutskever}, \binits{I.}},
\oauthor{\bsnm{Duvenaud}, \binits{D.}}:
{FFJORD}: {Free}-form {Continuous} {Dynamics} for {Scalable} {Reversible} {Generative} {Models}.
arXiv.
arXiv:1810.01367
(2018).
\doiurl{10.48550/arXiv.1810.01367}
\end{botherref}
\endbibitem

\bibitem[\protect\citeauthoryear{Grenioux et~al.}{2023}]{grenioux_sampling_2023}
\begin{bchapter}
\bauthor{\bsnm{Grenioux}, \binits{L.}},
\bauthor{\bsnm{Durmus}, \binits{A.O.}},
\bauthor{\bsnm{Moulines}, \binits{E.}},
\bauthor{\bsnm{Gabri\'{e}}, \binits{M.}}:
\bctitle{On sampling with approximate transport maps}.
In: \bbtitle{Proceedings of the 40th {International} {Conference} on {Machine} {Learning}}.
\bsertitle{Proceedings of Machine Learning Research},
vol. \bseriesno{202},
pp. \bfpage{11698}--\blpage{11733}.
\bpublisher{PMLR},
\blocation{Honolulu, United States}
(\byear{2023})
\end{bchapter}
\endbibitem

\bibitem[\protect\citeauthoryear{Grumitt et~al.}{2022}]{grumitt_deterministic_2022}
\begin{bchapter}
\bauthor{\bsnm{Grumitt}, \binits{R.}},
\bauthor{\bsnm{Dai}, \binits{B.}},
\bauthor{\bsnm{Seljak}, \binits{U.}}:
\bctitle{Deterministic {Langevin} {Monte} {Carlo} with {Normalizing} {Flows} for {Bayesian} {Inference}}.
In: \bbtitle{Advances in Neural Information Processing Systems},
vol. \bseriesno{35},
pp. \bfpage{11629}--\blpage{11641}.
\bpublisher{Curran Associates, Inc.},
\blocation{New Orleans, United States}
(\byear{2022})
\end{bchapter}
\endbibitem

\bibitem[\protect\citeauthoryear{Germain et~al.}{2015}]{germain_made_2015}
\begin{bchapter}
\bauthor{\bsnm{Germain}, \binits{M.}},
\bauthor{\bsnm{Gregor}, \binits{K.}},
\bauthor{\bsnm{Murray}, \binits{I.}},
\bauthor{\bsnm{Larochelle}, \binits{H.}}:
\bctitle{{MADE}: {Masked} {Autoencoder} for {Distribution} {Estimation}}.
In: \bbtitle{Proceedings of the 32nd {International} {Conference} on {Machine} {Learning}}.
\bsertitle{Proceedings of Machine Learning Research},
vol. \bseriesno{37},
pp. \bfpage{881}--\blpage{889}.
\bpublisher{PMLR},
\blocation{Lille, France}
(\byear{2015})
\end{bchapter}
\endbibitem

\bibitem[\protect\citeauthoryear{Grumitt et~al.}{2024}]{grumitt_flow_2024}
\begin{barticle}
\bauthor{\bsnm{Grumitt}, \binits{R.D.P.}},
\bauthor{\bsnm{Karamanis}, \binits{M.}},
\bauthor{\bsnm{Seljak}, \binits{U.}}:
\batitle{Flow {Annealed} {Kalman} {Inversion} for {Gradient}-{Free} {Inference} in {Bayesian} {Inverse} {Problems}}.
\bjtitle{Physical Sciences Forum}
\bvolume{9}(\bissue{1}),
\bfpage{21}
(\byear{2024})
\doiurl{10.3390/psf2023009021}
\end{barticle}
\endbibitem

\bibitem[\protect\citeauthoryear{Gabrié et~al.}{2022}]{gabrie_adaptive_2022}
\begin{barticle}
\bauthor{\bsnm{Gabrié}, \binits{M.}},
\bauthor{\bsnm{Rotskoff}, \binits{G.M.}},
\bauthor{\bsnm{Vanden-Eijnden}, \binits{E.}}:
\batitle{Adaptive {Monte} {Carlo} augmented with normalizing flows}.
\bjtitle{Proceedings of the National Academy of Sciences}
\bvolume{119}(\bissue{10}),
\bfpage{2109420119}
(\byear{2022})
\doiurl{10.1073/pnas.2109420119}
\end{barticle}
\endbibitem

\bibitem[\protect\citeauthoryear{Hoffman and Gelman}{2011}]{hoffman_no-u-turn_2011}
\begin{botherref}
\oauthor{\bsnm{Hoffman}, \binits{M.D.}},
\oauthor{\bsnm{Gelman}, \binits{A.}}:
The {No}-{U}-{Turn} {Sampler}: {Adaptively} {Setting} {Path} {Lengths} in {Hamiltonian} {Monte} {Carlo}.
arXiv.
arXiv:1111.4246
(2011).
\doiurl{10.48550/arXiv.1111.4246}
\end{botherref}
\endbibitem

\bibitem[\protect\citeauthoryear{Huang et~al.}{2018}]{huang_neural_2018}
\begin{bchapter}
\bauthor{\bsnm{Huang}, \binits{C.-W.}},
\bauthor{\bsnm{Krueger}, \binits{D.}},
\bauthor{\bsnm{Lacoste}, \binits{A.}},
\bauthor{\bsnm{Courville}, \binits{A.}}:
\bctitle{Neural {Autoregressive} {Flows}}.
In: \bbtitle{Proceedings of the 35th {International} {Conference} on {Machine} {Learning}}.
\bsertitle{Proceedings of Machine Learning Research},
vol. \bseriesno{80},
pp. \bfpage{2078}--\blpage{2087}.
\bpublisher{PMLR},
\blocation{Stockholm, Sweden}
(\byear{2018})
\end{bchapter}
\endbibitem

\bibitem[\protect\citeauthoryear{Hoffman et~al.}{2019}]{hoffman_neutra-lizing_2019}
\begin{botherref}
\oauthor{\bsnm{Hoffman}, \binits{M.D.}},
\oauthor{\bsnm{Sountsov}, \binits{P.}},
\oauthor{\bsnm{Dillon}, \binits{J.V.}},
\oauthor{\bsnm{Langmore}, \binits{I.}},
\oauthor{\bsnm{Tran}, \binits{D.}},
\oauthor{\bsnm{Vasudevan}, \binits{S.}}:
{NeuTra}-lizing {Bad} {Geometry} in {Hamiltonian} {Monte} {Carlo} {Using} {Neural} {Transport}.
arXiv.
arXiv:1903.03704
(2019).
\doiurl{10.48550/arXiv.1903.03704}
\end{botherref}
\endbibitem

\bibitem[\protect\citeauthoryear{Hutchinson}{1989}]{hutchinson_stochastic_1989}
\begin{barticle}
\bauthor{\bsnm{Hutchinson}, \binits{M.F.}}:
\batitle{A stochastic estimator of the trace of the influence matrix for {Laplacian} smoothing splines}.
\bjtitle{Communication in Statistics -- Simulation and Computation}
\bvolume{18},
\bfpage{1059}--\blpage{1076}
(\byear{1989})
\doiurl{10.1080/03610919008812866}
\end{barticle}
\endbibitem

\bibitem[\protect\citeauthoryear{Karamanis et~al.}{2022}]{karamanis_accelerating_2022}
\begin{barticle}
\bauthor{\bsnm{Karamanis}, \binits{M.}},
\bauthor{\bsnm{Beutler}, \binits{F.}},
\bauthor{\bsnm{Peacock}, \binits{J.A.}},
\bauthor{\bsnm{Nabergoj}, \binits{D.}},
\bauthor{\bsnm{Seljak}, \binits{U.}}:
\batitle{Accelerating astronomical and cosmological inference with preconditioned {Monte} {Carlo}}.
\bjtitle{Monthly Notices of the Royal Astronomical Society}
\bvolume{516}(\bissue{2}),
\bfpage{1644}--\blpage{1653}
(\byear{2022}).
\bcomment{Oxford University Press}
\end{barticle}
\endbibitem

\bibitem[\protect\citeauthoryear{Kingma et~al.}{2016}]{kingma_improved_2016}
\begin{bchapter}
\bauthor{\bsnm{Kingma}, \binits{D.P.}},
\bauthor{\bsnm{Salimans}, \binits{T.}},
\bauthor{\bsnm{Jozefowicz}, \binits{R.}},
\bauthor{\bsnm{Chen}, \binits{X.}},
\bauthor{\bsnm{Sutskever}, \binits{I.}},
\bauthor{\bsnm{Welling}, \binits{M.}}:
\bctitle{Improved {Variational} {Inference} with {Inverse} {Autoregressive} {Flow}}.
In: \bbtitle{Advances in {Neural} {Information} {Processing} {Systems}},
vol. \bseriesno{29}.
\bpublisher{Curran Associates, Inc.},
\blocation{Barcelona, Spain}
(\byear{2016})
\end{bchapter}
\endbibitem

\bibitem[\protect\citeauthoryear{Liu et~al.}{2016}]{liu_kernelized_2016}
\begin{bchapter}
\bauthor{\bsnm{Liu}, \binits{Q.}},
\bauthor{\bsnm{Lee}, \binits{J.D.}},
\bauthor{\bsnm{Jordan}, \binits{M.}}:
\bctitle{A kernelized stein discrepancy for goodness-of-fit tests}.
In: \bbtitle{Proceedings of the 33rd {International} {Conference} on {Machine} {Learning}}.
\bsertitle{Proceedings of Machine Learning Research},
vol. \bseriesno{48},
pp. \bfpage{276}--\blpage{284}.
\bpublisher{PMLR},
\blocation{New York, United States}
(\byear{2016})
\end{bchapter}
\endbibitem

\bibitem[\protect\citeauthoryear{Lee et~al.}{2021}]{lee_universal_2021}
\begin{bchapter}
\bauthor{\bsnm{Lee}, \binits{H.}},
\bauthor{\bsnm{Pabbaraju}, \binits{C.}},
\bauthor{\bsnm{Sevekari}, \binits{A.P.}},
\bauthor{\bsnm{Risteski}, \binits{A.}}:
\bctitle{Universal {Approximation} {Using} {Well}-{Conditioned} {Normalizing} {Flows}}.
In: \bbtitle{Advances in {Neural} {Information} {Processing} {Systems}},
vol. \bseriesno{34},
pp. \bfpage{12700}--\blpage{12711}.
\bpublisher{Curran Associates, Inc.},
\blocation{Virtual conference}
(\byear{2021})
\end{bchapter}
\endbibitem

\bibitem[\protect\citeauthoryear{Matthews et~al.}{2022}]{matthews_continual_2022}
\begin{bchapter}
\bauthor{\bsnm{Matthews}, \binits{A.G.D.G.}},
\bauthor{\bsnm{Arbel}, \binits{M.}},
\bauthor{\bsnm{Rezende}, \binits{D.J.}},
\bauthor{\bsnm{Doucet}, \binits{A.}}:
\bctitle{Continual {Repeated} {Annealed} {Flow} {Transport} {Monte} {Carlo}}.
In: \bbtitle{Proceedings of the 39th {International} {Conference} on {Machine} {Learning}}.
\bsertitle{Proceedings of Machine Learning Research},
vol. \bseriesno{162},
pp. \bfpage{15196}--\blpage{15219}.
\bpublisher{PMLR},
\blocation{Baltimore, United States}
(\byear{2022})
\end{bchapter}
\endbibitem

\bibitem[\protect\citeauthoryear{Mitrophanov}{2005}]{mitrophanov_sensitivity_2005}
\begin{barticle}
\bauthor{\bsnm{Mitrophanov}, \binits{A.Y.}}:
\batitle{Sensitivity and convergence of uniformly ergodic {Markov} chains}.
\bjtitle{Journal of Applied Probability}
\bvolume{42}(\bissue{4}),
\bfpage{1003}--\blpage{1014}
(\byear{2005})
\doiurl{10.1239/jap/1134587812}
\end{barticle}
\endbibitem

\bibitem[\protect\citeauthoryear{Midgley et~al.}{2023}]{midgley_flow_2023}
\begin{bchapter}
\bauthor{\bsnm{Midgley}, \binits{L.I.}},
\bauthor{\bsnm{Stimper}, \binits{V.}},
\bauthor{\bsnm{Simm}, \binits{G.N.C.}},
\bauthor{\bsnm{Schölkopf}, \binits{B.}},
\bauthor{\bsnm{Hernandez-Lobato}, \binits{J.M.}}:
\bctitle{Flow {Annealed} {Importance} {Sampling} {Bootstrap}}.
In: \bbtitle{The Eleventh International Conference on Learning Representations},
\bconflocation{Kigali, Rwanda}
(\byear{2023})
\end{bchapter}
\endbibitem

\bibitem[\protect\citeauthoryear{Magnusson et~al.}{2024}]{magnusson_posteriordb_2024}
\begin{botherref}
\oauthor{\bsnm{Magnusson}, \binits{M.}},
\oauthor{\bsnm{Torgander}, \binits{J.}},
\oauthor{\bsnm{Bürkner}, \binits{P.-C.}},
\oauthor{\bsnm{Zhang}, \binits{L.}},
\oauthor{\bsnm{Carpenter}, \binits{B.}},
\oauthor{\bsnm{Vehtari}, \binits{A.}}:
posteriordb: {Testing}, {Benchmarking} and {Developing} {Bayesian} {Inference} {Algorithms}.
arXiv.
arXiv:2407.04967
(2024)
\end{botherref}
\endbibitem

\bibitem[\protect\citeauthoryear{Paszke et~al.}{2019}]{paszke_pytorch_2019}
\begin{bchapter}
\bauthor{\bsnm{Paszke}, \binits{A.}},
\bauthor{\bsnm{Gross}, \binits{S.}},
\bauthor{\bsnm{Massa}, \binits{F.}},
\bauthor{\bsnm{Lerer}, \binits{A.}},
\bauthor{\bsnm{Bradbury}, \binits{J.}},
\bauthor{\bsnm{Chanan}, \binits{G.}},
\bauthor{\bsnm{Killeen}, \binits{T.}},
\bauthor{\bsnm{Lin}, \binits{Z.}},
\bauthor{\bsnm{Gimelshein}, \binits{N.}},
\bauthor{\bsnm{Antiga}, \binits{L.}},
\bauthor{\bsnm{Desmaison}, \binits{A.}},
\bauthor{\bsnm{Köpf}, \binits{A.}},
\bauthor{\bsnm{Yang}, \binits{E.}},
\bauthor{\bsnm{DeVito}, \binits{Z.}},
\bauthor{\bsnm{Raison}, \binits{M.}},
\bauthor{\bsnm{Tejani}, \binits{A.}},
\bauthor{\bsnm{Chilamkurthy}, \binits{S.}},
\bauthor{\bsnm{Steiner}, \binits{B.}},
\bauthor{\bsnm{Fang}, \binits{L.}},
\bauthor{\bsnm{Bai}, \binits{J.}},
\bauthor{\bsnm{Chintala}, \binits{S.}}:
\bctitle{{PyTorch}: an imperative style, high-performance deep learning library}.
In: \bbtitle{Advances in Neural Information Processing Systems},
vol. \bseriesno{32},
pp. \bfpage{8026}--\blpage{8037}.
\bpublisher{Curran Associates Inc.},
\blocation{Vancouver, Canada}
(\byear{2019})
\end{bchapter}
\endbibitem

\bibitem[\protect\citeauthoryear{Papamakarios et~al.}{2021}]{papamakarios_normalizing_2022}
\begin{barticle}
\bauthor{\bsnm{Papamakarios}, \binits{G.}},
\bauthor{\bsnm{Nalisnick}, \binits{E.}},
\bauthor{\bsnm{Rezende}, \binits{D.J.}},
\bauthor{\bsnm{Mohamed}, \binits{S.}},
\bauthor{\bsnm{Lakshminarayanan}, \binits{B.}}:
\batitle{Normalizing flows for probabilistic modeling and inference}.
\bjtitle{The Journal of Machine Learning Research}
\bvolume{22}(\bissue{57}),
\bfpage{1}--\blpage{64}
(\byear{2021})
\end{barticle}
\endbibitem

\bibitem[\protect\citeauthoryear{Papamakarios et~al.}{2017}]{papamakarios_masked_2017}
\begin{bchapter}
\bauthor{\bsnm{Papamakarios}, \binits{G.}},
\bauthor{\bsnm{Pavlakou}, \binits{T.}},
\bauthor{\bsnm{Murray}, \binits{I.}}:
\bctitle{Masked autoregressive flow for density estimation}.
In: \bbtitle{Advances in Neural Information Processing Systems},
vol. \bseriesno{30}.
\bpublisher{Curran Associates, Inc.},
\blocation{Long Beach, United States}
(\byear{2017})
\end{bchapter}
\endbibitem

\bibitem[\protect\citeauthoryear{Rezende and Mohamed}{2015}]{rezende_variational_2015}
\begin{bchapter}
\bauthor{\bsnm{Rezende}, \binits{D.}},
\bauthor{\bsnm{Mohamed}, \binits{S.}}:
\bctitle{Variational {Inference} with {Normalizing} {Flows}}.
In: \bbtitle{Proceedings of the 32nd {International} {Conference} on {Machine} {Learning}}.
\bsertitle{Proceedings of Machine Learning Research},
vol. \bseriesno{37},
pp. \bfpage{1530}--\blpage{1538}.
\bpublisher{PMLR},
\blocation{Lille, France}
(\byear{2015})
\end{bchapter}
\endbibitem

\bibitem[\protect\citeauthoryear{Schoenholz and Cubuk}{2021}]{schoenholz_jax_2021}
\begin{barticle}
\bauthor{\bsnm{Schoenholz}, \binits{S.S.}},
\bauthor{\bsnm{Cubuk}, \binits{E.D.}}:
\batitle{{JAX}, {M}.{D}. {A} framework for differentiable physics}.
\bjtitle{Journal of Statistical Mechanics: Theory and Experiment}
\bvolume{2021}(\bissue{12}),
\bfpage{124016}
(\byear{2021})
\doiurl{10.1088/1742-5468/ac3ae9}
\end{barticle}
\endbibitem

\bibitem[\protect\citeauthoryear{Sch{\"a}r et~al.}{2024}]{schar_parallel_2024}
\begin{bchapter}
\bauthor{\bsnm{Sch{\"a}r}, \binits{P.}},
\bauthor{\bsnm{Habeck}, \binits{M.}},
\bauthor{\bsnm{Rudolf}, \binits{D.}}:
\bctitle{Parallel affine transformation tuning of {Markov Chain Monte Carlo}}.
In: \bbtitle{Proceedings of the 41st {International} {Conference} on {Machine} {Learning}}.
\bsertitle{Proceedings of Machine Learning Research},
vol. \bseriesno{235},
pp. \bfpage{43571}--\blpage{43607}.
\bpublisher{PMLR},
\blocation{Vienna, Austria}
(\byear{2024})
\end{bchapter}
\endbibitem

\bibitem[\protect\citeauthoryear{Samsonov et~al.}{2022}]{samsonov_local-global_2024}
\begin{bchapter}
\bauthor{\bsnm{Samsonov}, \binits{S.}},
\bauthor{\bsnm{Lagutin}, \binits{E.}},
\bauthor{\bsnm{Gabri\'{e}}, \binits{M.}},
\bauthor{\bsnm{Durmus}, \binits{A.}},
\bauthor{\bsnm{Naumov}, \binits{A.}},
\bauthor{\bsnm{Moulines}, \binits{E.}}:
\bctitle{Local-global {MCMC} fkernels: the best of both worlds}.
In: \bbtitle{Advances in Neural Information Processing Systems},
vol. \bseriesno{35},
pp. \bfpage{5178}--\blpage{5193}.
\bpublisher{Curran Associates Inc.},
\blocation{New Orleans, United States}
(\byear{2022})
\end{bchapter}
\endbibitem

\bibitem[\protect\citeauthoryear{Salman et~al.}{2018}]{salman_deep_2018}
\begin{botherref}
\oauthor{\bsnm{Salman}, \binits{H.}},
\oauthor{\bsnm{Yadollahpour}, \binits{P.}},
\oauthor{\bsnm{Fletcher}, \binits{T.}},
\oauthor{\bsnm{Batmanghelich}, \binits{K.}}:
Deep {Diffeomorphic} {Normalizing} {Flows}.
arXiv.
arXiv:1810.03256
(2018).
\doiurl{10.48550/arXiv.1810.03256}
\end{botherref}
\endbibitem

\bibitem[\protect\citeauthoryear{Tabak and Turner}{2013}]{tabak_family_2013}
\begin{barticle}
\bauthor{\bsnm{Tabak}, \binits{E.G.}},
\bauthor{\bsnm{Turner}, \binits{C.V.}}:
\batitle{A {Family} of {Nonparametric} {Density} {Estimation} {Algorithms}}.
\bjtitle{Communications on Pure and Applied Mathematics}
\bvolume{66}(\bissue{2}),
\bfpage{145}--\blpage{164}
(\byear{2013})
\doiurl{10.1002/cpa.21423}
\end{barticle}
\endbibitem

\bibitem[\protect\citeauthoryear{Urbano et~al.}{2019}]{urbano_new_2019}
\begin{bchapter}
\bauthor{\bsnm{Urbano}, \binits{J.}},
\bauthor{\bsnm{Lima}, \binits{H.}},
\bauthor{\bsnm{Hanjalic}, \binits{A.}}:
\bctitle{A {New} {Perspective} on {Score} {Standardization}}.
In: \bbtitle{Proceedings of the 42nd {International} {ACM} {SIGIR} {Conference} on {Research} and {Development} in {Information} {Retrieval}},
pp. \bfpage{1061}--\blpage{1064}.
\bpublisher{Association for Computing Machinery},
\blocation{Paris, France}
(\byear{2019})
\end{bchapter}
\endbibitem

\bibitem[\protect\citeauthoryear{Vitter}{1985}]{vitter_random_1985}
\begin{barticle}
\bauthor{\bsnm{Vitter}, \binits{J.S.}}:
\batitle{Random sampling with a reservoir}.
\bjtitle{ACM Transactions on Mathematical Software}
\bvolume{11}(\bissue{1}),
\bfpage{37}--\blpage{57}
(\byear{1985})
\doiurl{10.1145/3147.3165}
\end{barticle}
\endbibitem

\bibitem[\protect\citeauthoryear{Wu et~al.}{2020}]{wu_stochastic_2020}
\begin{bchapter}
\bauthor{\bsnm{Wu}, \binits{H.}},
\bauthor{\bsnm{Köhler}, \binits{J.}},
\bauthor{\bsnm{Noe}, \binits{F.}}:
\bctitle{Stochastic {Normalizing} {Flows}}.
In: \bbtitle{Advances in {Neural} {Information} {Processing} {Systems}},
vol. \bseriesno{33},
pp. \bfpage{5933}--\blpage{5944}.
\bpublisher{Curran Associates, Inc.},
\blocation{Vancouver, Canada (virtual conference)}
(\byear{2020})
\end{bchapter}
\endbibitem

\bibitem[\protect\citeauthoryear{Williams et~al.}{2021}]{williams_nested_2021}
\begin{barticle}
\bauthor{\bsnm{Williams}, \binits{M.J.}},
\bauthor{\bsnm{Veitch}, \binits{J.}},
\bauthor{\bsnm{Messenger}, \binits{C.}}:
\batitle{Nested {Sampling} with {Normalising} {Flows} for {Gravitational}-{Wave} {Inference}}.
\bjtitle{Physical Review D}
\bvolume{103}(\bissue{10}),
\bfpage{103006}
(\byear{2021})
\doiurl{10.1103/PhysRevD.103.103006}
\end{barticle}
\endbibitem

\end{thebibliography}

\newpage
\begin{appendices}

\section{Additional results}\label{app:additional-results}
In this section, we discuss additional results regarding NF operation speeds, autoregressive NF components, and experiments with a smaller time budget.
\subsection{NF operation speed for moderate dimensional targets}
\begin{figure}[t]
    \centering
    \includegraphics[width=\linewidth]{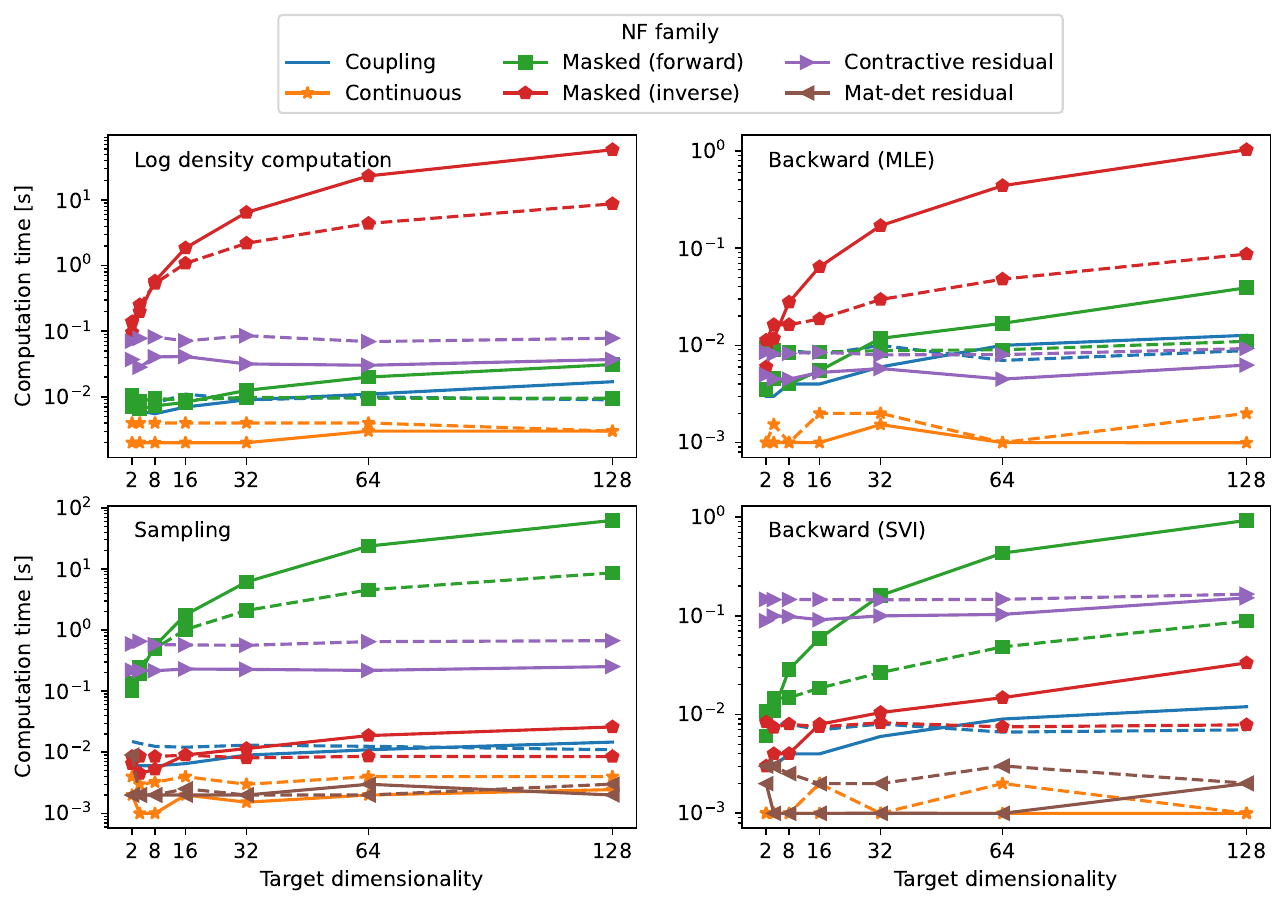}
    \caption{Computation time in seconds for density evaluation, sampling, and model parameter gradient computation via automatic differentiation (backward). Solid lines denote operations on the CPU, and dashed lines are operations on the GPU. Shown values are medians across operation times for NFs belonging to corresponding families. Operation times are averages of 100 trials with a hundred standard Gaussian vectors in 100D for each operation.}
    \label{fig:cpu-vs-gpu}
\end{figure}

Executing NF operations on the GPU can be much faster than the CPU when target dimensionality is high, such as for distributions of images.
Such NFs consist of convolutional neural networks whose operations can be efficiently parallelized on the GPU.
However, many analyses are performed on statistical models with fewer parameters and often on consumer-grade laptops.
In such cases, it is practical to know whether GPUs are necessary for efficient NF operations or if CPUs suffice.
In Figure~\ref{fig:cpu-vs-gpu}, we compare the efficiency of GPU and CPU for three essential NF operations: computing the log probability of data points, sampling new data points, and computing the gradient of the loss.
We use an AMD Ryzen 9 3900X 12-core CPU at 3.8 GHz and an NVIDIA RTX 2080S GPU in these experiments.

We find that MA and IA architectures are faster on the GPU for all operations.
Out of the three autoregressive families, coupling NFs are the fastest on the CPU.
Moreover, coupling NFs are faster on the CPU than the GPU for 64 or fewer dimensions.
In continuous NFs and both residual NF families, we find that GPU operations are generally slower than CPU operations.
Our results have practical relevance to the further development of general-purpose MCMC packages, as we find that the CPU is sufficient for key NF operations.
This means practitioners and researchers can leverage NFMC for analyses with minimal hardware.
Our findings are also promising for embedded systems without GPU support.
We note that we perform autodifferentiation in an eager execution framework, and further work is needed to evaluate gradient computation speeds in graph execution configurations.

\subsection{Autoregressive conditioner and transformer comparison in NeuTra MCMC}
We compare different combinations of autoregressive conditioners (MADE and coupling) and transformers (affine maps, splines, and neural networks) in terms of moment estimation quality.
We focused our comparison to \neutramcmc sampling for a fair comparison, as we only used MADE conditioners in this context, not Jump MCMC.
We show the results in Table~\ref{tab:autoregressive-nf-comparison}.

\begin{table}

            \renewrobustcmd{\bfseries}{\fontseries{b}\selectfont}
            \renewrobustcmd{\boldmath}{}
            \sisetup{%
                table-align-uncertainty=true,
                detect-all,
                separate-uncertainty=true,
                mode=text,
                round-mode=uncertainty,
                round-precision=2,
                table-format = 2.2(2),
            }
            \tablefontsize
\begin{tabular}{l
S
S
S
S
S}
\toprule
{Combination} & {Gaussian} & {Non-Gaussian} & {Multimodal} & {Real-world} & {All} \\
\midrule
{C-Affine} & -0.44(0.28) & 0.88 & 0.59(0.17) & \bfseries -0.37(0.32) & 0.46(0.23) \\
{C-Spline} & \bfseries -0.59(0.56) & -0.59(0.88) & -0.44(0.65) & -0.15(0.4) & \bfseries -0.49(0.18) \\
{C-NN} & 1.17(0.29) & \bfseries -0.88 & 0.88(0.41) & 0.37(0.37) & 0.29(0.26) \\
\midrule
{MADE-Affine} & 0.44(0.44) & 1.46 & -0.15(0.5) & 0.22(0.46) & 0.52(0.2) \\
{MADE-Spline} & \bfseries -0.73(0.37) & 0.0(0.29) & \bfseries -0.59(0.38) & 0.15(0.24) & -0.23(0.22) \\
{MADE-NN} & 0.15(0.55) & \bfseries -0.88(0.59) & -0.29(0.63) & -0.22(0.36) & \bfseries -0.55(0.22) \\
\bottomrule
\end{tabular}
\caption{$\overline{r} \pm \hat{\sigma}$ for all conditioner-transformer combinations in autoregressive NFs, estimated with default hyperparameters for each benchmark. NN denotes neural network transformers, and C denotes coupling conditioners. The top 20\% combinations are shown in bold. Ranks are computed separately for each target family.}
\label{tab:autoregressive-nf-comparison}
\end{table}

The combination of MADE conditioners and NN transformers attains the best $\overline{r}$ across all targets, with C-spline models ranking second.
Interestingly, C-spline models rank better than C-NN models, and MADE-NN spline models are better than MADE-spline models.
This suggests that autoregressive NF performance in \neutramcmc should be assessed by jointly observing the conditioner and transformer, as certain combinations like MADE-NN and C-spline may possess better inductive biases than others.

Nevertheless, MADE models seem to rank somewhat better than models with coupling conditioners, which is consistent with our previous findings in Tables~\ref{tab:nf-per-family-neutra} and~\ref{tab:target-variations-neutra}.
There are some caveats to this conclusion when observing individual target families:
\begin{itemize}
    \item On non-Gaussian targets, MADE-NN achieves the best $\overline{r}$, but has a very high uncertainty. Its performance is matched by C-NN, which always ranks the same on the non-Gaussian target family. The true performance of MADE is thus inconclusive. However, we notice that NN transformers are the best in both cases.
    \item On multimodal targets, all MADE configurations achieve $\overline{r} < 0$, which indicates good performance. C-spline models could also be promising candidates, however their uncertainty is high.
    \item On real-world targets, only MADE-NN achieves $\overline{r} < 0$ and is outperformed by C-affine models. We again note the high uncertainty which prevents us from making definitive conclusions.
\end{itemize}

We further compare the two conditioners by observing the percentage of \neutramcmc experiments with identical configurations except for the conditioner.
Configurations include the choice of sampler (\neutrahmc or \neutramh), transformer (as defined in Section~\ref{subsec:nf-architectures}), target distribution, and NF hyperparameters.
We find that MADE conditioners perform somewhat better, beating coupling conditioners 61\% of the time when using affine transformers, 52\% with splines, 56\% with NN transformers, and 56\% of the time across all transformers.

\subsection{Experiments with a small time budget}\label{app:subsec:small-time-budget}
We investigate sampling quality when substantially reducing both the allotted warm-up and sampling time.
These experiments are indicative of short-run NFMC performance and can provide useful guidelines when adopting NFMC into mainstream programming packages for sampling, as many of their users typically only deal with quick analyses.
We limit warm-up time to 2 minutes and sampling time to 5 minutes.

\subsubsection{Sampler comparison on short NFMC runs}

\begin{figure}[ht]
    \begin{subfigure}[t]{0.525\textwidth}
        \centering
        \includegraphics[width=\textwidth, trim={0 0 266 0}, clip]{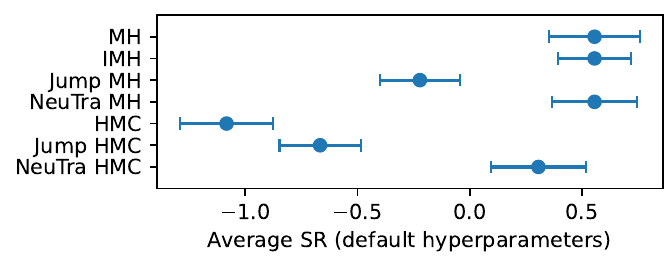}
        \caption{$\overline{r} \pm \hat{\sigma}$ across all targets and NFs for each sampler, using $b^2$ limited to experiments with default NF hyperparameters.}
        \label{subfig:stb-nfmc-all-benchmarks-left}
    \end{subfigure}
    \hfill
    \begin{subfigure}[t]{0.428\textwidth}
        \centering
        \includegraphics[width=\textwidth, trim={325 0 0 0}, clip]{stb-nfmc-all-benchmarks-stb.pdf}
        \caption{$\overline{r} \pm \hat{\sigma}$ across all targets and NFs for each sampler, values estimated with minimum $b^2$ across all NF hyperparameter sets.}
        \label{subfig:stb-nfmc-all-benchmarks-right}
    \end{subfigure}
    \caption{Numerical comparison of investigated NFMC methods on the entire benchmark. Each experiment consisted of two minutes of warm-up and five minutes of sampling.}
\end{figure}

We present SR values for different samplers in Figures~\ref{subfig:stb-nfmc-all-benchmarks-left} and~\ref{subfig:stb-nfmc-all-benchmarks-right}.
We observe some differences in comparison to Figures~\ref{subfig:nfmc-all-benchmarks-left} and~\ref{subfig:nfmc-all-benchmarks-right}.

HMC expectedly performs better than Jump HMC when using default NF hyperparameters, compared to matching the performance of Jump HMC performance in runs with longer warm-up and sampling stages.
Interestingly, the opposite holds when observing minimum $b^2$.
This suggests that choosing an appropriate NF architecture for sampling can improve moment estimates even with limited computational time.
A major practical consequence is that we may perform several short runs of NFMC to gauge the potential of different hyperparameter configurations. This reduces the number of repeated NFMC runs with poor hyperparameter choices, which is particularly promising for experiments with long chains or computationally expensive target density evaluations.
An extensive hyperparameter search may not be a priority for all analyses. Our results encourage the development of hyperparameter tuning methods for such cases.

When using default NF hyperparameters, IMH and \neutramh are comparable to MH, while Jump MH remains the best of the investigated gradient-free samplers.
These results suggest that gradient-free sampling can greatly benefit from independent NF jumps even without hyperparameter tuning, which is especially useful for quick tests.
IMH and Jump MH are comparable when considering minimum $b^2$. As long as the NF architecture and hyperparameters are chosen well, this suggests that independent NF jumps can be more effective than local MH exploration with minimal hyperparameter tuning effort.

We further investigate sampler performance for different target families in Table~\ref{tab:nfmc-per-family-stb}.
Jump HMC is again the decisive winner, just as in longer NFMC runs (see Table~\ref{tab:nfmc-per-family}).
Jump MH narrowly remains the best gradient-free sampler overall, only beaten by IMH on real-world targets.
The main difference compared to longer NFMC runs is that MH and HMC rank noticeably better on short runs.
This is reasonable, as longer runs allow better NF fits.
Moreover, there is greater uncertainty in SR for non-Gaussian targets.
Jump MH also achieves a worse rank on non-Gaussian targets compared to longer runs.
The takeaway is that short-run NFMC is not necessarily suitable for challenging non-Gaussian targets, especially if selecting a suitable NF is very time-consuming.

\begin{table}

            \renewrobustcmd{\bfseries}{\fontseries{b}\selectfont}
            \renewrobustcmd{\boldmath}{}
            \sisetup{%
                table-align-uncertainty=true,
                detect-all,
                separate-uncertainty=true,
                mode=text,
                round-mode=uncertainty,
                round-precision=2,
                table-format = 2.2(2),
            }
            \tablefontsize
\begin{tabular}{l
S
S
S
S
S}
\toprule
{Sampler} & {Gaussian} & {Non-Gaussian} & {Multimodal} & {Real-world} & {All} \\
\midrule
{MH} & 1.00(0.20) & 0.75(0.75) & 0.75(0.32) & 1.31(0.13) & 1.06(0.13) \\
{IMH} & 0.62(0.52) & 0.50(1.00) & 0.12(0.24) & -0.56(0.29) & -0.03(0.22) \\
{Jump MH} & -0.25(0.32) & -0.50(0.50) & -0.38(0.12) & 0.31(0.31) & -0.06(0.18) \\
{NeuTra MH} & 0.88(0.31) & 1.0 & 0.75(0.60) & 0.25(0.23) & 0.58(0.18) \\
{HMC} & -1.12(0.12) & 0.00(0.50) & -0.38(0.62) & -0.44(0.35) & -0.53(0.22) \\
{Jump HMC} & \bfseries -1.25(0.25) & \bfseries -1.5 & \bfseries -0.88(0.62) & \bfseries -0.69(0.33) & \bfseries -0.94(0.21) \\
{NeuTra HMC} & 0.12(0.24) & -0.25(0.75) & 0.00(0.54) & -0.19(0.33) & -0.08(0.19) \\
\bottomrule
\end{tabular}
\caption{$\overline{r} \pm \hat{\sigma}$ for all samplers and target families given 2 minutes of warm-up time and 5 minutes of sampling time. Samplers with the best $\overline{r}$ are shown in bold for each target family. We estimate $\overline{r} \pm \hat{\sigma}$ with the minimum $b^2$ across all NFs for each target within a family. Entries without $\hat{\sigma}$ always attain the same $\overline{r}$. Ranks are computed separately for each target family.}
\label{tab:nfmc-per-family-stb}
\end{table}

\subsubsection{NF comparison on short NFMC runs}

We compare different NF architectures on short Jump MCMC runs in Table~\ref{tab:nf-per-family-stb}.
We find \cnfrkreg to be in the top 20\% for all target families, including Gaussian and real-world targets, where it performed worse during long runs.
\cnfeuler and \cnfrk NFs attain $\overline{r} > 0$ on Gaussians, which is similar to the long run results in Table~\ref{tab:nf-per-family}, where we only observed $\overline{r} > 0$ on Gaussians with \cnfeuler and \cnfrk.
However, the combined ranks suggest that continuous NF models are consistently among the best choices regardless of the allotted computational time.
This implies that continuous NF models are among the quickest to efficiently train with few training samples from NFMC.

The main difference regarding autoregressive NFs is the good performance of C-LR-NSF, which ranks second best among all NFs.
LRS transformers have more parameters than affine maps in NICE and \realnvp, yet fewer than RQS and NAF transformers.
This suggests that the LRS capacity is most beneficial for short runs of Jump MCMC.
We also find residual NFs to perform worse on short runs, suggesting that their applicability is somewhat limited.

\begin{table}

            \renewrobustcmd{\bfseries}{\fontseries{b}\selectfont}
            \renewrobustcmd{\boldmath}{}
            \sisetup{%
                table-align-uncertainty=true,
                detect-all,
                separate-uncertainty=true,
                mode=text,
                round-mode=uncertainty,
                round-precision=2,
                table-format = 2.2(2),
            }
            \tablefontsize
\begin{tabular}{l
S
S
S
S
S}
\toprule
{NF} & {Gaussian} & {Non-Gaussian} & {Multimodal} & {Real-world} & {All} \\
\midrule
{NICE} & \bfseries -0.87(0.34) & 0.14(0.87) & -0.22(0.45) & 0.29(0.35) & -0.10(0.23) \\
{\realnvp} & -0.22(0.14) & -0.29(0.72) & -0.22(0.38) & -0.33(0.37) & -0.27(0.19) \\
{C-LR-NSF} & -0.58(0.28) & 0.00(0.14) & \bfseries -0.58(0.60) & \bfseries -0.58(0.20) & \bfseries -0.51(0.16) \\
{C-RQ-NSF} & \bfseries -1.16(0.34) & 1.30(0.29) & -0.29(0.34) & 0.36(0.42) & -0.02(0.27) \\
{C-\nafdeep} & 0.29(0.34) & 0.00(1.30) & -0.07(0.62) & 0.80(0.26) & 0.40(0.23) \\
{C-\nafdense} & 1.3 & 0.72(0.58) & 0.87(0.43) & 0.87(0.20) & 0.95(0.14) \\
{C-\nafboth} & 1.59 & 0.43(1.16) & 1.59 & 0.80(0.36) & 1.11(0.21) \\
\midrule
{i-ResNet} & -0.43(0.20) & 0.00(0.43) & 0.07(0.36) & -0.47(0.27) & -0.29(0.16) \\
{ResFlow} & 0.29(0.36) & 0.29(0.43) & 0.14(0.35) & \bfseries -0.58(0.23) & -0.13(0.17) \\
\midrule
{\cnfeuler} & 0.80(0.14) & \bfseries -1.45(0.14) & \bfseries -0.51(0.49) & -0.47(0.32) & \bfseries -0.31(0.23) \\
{\cnfrk} & 0.07(0.46) & \bfseries -0.43(1.16) & 0.00(0.51) & 0.04(0.37) & -0.02(0.23) \\
{\cnfrkreg} & \bfseries -1.09(0.42) & \bfseries -0.72(0.29) & \bfseries -0.80(0.52) & \bfseries -0.72(0.28) & \bfseries -0.82(0.18) \\
\bottomrule
\end{tabular}
\caption{$\overline{r} \pm \hat{\sigma}$ for all NFs and target families in IMH, Jump MH, and Jump HMC given 2 minutes of warm-up time and 5 minutes of sampling time. NFs in the top 20th percentile are shown in bold for each target family. We estimate $\overline{r} \pm \hat{\sigma}$ with $b^2$ from runs with default hyperparameters. Entries without $\hat{\sigma}$ always attain the same $\overline{r}$. Ranks are computed separately for each target family.}
\label{tab:nf-per-family-stb}
\end{table}

We compare NFs for short \neutramcmc runs in Table~\ref{tab:nf-per-family-neutra-stb}.
The radial flow shows the most striking change in performance.
Whereas it decisively ranked best on long NFMC runs (c.f.\ Table~\ref{tab:nf-per-family-neutra}), it is among the worst here and clearly the worst choice for Gaussian and synthetic non-Gaussian targets.
When considering the entire benchmark, all coupling NFs rank better than in long runs.
Their relative ranks remain similar on Gaussian and synthetic non-Gaussian targets but mostly change on multimodal and real-world targets.
It is difficult to draw conclusions for the latter two families due to the high uncertainty.
We find that C-\nafdense and C-\nafboth attain $\overline{r} < 0$ on all families, which contributes to them ranking well on the entire benchmark.
The overall ranks of IA methods are largely the same as in long runs, except for worse rankings attained by IA-\nafdense and IA-\nafboth.
Both residual architectures rank worse than in long runs.
We also note that the performance of CNF models and contractive residual NFs aligns with previous long-run experiments: contractive residual NFs perform well on Jump MCMC and NeuTra MCMC in both long and short runs, while CNF again performs well on Jump MCMC and again not as well in NeuTra MCMC.

\begin{table}[t]

            \renewrobustcmd{\bfseries}{\fontseries{b}\selectfont}
            \renewrobustcmd{\boldmath}{}
            \sisetup{%
                table-align-uncertainty=true,
                detect-all,
                separate-uncertainty=true,
                mode=text,
                round-mode=uncertainty,
                round-precision=2,
                table-format = 2.2(2),
            }
            \tablefontsize
\begin{tabular}{l
S
S
S
S
S}
\toprule
{NF} & {Gaussian} & {Non-Gaussian} & {Multimodal} & {Real-world} & {All} \\
\midrule
{NICE} & \bfseries -0.70(0.41) & 0.74(0.08) & \bfseries -0.58(0.46) & -0.23(0.31) & -0.30(0.21) \\
{\realnvp} & -0.50(0.40) & 0.74(0.41) & 0.41(0.27) & -0.15(0.33) & -0.00(0.20) \\
{C-LR-NSF} & \bfseries -1.03(0.24) & 0.50(0.17) & -0.08(0.47) & -0.20(0.38) & -0.28(0.22) \\
{C-RQ-NSF} & \bfseries -0.58(0.27) & \bfseries -0.91(0.74) & \bfseries -0.54(0.53) & 0.36(0.31) & -0.22(0.22) \\
{C-\nafdeep} & 0.58(0.26) & 0.00(0.17) & 0.74(0.46) & 0.51(0.32) & 0.52(0.18) \\
{C-\nafdense} & -0.37(0.61) & \bfseries -1.32(0.17) & -0.08(0.60) & \bfseries -0.56(0.36) & \bfseries -0.50(0.24) \\
{C-\nafboth} & -0.41(0.64) & \bfseries -1.32(0.17) & -0.45(0.57) & -0.28(0.38) & \bfseries -0.47(0.25) \\
\midrule
{IAF} & -0.50(0.36) & 1.16(0.17) & -0.29(0.37) & 0.51(0.32) & 0.18(0.22) \\
{IA-LR-NSF} & -0.50(0.39) & 0.25(0.25) & -0.12(0.37) & -0.19(0.21) & -0.19(0.15) \\
{IA-RQ-NSF} & \bfseries -0.83(0.26) & -0.08(0.25) & -0.21(0.54) & -0.44(0.28) & \bfseries -0.43(0.18) \\
{IA-\nafdeep} & -0.41(0.49) & -0.5 & -0.41(0.14) & \bfseries -0.59(0.35) & \bfseries -0.50(0.18) \\
{IA-\nafdense} & 0.12(0.55) & 0.00(0.99) & 0.74(0.43) & 0.75(0.22) & 0.53(0.20) \\
{IA-\nafboth} & 1.16(0.07) & -0.17(0.66) & 1.45(0.04) & 0.74(0.37) & 0.90(0.20) \\
\midrule
{i-ResNet} & -0.12(0.30) & -0.50(0.17) & -0.12(0.54) & \bfseries -0.52(0.29) & -0.34(0.18) \\
{ResFlow} & -0.50(0.43) & \bfseries -0.74(0.08) & 0.08(0.37) & -0.39(0.34) & -0.35(0.19) \\
\midrule
{Planar} & 0.04(0.26) & 0.41(0.41) & 0.12(0.35) & \bfseries -0.54(0.28) & -0.16(0.17) \\
{Radial} & 1.65 & 1.65 & 0.83(0.83) & -0.03(0.46) & 0.72(0.31) \\
{Sylvester} & -0.41(0.21) & 0.08(1.40) & \bfseries -0.58(0.14) & -0.22(0.22) & -0.34(0.19) \\
\midrule
{\cnfeuler} & 1.49 & 0.25(1.24) & \bfseries -0.62(0.71) & 0.90(0.35) & 0.62(0.29) \\
{\cnfrk} & 1.28(0.04) & 0.00(1.32) & -0.45(0.66) & 0.60(0.43) & 0.45(0.29) \\
{\cnfrkreg} & 0.54(0.41) & -0.25(1.40) & 0.17(0.54) & -0.08(0.36) & 0.09(0.24) \\
\bottomrule
\end{tabular}
\caption{$\overline{r} \pm \hat{\sigma}$ for all NFs and target families in NeuTra MH and NeuTra HMC given 2 minutes of warm-up time and 5 minutes of sampling time. NFs in the top 20th percentile are shown in bold for each target family. We estimate $\overline{r} \pm \hat{\sigma}$ with $b^2$ from runs with default hyperparameters. Entries without $\hat{\sigma}$ always attain the same $\overline{r}$. Ranks are computed separately for each target family.}
\label{tab:nf-per-family-neutra-stb}
\end{table}

\subsubsection{Short summary of findings}
One of our key findings is that NFMC can still perform well even without long warm-up and tuning stages.
For gradient-free sampling, we found Jump MH to perform better than all other gradient-free samplers just by using default NF hyperparameters.
This makes it a good primary choice of sampler for short sampling runs.
For gradient-based sampling, we found HMC to perform best compared to NFMC with default hyperparameters.
If we choose good hyperparameters, Jump HMC will rank best among gradient-based samplers.
The best architectures for it are \cnfrkreg, C-LR-NSF, and \cnfeuler.
In many cases, we can afford some NF hyperparameter tuning time, which makes Jump HMC a suitable choice for gradient-based sampling.

\subsection{Verifying results via kernelized Stein discrepancy}\label{app:subsec:ksd}
The squared bias of the second moment is a common evaluation metric for MCMC and can be related to the well-known mean squared error (see Appendix~\ref{app:sec:comparison-metrics}).
In certain cases, measuring second moment error via $b^2$ is not sufficient to evaluate the quality of MCMC samples.
For example, if $X$ is a unidimensional target distribution, $Y = X$ is the perfect model, and $Z = N(0, \mathrm{Var}[X])$ is a Gaussian approximation, then $b^2 = 0$ for both $Y$ and $Z$.
Our a priori position is that such cases are unlikely to happen because the Metropolis-Hastings accept/reject step intuitively ensures that visited states at least approximately follow the geometry of the target distribution.
However, there may exist other pathologies that could be better handled with a different metric.
Furthermore, we may be interested in other properties of the MCMC sample distribution besides second-moment estimation.

An alternative metric that evaluates sample quality is Kernelized Stein discrepancy~\citep[KSD]{liu_kernelized_2016}.
KSD measures how far a given probability distribution $Y$ is from a target distribution $X$, with the density of the latter known up to a normalization constant.
While $b^2$ focuses on second moment estimates, KSD compares the distributions in a global manner, being sensitive to mean, variance, skewness, and other distribution properties.
Let $\mathcal{F}$ be a set of smooth functions $f$ that satisfy
\begin{align}
    \mathbb{E}_{X}[s_Y(x)f(x)+\nabla_x f(x)] = 0,
\end{align}
where $s_Y(x) = \nabla_x \log p_Y(x)$.
Then~\cite{liu_kernelized_2016} define Stein discrepancy as:
\begin{align}
    \mathbb{S}(X, Y) = \max_{f \in \mathcal{F}}( \mathbb{E}_X[s_Y(x)f(x) + \nabla_x f(x)] )^2
\end{align}
with $\mathbb{S}(X, Y) > 0$ whenever $X \neq Y$.
This quantity is often intractable as it requires difficult variational optimization. They instead propose KSD, a kernelized variant of Stein discrepancy, which can be reformulated to only require the target score function $s_Y(x)$, samples from $X$, and a kernel $k(\cdot, \cdot)$:
\begin{align}
    \mathbb{S}(X, Y) &= \mathbb{E}_{x,x^\prime\sim X}[u_Y(x, x^\prime)], \\
    u_Y(x, x^\prime) &= s_Y(x)^\top k(x,x^\prime) s_Y(x^\prime) + s_Y(x)^\top \nabla_{x^\prime} k(x, x^\prime) \\
    &+ \nabla_{x} k(x, x^\prime)^\top s_Y(x^\prime) + \trace{\nabla_{x, x^\prime}k(x,x^\prime)}. \label{eqn:ksd}
\end{align}
Here, $k$ is a kernel in the Stein class of $X$.
The gradient and trace terms can be computed efficiently when $k$ is a radial basis function (RBF) kernel, which yields a tractable version of KSD for some $\sigma > 0$:
\begin{align}
    k(x, x^\prime) = \exp\left(-||x - x^\prime||^2 / (2\sigma^2) \right).
\end{align}

\begin{figure}[ht]
    \begin{subfigure}[t]{0.525\textwidth}
        \centering
        \includegraphics[width=\textwidth, trim={0 0 266 0}, clip]{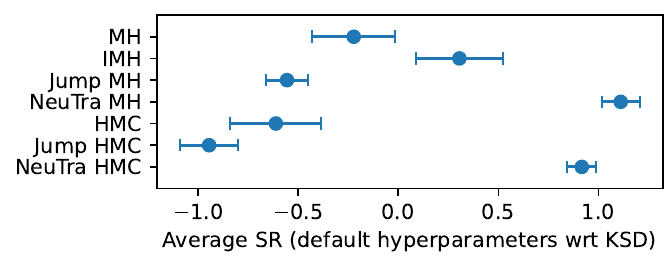}
        \caption{$\overline{r} \pm \hat{\sigma}$ across all targets and NFs for each sampler, using KSD limited to experiments with default NF hyperparameters.}
        \label{subfig:stb-nfmc-all-benchmarks-left-ksd}
    \end{subfigure}
    \hfill
    \begin{subfigure}[t]{0.428\textwidth}
        \centering
        \includegraphics[width=\textwidth, trim={325 0 0 0}, clip]{stb-nfmc-all-benchmarks-ksd.pdf}
        \caption{$\overline{r} \pm \hat{\sigma}$ across all targets and NFs for each sampler, values estimated with minimum KSD across all NF hyperparameter sets.}
        \label{subfig:stb-nfmc-all-benchmarks-right-ksd}
    \end{subfigure}
    \caption{Numerical comparison of investigated NFMC methods on the entire benchmark according to KSD. Each experiment consisted of two minutes of warm-up and five minutes of sampling.}
\end{figure}

We repeated the experiments in Section~\ref{app:subsec:small-time-budget} and observed SR according to KSD instead of $b^2$, treating $X$ as the empirical distribution of MCMC draws and $Y$ as the target distribution in Equation~\ref{eqn:ksd}.
We limited the number of samples to $n = 1000$ as computations involving the RBF kernel involve a costly computation of an $n \times n$ matrix of distances between samples.\footnote{We checked the accuracy in terms of $n$ on an example with $X = N(0, I)$ and $Y = N(0, 3.5^2I)$. KSD between $X$ and the empirical distribution of iid samples from $X$ was equal to $0.04$ and $0.02$ with $n = 1000$ and $n = 10000$ draws from $X$, respectively. When iid samples were drawn from $Y$, the corresponding KSD values were 1.60 and 1.63, which means a significant difference in both cases.}
We used reservoir sampling~\citep{vitter_random_1985} to select the samples on-the-fly with a fixed memory budget.
We chose $\sigma$ to be the median of all sample distances, following the median bandwidth heuristic convention.
We show sampler comparison results in Figures~\ref{subfig:stb-nfmc-all-benchmarks-left-ksd} and~\ref{subfig:stb-nfmc-all-benchmarks-right-ksd}.

Our findings are largely consistent with $b^2$ experiments.
When tuning NF hyperparameters, Jump HMC remains the best sampler, NeuTra HMC performs worse than HMC.
Moreover, Jump HMC attains a better SR value than HMC when using the default hyperparameter set.
This is in contrast to $b^2$ experiments, where the roles were reversed.
It suggests that, while Jump HMC performs worse in second-moment estimation, it manages to capture the overall distribution better.
We also observe that IMH with the default hyperparameter set performs worse than regular MH.
This adds to the results in Figure~\ref{subfig:stb-nfmc-all-benchmarks-left}: while there is little difference in second moment error, IMH alone cannot adequately describe the target distribution globally.
Introducing local MCMC transitions via Jump MH alleviates this problem and ranks better than both MH and IMH.
Both MH and IMH achieve a similar SR value using tuned hyperparameters, which is consistent with the result in Figure~\ref{subfig:stb-nfmc-all-benchmarks-right}.
They also outrank HMC in this setting, adding further evidence that global NF proposals are an efficient exploration method.
This experiment also reveals that while NeuTra MH can rank better than MH in second moment estimation, it ranks worse according to KSD.

We also ranked NF architectures according to KSD on short-run experiments.
We show results for Jump MCMC and IMH in Table~\ref{tab:nf-per-family-stb-ksd}.
We observe similarities with $b^2$ rankings on Jump MCMC and IMH in Table~\ref{tab:nf-per-family-stb}:
\begin{itemize}
    \item Autoregressive NFs with simple transformers consistently rank better than their NAF counterparts.
    \item CNF models rank the best overall.
\end{itemize}
We note that residual NFs rank worse relative to comparisons with $b^2$, implying that they do not capture target distribution characteristics globally, even though they attain better performance than competing NFs on second moment estimation.
\begin{table}

            \renewrobustcmd{\bfseries}{\fontseries{b}\selectfont}
            \renewrobustcmd{\boldmath}{}
            \sisetup{%
                table-align-uncertainty=true,
                detect-all,
                separate-uncertainty=true,
                mode=text,
                round-mode=uncertainty,
                round-precision=2,
                table-format = 2.2(2),
                table-column-width=1.7cm
            }
            
\begin{tabular}{l
S
S
S
S
S}
\toprule
{NF} & {Gaussian} & {Non-Gaussian} & {Multimodal} & {Real-world} & {All} \\
\midrule
{NICE} & -0.51(0.22) & \bfseries -1.01(0.29) & -0.29(0.25) & -0.04(0.38) & -0.31(0.19) \\
{Real NVP} & -0.72(0.12) & -0.58(0.43) & \bfseries -1.09(0.51) & 0.54(0.33) & -0.23(0.25) \\
{C-LR-NSF} & -0.36(0.25) & -0.58(0.14) & -0.14(0.31) & -0.11(0.36) & -0.23(0.18) \\
{C-RQ-NSF} & \bfseries -1.01(0.49) & \bfseries -1.30(0.29) & -0.29(0.54) & -0.00(0.38) & \bfseries -0.43(0.25) \\
{C-NAF$_\mathrm{deep}$} & 0.51(0.07) & 0.58(0.14) & 0.65(0.22) & 0.07(0.38) & 0.35(0.18) \\
{C-NAF$_\mathrm{dense}$} & 1.01(0.12) & 1.16(0.14) & 0.51(0.55) & 0.91(0.28) & 0.87(0.17) \\
{C-NAF$_\mathrm{both}$} & 1.23(0.14) & 1.45(0.14) & 0.36(0.58) & 0.29(0.20) & 0.64(0.19) \\
\midrule
{i-ResNet} & 0.58(0.36) & 0.87(0.14) & 0.51(0.56) & -0.14(0.41) & 0.27(0.24) \\
{ResFlow} & 1.52(0.07) & 1.01(0.58) & 1.01(0.49) & 0.80(0.30) & 1.03(0.18) \\
\midrule
{CNF$_\mathrm{Euler}$} & \bfseries -1.01(0.29) & -0.58(0.72) & \bfseries -0.58(0.65) & \bfseries -0.87(0.21) & \bfseries -0.80(0.18) \\
{CNF$_\mathrm{RK}$} & -0.00(0.14) & -0.14(0.29) & -0.00(0.36) & \bfseries -0.83(0.22) & -0.39(0.16) \\
{CNF$_\mathrm{RK(R)}$} & \bfseries -1.23(0.22) & \bfseries -0.87(0.72) & \bfseries -0.65(0.30) & \bfseries -0.62(0.22) & \bfseries -0.79(0.15) \\
\bottomrule
\end{tabular}
\caption{$\overline{r} \pm \hat{\sigma}$ for all NFs and target families in IMH, Jump MH, and Jump HMC given 2 minutes of warm-up time and 5 minutes of sampling time. NFs in the top 20th percentile are shown in bold for each target family. We estimate $\overline{r} \pm \hat{\sigma}$ with KSD from runs with default hyperparameters. Entries without $\hat{\sigma}$ always attain the same $\overline{r}$. Ranks computed separately for each target family.}
\label{tab:nf-per-family-stb-ksd}
\end{table}
We show results for NeuTra MCMC in Table~\ref{tab:nf-per-family-neutra}.
Contractive residual NFs rank better compared to other NFs than in $b^2$ experiments.
Conversely, CNF models are the best in this scenario, both overall and for almost every target distribution family.
Autoregressive NFs attain $\overline{r}$ between -0.16 and 0.44, exhibiting less variance of average ranks than the $b^2$ experiments.
Moreover, while certain autoregressive architectures rank best when observing $b^2$, they are overall worse when considering the global structure of MCMC draws.

In summary, the ranks of samplers are consistent regardless of the choice of $b^2$ or KSD as the metric.
Architecture rankings are also consistent when observing Jump MCMC, with the exception of residual NFs that perform better within second-moment estimation than the global sample distribution view.
Finally, KSD clarifies the relative ranks of architecture families in NeuTra MCMC, ranking CNFs as the best, contractive residual NFs as second, then autoregressive NFs, and the remainder of matrix determinant residual NFs.

\begin{table}

            \renewrobustcmd{\bfseries}{\fontseries{b}\selectfont}
            \renewrobustcmd{\boldmath}{}
            \sisetup{%
                table-align-uncertainty=true,
                detect-all,
                separate-uncertainty=true,
                mode=text,
                round-mode=uncertainty,
                round-precision=2,
                table-format = 2.2(2),
                table-column-width=1.7cm
            }
            
\begin{tabular}{l
S
S
S
S
S}
\toprule
{NF} & {Gaussian} & {Non-Gaussian} & {Multimodal} & {Real-world} & {All} \\
\midrule
{NICE} & 0.62(0.41) & 0.90(0.74) & -0.01(0.45) & 0.48(0.39) & 0.44(0.22) \\
{Real NVP} & 0.08(0.31) & -0.13(0.96) & 0.14(0.63) & 0.39(0.42) & 0.18(0.24) \\
{C-LR-NSF} & 0.04(0.47) & 1.49 & -0.19(0.80) & 0.81(0.27) & 0.49(0.24) \\
{C-RQ-NSF} & -0.21(0.45) & 1.46(0.19) & -0.01(0.30) & 0.68(0.22) & 0.40(0.20) \\
{C-NAF$_\mathrm{deep}$} & -0.17(0.50) & -0.74(0.74) & 1.48(0.07) & 0.34(0.42) & 0.36(0.27) \\
{C-NAF$_\mathrm{dense}$} & -0.21(0.52) & 0.51(0.15) & 0.31(0.59) & -0.15(0.38) & 0.02(0.23) \\
{C-NAF$_\mathrm{both}$} & -0.17(0.41) & -0.20(0.53) & 0.05(0.66) & -0.02(0.35) & -0.06(0.22) \\
\midrule
{IAF} & 0.08(0.59) & -0.09(0.09) & \bfseries -0.35(0.61) & -0.22(0.38) & -0.16(0.24) \\
{IA-LR-NSF} & 0.12(0.53) & 0.50 & -0.00(0.63) & 0.08(0.33) & 0.10(0.23) \\
{IA-RQ-NSF} & \bfseries -0.37(0.42) & 1.21(0.11) & 0.07(0.36) & 0.34(0.49) & 0.20(0.24) \\
{IA-NAF$_\mathrm{deep}$} & 0.21(0.28) & -0.36(0.19) & 0.15(0.08) & 0.29(0.40) & 0.17(0.19) \\
{IA-NAF$_\mathrm{dense}$} & 0.08(0.39) & \bfseries -1.03(0.12) & -0.29(0.39) & 0.08(0.45) & -0.15(0.22) \\
{IA-NAF$_\mathrm{both}$} & \bfseries -0.37(0.43) & 0.77(0.22) & -0.28(0.43) & -0.14(0.32) & -0.12(0.20) \\
\midrule
{i-ResNet} & 0.33(0.29) & -0.16(0.34) & \bfseries -0.41(0.34) & \bfseries -0.72(0.41) & -0.31(0.21) \\
{ResFlow} & -0.04(0.14) & -0.30(1.03) & \bfseries -0.94(0.48) & -0.43(0.31) & \bfseries -0.44(0.20) \\
\midrule
{Planar} & 0.37(0.43) & 1.31(0.15) & 0.36(0.30) & -0.06(0.44) & 0.30(0.23) \\
{Radial} & 1.65 & \bfseries -1.01(0.64) & 0.03(0.86) & 0.33(0.53) & 0.46(0.35) \\
{Sylvester} & 1.45(0.04) & 0.29(0.62) & 0.46(0.43) & 0.51(0.10) & 0.77(0.21) \\
\midrule
{CNF$_\mathrm{Euler}$} & \bfseries -1.45(0.04) & \bfseries -1.15(0.49) & -0.29(0.69) & \bfseries -0.65(0.40) & \bfseries -0.80(0.25) \\
{CNF$_\mathrm{RK}$} & \bfseries -0.83(0.77) & \bfseries -1.13(0.14) & \bfseries -0.48(0.75) & \bfseries -0.74(0.32) & \bfseries -0.74(0.26) \\
{CNF$_\mathrm{RK(R)}$} & \bfseries -1.24(0.31) & \bfseries -1.14(0.32) & 0.12(0.67) & \bfseries -0.54(0.24) & \bfseries -0.62(0.22) \\
\bottomrule
\end{tabular}
\caption{$\overline{r} \pm \hat{\sigma}$ for all NFs and target families in NeuTra MH and NeuTra HMC given 2 minutes of warm-up time and 5 minutes of sampling time. NFs in the top 20th percentile are shown in bold for each target family. We estimate $\overline{r} \pm \hat{\sigma}$ with KSD from runs with default hyperparameters. Entries without $\hat{\sigma}$ always attain the same $\overline{r}$. Ranks computed separately for each target family.}
\label{tab:nf-per-family-neutra-stb-ksd}
\end{table}

\subsection{Jump MCMC with the i-SIR global kernel}\label{app:subsec:isir}
\cite{samsonov_local-global_2024} propose using an \textit{iterated sampling importance resampling} (i-SIR) kernel to perform global transitions in a Jump MCMC scheme.
The i-SIR kernel receives as input a state $x_t$ and draws in parallel $m-1$ independent candidate states $x_{t+1, i}^{\prime} \sim Q$ from an NF $Q$, where $i = 2, \dots m$. The current state is also set as a candidate with $x_{t+1, 1}^\prime = x_t$. The kernel associates a weight to each candidate as $w_i = w(x_{t+1,i}^{\prime}) / \sum_{j=1}^m w(x_{t+1,j}^{\prime})$.
The next state is then chosen as $x_{t+1} = x_{t+1, k}^\prime$, where $k \sim \mathrm{Categorical}(w_1, \dots, w_m)$.
i-SIR gives rise to a Markov chain with a kernel that is reversible with respect to the target distribution, Harris recurrent, and ergodic.
It represents an alternative to independent NF proposals, which can be thought of as IMH transitions.

\cite{grenioux_sampling_2023} compared i-SIR and IMH as global kernels within Jump MCMC and found that i-SIR exhibits better acceptance rates than IMH.\footnote{As i-SIR does not perform a classic accept/reject step, an operational definition of acceptance used by~\citep{grenioux_sampling_2023} is when the sampled categorical index corresponds to a newly drawn candidate instead of the current state.}
This is reasonable, as i-SIR can effectively choose between multiple candidates instead of just one, as in IMH.
The practical success of i-SIR as a global jump kernel appears to be linked to the computational efficiency of NF operations and the target density evaluation speed.
This is firstly because each i-SIR transition involves sampling $m-1$ candidate states from the NF.
Second, using the logit weight function $w(x) = \mathrm{softmax}(u(x)), u(x) = \log p_X(x) - \log q(x)$ as per~\citep{samsonov_local-global_2024, grenioux_sampling_2023} involves $m$ target density computations.
We evaluate global i-SIR proposals in Jump MCMC using different NF architectures.

\begin{figure}[ht]
    \begin{subfigure}[t]{0.555\textwidth}
        \centering
        \includegraphics[width=\textwidth, trim={0 0 230 0}, clip]{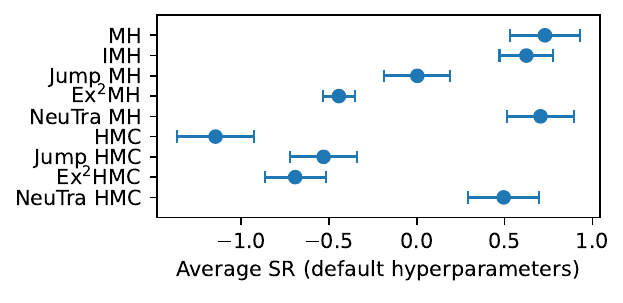}
        \caption{$\overline{r} \pm \hat{\sigma}$ across all targets and NFs for each sampler, using $b^2$ limited to experiments with default NF hyperparameters.}
        \label{subfig:stb-nfmc-all-benchmarks-left-isir}
    \end{subfigure}
    \hfill
    \begin{subfigure}[t]{0.428\textwidth}
        \centering
        \includegraphics[width=\textwidth, trim={300 0 0 0}, clip]{nfmc-all-benchmarks-ex2mcmc.pdf}
        \caption{$\overline{r} \pm \hat{\sigma}$ across all targets and NFs for each sampler, values estimated with minimum $b^2$ across all NF hyperparameter sets.}
        \label{subfig:stb-nfmc-all-benchmarks-right-isir}
    \end{subfigure}
    \caption{Numerical comparison i-SIR (corresponding to Ex$^2$MCMC samplers) to other sampling methods on the entire benchmark according to $b^2$. Each experiment consisted of two minutes of warm-up and five minutes of sampling.}
\end{figure}

We repeated the short-run experiments in Section~\ref{app:subsec:small-time-budget} by replacing IMH with i-SIR as the global jump kernel in Jump MH and Jump HMC.
We used $m = 20$ candidates in our experiments.
Following~\cite{samsonov_local-global_2024}, we refer to Jump MCMC with the i-SIR kernel as Ex$^2$MCMC, short for an explore-exploit MCMC sampling strategy.
We show the results in Figures~\ref{subfig:stb-nfmc-all-benchmarks-left-isir} and~\ref{subfig:stb-nfmc-all-benchmarks-right-isir}.

When using the default set of NF hyperparameters, we find i-SIR to perform better than IMH.
Both Ex$^2$HMC and Jump HMC perform roughly the same, considering the estimated uncertainty.
However, Ex$^2$MH ranks decisively better than all other gradient-free methods, including Jump MH.
Furthermore, using the tuned set of NF hyperparameters allows Ex$^2$MH to even match the performance of HMC.
This suggests that on target distributions comparable to our benchmark, sampling multiple global candidates in each iteration is preferable to a single candidate, despite the added cost of evaluating $p_X$ on each candidate.
In relation to our primary hypothesis, this further strengthens the case that adding global NF jumps to MCMC improves or matches the performance of classic MCMC.

\subsection{Effects of stochastic Jacobian determinant estimation on MCMC bias}\label{app:subsec:jac}
NeuTra MCMC, Jump MCMC, and IMH all rely on the Jacobian determinant of the NF transformation $f$.
In residual and continuous NFs, the determinant is estimated stochastically via roulette, power series, and Hutchinson trace estimators.
In NeuTra MCMC, this implies that the adjusted log density $\log \widetilde{p}(x)$ includes a log Jacobian determinant term with some degree of randomness.
In Jump MCMC and IMH, this similarly implies that the acceptance rate of independent NF jumps contains randomness in the log NF density $\log q(x)$ via the log Jacobian determinant of the transformation $f$.

For IMH transitions within Jump MCMC and standalone IMH, we relate the phenomenon to noisy Metropolis-Hastings~\citep{alquier_noisy_2016}.
Given an IMH transition kernel $P$ and an approximate IMH transition kernel $\hat{P}$, we can bound the distance between the corresponding Markov chains.
Specifically, Corollary 2.3 in~\citep{alquier_noisy_2016} states that if $P$ is a kernel corresponding to a uniformly ergodic Markov chain with acceptance probability $\alpha(x, x^\prime)$ and $\hat{P}$ is a kernel with stochastic acceptance probability $\hat{\alpha}(x, x^\prime, y^\prime)$ for noise $y^\prime$ drawn from a distribution $F_{x^\prime}$ such that
\begin{align}
    \mathbb{E}_{y^\prime \sim F_{x^\prime}} \left[ \left|  \alpha(x, x^\prime) - \hat{\alpha}(x, x^\prime, y^\prime) \right| \right] \leq \Delta(x, x^\prime),
\end{align}
then for some $C<\infty, 0 \leq \rho < 1$, the total variation (TV) distance between the two chains is bounded as:
\begin{align}
    ||\delta_{x_0} P^{n} - \delta_{x_0} \hat{P}^{n}||_{\mathrm{TV}} \leq \left(\lambda+\frac{C \rho^\lambda}{1 - \rho}\right) \sup_{x} \int d x^\prime h(x^\prime|x)\Delta(x,x^\prime), \label{eqn:app:tv-dist-1}
\end{align}
for any $n\in \mathbb{N}$ and any starting point $x_0$, where $\delta$ is the Dirac delta measure, $\Delta$ is a pointwise bound on expected acceptance error, $\lambda = \left\lceil \frac{\log(1/C)}{\log(\rho)} \right\rceil$, and $h$ is the Metropolis-Hastings transition conditioned on the current state $x$.
The values $C, \rho$ specify uniform ergodicity of the transition kernel $P$ with respect to the target $p_X$:
\begin{align}
    \sup_{x_0} ||\delta_{x_0}P^n - p_X||_{\mathrm{TV}} \leq C \rho^n. \label{eqn:app:uniform-ergodicity}
\end{align}
When we select $\hat{\alpha}$ such that $\Delta \ll 1$, we obtain $|| \delta_{x_0}P^n - \delta_{x_0}\hat{P}^n ||_{\mathrm{TV}} \ll 1$, yielding:
\begin{align}
    \limsup_{n\rightarrow\infty} ||\delta_{x_0} \hat{P}^n - p_X||_{\mathrm{TV}} \leq \Delta \left(\lambda + \frac{C\rho^\lambda}{1-\rho} \right).\label{eqn:app:limsup-tv}
\end{align}

Corollary 3.1 in~\citep{mitrophanov_sensitivity_2005} provides a general result that can be related not only to Jump MCMC and IMH, but also to NeuTra MCMC, as it does not explicitly assume an approximate accept/reject step.
Given a uniformly ergodic Markov chain with a transition kernel $P$ (i.e., satisfying Equation~\ref{eqn:app:uniform-ergodicity}) and an approximate kernel $\hat{P}$, we have for any $n \in \mathbb{N}$ and any starting point $x_0$:
\begin{align}
        ||\delta_{x_0} P^{n} - \delta_{x_0} \hat{P}^{n}||_{\mathrm{TV}} \leq \left(\lambda+\frac{C \rho^\lambda}{1 - \rho}\right) || P - \hat{P} ||_{\mathrm{TV}}. \label{eqn:app:tv-dist-2}
\end{align}
Following this, \cite{mitrophanov_sensitivity_2005} also provides an upper bound on the TV distance between the stationary distributions of $P$ and $\hat{P}$.

Equations~\ref{eqn:app:limsup-tv} and~\ref{eqn:app:tv-dist-2} provide an important justification for MCMC methods where the acceptance rate may be stochastic, including NFMC methods where $\hat{\alpha}$ is influenced by Jacobian determinant estimators.
Equation~\ref{eqn:app:tv-dist-2} can also be connected to CNF bijections and residual NF inverses, whose data transformations rely on approximate numerical integration and the Banach fixed point theorem, respectively.
We can relate~\ref{eqn:app:limsup-tv} to IMH by simplifying the transition kernel $h$ to be independent of the current state.
We again note that the non-truncated power series estimator in~\citep{behrmann_invertible_2019}, the roulette estimator in~\citep{chen_residual_2019}, and the Hutchinson trace estimator in e.g.,~\citep{grathwohl_ffjord_2018} are all unbiased, which may contribute to lowering the TV distance in the above equations.
Moreover, in practical implementations of i-ResNet, \cite{behrmann_invertible_2019} find that truncating the power series estimator exhibits a bias of less than 0.001 bits per dimension after only 5-10 series terms.
\cite{grathwohl_ffjord_2018} discuss the error incurred by numerical integration.
They find that decreasing ODE solver tolerance reduces the error in the integral over the entire probability density.
Specifically, using a tolerance of $10^{-7}$ yields an integration error of approximately $10^{-7}$ on a multimodal unidimensional example.
The decreasing tolerance linearly decreases integration error on a log-log plot, i.e., dividing tolerance by 10 approximately divides integration error by a positive constant.
However, using the less precise Euler solver can substantially increase determinant bias.

\begin{figure}[h]
    \centering
    \includegraphics[width=\linewidth]{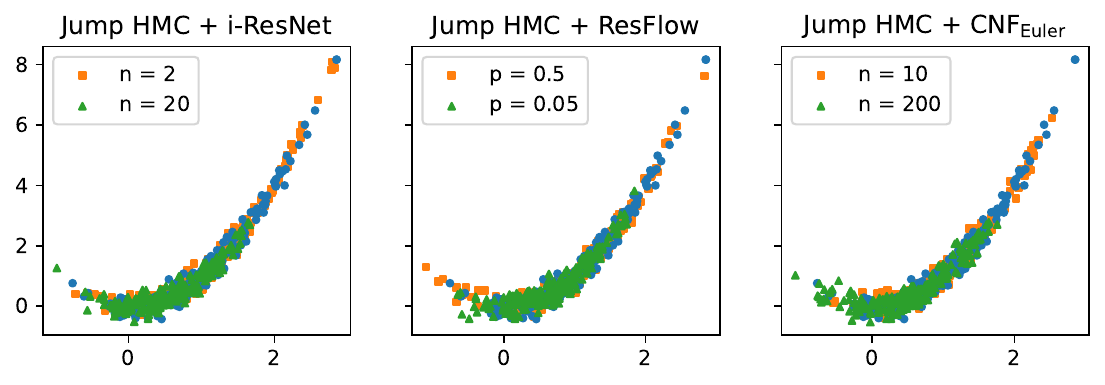}
    \caption{Comparison of Jacobian estimators and the Euler integrator with respect to the number of power series iterations for i-ResNet, Geometric probability for ResFlow, and the number of integration steps for \cnfeuler. Blue circles represent training samples from HMC, orange squares represent samples from low-accuracy approximation/integration, and green triangles represent samples from high-accuracy approximation/integration. Scatterplots are limited to the first two dimensions of the Rosenbrock target and 200 randomly chosen samples out of $10^4$ total samples.}
    \label{fig:app:jac-est}
\end{figure}

We empirically investigated the impact of approximate kernels and accept/reject steps in i-ResNet, ResFlow, and \cnfeuler.
We considered the 100-dimensional Rosenbrock banana target distribution and observed how the 2D scatterplot of NFMC draws changes as we vary the bijection accuracy of these NFs.
Specifically, we considered a low-accuracy and a high-accuracy setting:
\begin{itemize}
    \item We used $n=2$ and $n=20$ power series iterations in i-ResNet.
    \item We used $p=0.5$ and $p=0.05$ as the Geometric probability in ResFlow.
    \item We used $n=10$ and $n=200$ Euler steps in \cnfeuler.
\end{itemize}
We ran HMC with 100 chains for 1000 warmup iterations and 1000 sampling iterations, then Jump HMC for 100 iterations, i.e., 100 jumps and 100 HMC iterations per jump.
Other settings were the same as in our main experiments.
We show scatterplots of the first and second dimensions in Figure~\ref{fig:app:jac-est}.

We find that lower-accuracy bijections capture the right tail of the target better than higher-accuracy ones.
Focusing on jumps alone, we observed an acceptance rate of 0.48 in \cnfeuler with $n=10$ and 0.62 in \cnfeuler with $n=200$.
Both these values are comparable and relatively high, so jumps do have an impact on sampler performance.
Since Equation~\ref{eqn:app:tv-dist-2} states that the chain distance is bounded as a function of kernel distance, it suggests that the high-accuracy bijections give rise to chains that are closer to the non-approximate reference chain than the low-accuracy bijections.
Despite this, and the possible bias incurred by low-accuracy Jacobian approximations and integration, the samples relating to such low-accuracy kernels appear to better capture tail behavior.
Future works on NFMC could build on the results by~\cite{alquier_noisy_2016} and~\cite{mitrophanov_sensitivity_2005} to better understand how approximate NF kernels impact sample quality, either from an empirical or a theoretical perspective.
While this ablation required fixed-length chains for a fair comparison, we note that all high-accuracy bijections yielded a significantly longer NF training time and a moderately longer NFMC sampling time.
This may further impact the practical performance of approximate NF kernels and is an important consideration for future works.

\section{Experiment details}
We provide experiment details, including used hardware, sampler warm-up procedures, NF training details, and NF hyperparameter choices.
\subsection{Hardware configuration}\label{app:sec:experiment-configuration}
Unless otherwise noted, we ran all experiments with the AMD EPYC 7702P CPU.
To estimate ground truth moments, we ran standard HMC (without NF extensions) with 100 parallel chains for 20 hours.
For each experiment, we ran warm-up for 3 hours and sampling for 7 hours, using 8 GB of memory.
The total sequential computation time for the experiments in this paper was roughly 5 to 6 years (not accounting for repeated runs).

\subsection{MCMC and NFMC warm-up}
We warmed up MH and HMC by sampling while adapting their parameters. We adapted HMC step size with dual averaging, mass matrices in each sampler by $M_{t+1}^{-1} = M_t^{-1} + \sqrt{\mathrm{Cov}[x_t]} \cdot 0.999^t$,
where $x_t$ are the current chain states.
\neutramcmc first performs stochastic variational inference for as long as possible (at most 3 hours), then we warm up the inner MCMC sampler on the adjusted log density and obtain MCMC samples.
Jump MCMC has the same warm-up procedure as \neutramcmc, except that we also fit the NF again to samples from the MCMC fit.

\subsection{NF training details}\label{app:subsec:nf-training-details}
In maximum likelihood fitting (\ie given training samples), we trained all NFs with the Adam optimizer, step size 0.05 and batch size 1024.
When given a validation set (in maximum likelihood fitting), we stopped training after no validation loss improvement in 5000 consecutive steps.
In SVI, we stopped training after 5000 steps of no training loss improvement.
We kept the best weights according to validation loss in both cases.
We used a single sample in SVI.

At the time of writing, \cite{agrawal_disentangling_2024} performed a study of SVI for the \realnvp architecture.
They observe that large batch sizes reduce gradient variance and thus improve fit quality.
They observe a similar effect when using a reduced variance gradient estimator, however they state that it is impractical for NFs with expensive bijection inversion costs.
Our preliminary tests showed that large batch sizes and the reduced variance estimator take up a large chunk of the computational budget for certain NFs due to very slow autodifferentiation, thus negatively impacting their performance.
Applying large batch sizes and the estimator to only select NFs would prevent a fair comparison and potentially add excessive variation to our results.
We thus avoid these two approaches in SVI fits.
Our choice promotes a fair comparison because the fitting routines are identical for all NFs.

\subsection{NF hyperparameter choices}\label{app:sec:nf-hyperparameters}
We set NF hyperparameters such that all NFs successfully passed a series of automated tests, which ensured numerical stability in the following aspects:
\begin{itemize}
    \item We reconstruct an input by first passing it to the forward bijection method, then the inverse bijection method. The reconstruction error must not be too great.
    \item No trainable parameter or the loss may take on a NaN value during forward passes, inverse passes, and loss gradient computation.
\end{itemize}
For autoregressive NFs, we use the following hyperparameter combinations:
\begin{itemize}
    \item We used 2, 5, or 10 bijective layers in the composition.
    \item We used either conditioner hidden size 10 and two conditioner layers or conditioner hidden size 100 and five conditioner layers.
\end{itemize}
We split input tensors in half across the first dimension for coupling NFs.
We used the Tanh activation in all conditioners, as it ensured controllable magnitudes of outputs.
We noticed that the ReLU activation can result in predicting parameters in large magnitudes, which causes divergences in NFMC.
We used 8 splines in all LRS and RQS transformers.
For NAF, we used one dense layer with 8 neurons for \naftransformerdense, two hidden layers with hidden size $\maxtwo{5 \ceil{\logten d}}{4}$ in \naftransformerdeep, and two layers with 8 neurons in \naftransformerboth.
For Sylvester flows, we used $m = \frac{d}{2}$ columns in $Q$ for the $QR$ decomposition.
We used 2, 5, and 10 layers in matrix determinant residual NFs.
For contractive residual NFs, we used a spectrally normalized neural network with 1 hidden layer, $3 \maxtwo{\ceil{\logten d}}{4}$ hidden neurons, and TanH activations.
We used $p = 0.5$ for the Roulette estimator in ResFlow.
To parameterize $g_\phi$ in continuous NFs, we used 1, 5, or 10 hidden layers with 10 or 100 hidden neurons.
We implement time-dependence in $g_\phi$ for \cnfrkreg and \cnfrk by concatenating the time variable to the remainder of the input.
Any other hyperparameter choices can be found in the linked repositories.

\section{Sampler definitions}\label{app:sampler-definitions}
In this section, we define each investigated MCMC and NFMC sampler.

\subsection{Metropolis-Hastings}\label{app:subsec:mh}
The (random-walk) Metropolis-Hastings sampler defines its proposal and log acceptance ratio as follows:
\begin{align*}
    x^\prime_{t+1} &= x_t + M^{-1} u_t, \; u_t \sim N(0, I), \\
    \log \alpha_t &= \log p_X(x^\prime_t) -\log p_X(x_t).
\end{align*}
Here, $x_t$ is the current state, $x^\prime_{t+1}$ is the proposed next state, $M^{-1}$ is the inverse of the diagonal mass matrix.
We set $x_{t+1} = x^\prime_{t+1}$ if $\log \alpha_t > \log w_t, w_t \sim U(0, 1)$, otherwise we set $x_{t+1} = x_t$.

\subsection{Hamiltonian Monte Carlo}
Hamiltonian Monte Carlo (with the leapfrog integrator) defines each trajectory step as:
\begin{align*}
    r^{(k + 1/2)} &= r^{(k)} + h / 2 \nabla \log p_X(x^{(t)}), \\
    x^{(k + 1)} &= x^{(k)} + h M^{-1} r^{(k + 1/2)}, \\
    r^{(k + 1)} &= r^{(k + 1/2)} + h / 2 \nabla \log p_X(x^{(k+1)}).
\end{align*}
Here, $x^{(k)}, r^{(k)}$ are the current trajectory state and momentum, $x^{(k+1)}, r^{(k+1)}$ are the next trajectory state and momentum, $r^{(k+1/2)}$ is the intermediate momentum, $M^{-1}$ is the inverse of the diagonal mass matrix.
The trajectory has $L > 0$ steps with $r^{(1)}= M^{-1} u, u \sim N(0, I)$.
The proposed next state and log acceptance ratio are defined as:
\begin{align*}
    x^\prime_{t+1} &= x_t^{(L)}, \\
    \log \alpha_t &= \log p_X(x^\prime_{t+1}) - \log p_X(x_t) - 0.5 \left(r^{(L)\top}_t M^{-1} r^{(L)}_t - r^{(1)\top}_t M^{-1} r^{(1)}_t\right).
\end{align*}
The momentum $r^{(1)}$ is refreshed at the beginning of each trajectory by newly sampling $u \sim N(0, I)$.
We set the new state as in subsection~\ref{app:subsec:mh}.

\subsection{NeuTra MCMC}
\neutramcmc uses an NF $Q$ with the bijection whose inverse map $f^{-1}$ transforms a latent space point $z$ to a target space point $x = f^{-1}(z)$.
Given a target distribution $p_X(x)$, it runs an MCMC sampler with the transformed target log density:
\begin{align*}
    \log \widetilde{p}(z) = \log p_X(f^{-1}(z)) + \logabsdetjac{f^{-1}(z)}{z}.
\end{align*}
We set the new state as in subsection~\ref{app:subsec:mh}.
This generates a chain of $n$ states.
Afterwards, all $n$ latent points $z_i$ are transformed to target space samples via $x_i = f^{-1}(z_i)$ for $i = 1, \dots, n$.

\subsection{Jump MCMC and IMH}
Every $K$-th step of Jump MCMC with an NF $Q$ proposes a new state as in Equation~\ref{eqn:jump-mcmc-x-prime} and computes the log acceptance ratio as in Equation~\ref{eqn:jump-mcmc-log-alpha}.
We set the new state as in subsection~\ref{app:subsec:mh}.
Other steps are performed with an MCMC sampler.
When $K=1$, Jump MCMC reduces to IMH.

\section{Benchmark distribution details}
In this section, we give precise definitions for all target distributions in our benchmark.
We parameterize univariate Gaussian and half-Cauchy distributions with the standard deviation (not the variance).

\subsection{Synthetic Gaussian target distributions}\label{app:gaussian-targets}
We use the following Gaussian distributions in our benchmark:
\begin{itemize}
    \item Standard Gaussian in 100 dimensions.
    \item Diagonal Gaussian in 100 dimensions with zero mean and standard deviation linearly spaced between 1 and 10.
    \item Full-rank Gaussian in 100 dimensions with zero mean and eigenvalues $\lambda_1, \dots, \lambda_{100}$ linearly spaced between 1 and 10, giving rise to covariance $\Sigma = Q\Lambda Q^\top$ with $\Lambda = \diag\left(\lambda_1, \dots, \lambda_{100}\right)$ and $Q$ orthonormal.
    \item Ill-conditioned full-rank Gaussian in 100 dimensions with zero mean and eigenvalue reciprocals $\lambda_i^{-1} \sim \mathrm{Gamma}(0.5, 1)$, giving rise to covariance $\Sigma = Q\Lambda Q^\top$ with $\Lambda = \diag\left(\lambda_1, \dots, \lambda_{100}\right)$ and $Q$ orthonormal.
\end{itemize}
The orthonormal rotation matrix $Q$ is generated by decomposing a $100 \times 100$ standard normal matrix $A$ into $Q_0 R = A$ where $Q_0$ is orthonormal and $R$ is upper triangular.
We then proceed with $Q = Q_0 \diag (\mathrm{sign} (\diag (R)))$, which multiplies the diagonal of $Q_0$ with the sign of the diagonal of $R$ to give $Q$ a determinant of 1, while keeping all off-diagonal elements of $Q$ the same as $Q_0$.

\subsection{Synthetic non-Gaussian unimodal target distributions}\label{app:synthetic-non-gaussian-targets}
We use the following synthetic non-Gaussian unimodal distributions in our benchmark:
\begin{itemize}
    \item Funnel distribution in 100 dimensions. The first dimension is given by $N(0, 3)$, all remaining dimensions are given by $x_i | x_1 \sim N(0, \exp(x_1 / 2))$.
    \item Rosenbrock banana distribution in 100 dimensions with scale 10.
\end{itemize}
The Rosebrock banana log density for an input $x \in \mathbb{R}^D$ (with $D$ even and scale $s$) is defined as:
\begin{align*}
    \log p_X(x) = -\sum_{d=1}^{D/2} s (x_{2d-1}^2 - x_{2d})^2 + (x_{2d-1} - 1)^2 - C,
\end{align*}
where $C$ is the log of the normalization constant.

\subsection{Synthetic multimodal target distributions}\label{app:multimodal-targets}
We use the following multimodal distributions in our benchmark:
\begin{itemize}
    \item A mixture with three diagonal Gaussian components in 100 dimensions. Component means are $-5, 0,$, and $5$, respectively, in all dimensions. Component standard deviations are $0.7$ in all dimensions. Component weights are $1/3$ for all three components.
    \item A mixture with twenty diagonal Gaussian components in 100 dimensions. Component means are randomly sampled from $N(0, 10)$ in all dimensions. Component standard deviations are $1$ in all dimensions. Component weights given by $\mathrm{softmax}(x_1, \dots, x_{20})$ where $x_i \sim N(0, 1)$.
    \item A double well distribution in 10 dimensions (containing $2^{10}$ modes).
    \item A double well distribution in 100 dimensions (containing $2^{100}$ modes).
\end{itemize}
The double well log density for an input $x \in \mathbb{R}^D$ is defined as
\begin{align*}
    \log p_X(x) = -\sum_{d=1}^D (x^2 - 4)^2 - C,
\end{align*}
where $C$ is the log of the normalization constant.

\subsection{Real-world target distributions}\label{app:real-world-targets}

We define the real-world target distributions included in our benchmark. We acquire data for likelihood functions from the repository by~\cite{magnusson_posteriordb_2024}.

\subsubsection{Eight schools}
Given a parameter vector $(\mu, \widetilde{\tau}, \theta^\prime)$ with $\mu \in \mathbb{R}$, $\widetilde{\tau} \in \mathbb{R}$, $\theta^\prime \in \mathbb{R}^8$ and measurements $y_i \in \mathbb{R}$, $\sigma_i > 0, i = 1, \dots, 8$, the 10D eight schools model defined as:
\begin{align*}
    \tau &= \log (1 + \exp (\widetilde{\tau})), \; \theta = \mu + \tau \theta^\prime, \\
    \mu &\sim N(0, 10), \; \tau \sim \mathrm{LogNormal}(5, 1), \; \theta^\prime_i \sim_{iid} N(0, 1), \\
    y_i &\sim N(\theta_i, \sigma_i).
\end{align*}

\subsubsection{German credit}
Given a parameter vector $(\widetilde{\tau}, \beta)$ with $\widetilde{\tau} \in \mathbb{R}$, $\beta \in \mathbb{R}^{25}$ and measurements $(x_j, y_j)$ with $x_j \in \mathbb{R}^{25}$, $y_j \in \{0, 1\}$, the 26D German credit model is defined as:
\begin{align*}
    \tau &= \log (1 + \exp (\widetilde{\tau})), \\
    \tau &\sim \mathrm{Gamma}(0.5, 0.5), \; \beta_i \sim_{iid} N(0, 1), \\
    y_j &\sim \mathrm{Bernoulli}(\sigma(\tau \beta^\top x_j)).
\end{align*}
We use the shape-rate parameterization for the Gamma distribution. Bernoulli parameters are computed with the sigmoid function $\sigma$.

\subsubsection{Sparse German credit}
Given a parameter vector $(\widetilde{\tau}, \widetilde{\lambda}, \beta)$ with $\widetilde{\tau} \in \mathbb{R}, \widetilde{\lambda} \in \mathbb{R}^{25}, \beta \in \mathbb{R}^{25}$ and measurements $(x_j, y_j)$ with $x_j \in \mathbb{R}^{25}$, $y_j \in \{0, 1\}$, the 51D sparse German credit model is defined as:
\begin{align*}
    \tau &= \log (1 + \exp (\widetilde{\tau})),\; \lambda = \log (1 + \exp (\widetilde{\lambda})), \\
    \tau, \lambda_i &\sim_{iid} \mathrm{Gamma}(0.5, 0.5), \; \beta_i \sim_{iid} N(0, 1), \\
    y_j &\sim \mathrm{Bernoulli}(\sigma(\tau (\beta \lambda)^\top x_j)).
\end{align*}
We use the shape-rate parameterization for the Gamma distribution.
The product between $\beta$ and $\lambda$ is element-wise multiplication. Bernoulli parameters are computed with the sigmoid function $\sigma$.

\subsubsection{Radon (varying intercepts)}
Given a parameter vector $(\mu_b, \widetilde{\sigma}_b, \widetilde{\sigma}_y, a, b)$ with $\mu_b, \widetilde{\sigma}_b, \widetilde{\sigma}_y, a \in \mathbb{R}, b \in \mathbb{R}^{85}$ and measurements $(r_j, f_j)$ with $r_j \in \mathbb{R}$, $f_j \in \{0, 1\}$, the 89D radon model with varying intercepts is defined as:
\begin{align*}
    \sigma_b &= \log (1 + \exp (\widetilde{\sigma}_b)),\; \sigma_y = \log (1 + \exp (\widetilde{\sigma}_y)) \\
    \mu_b, a &\sim N(0, 10^5), \; \sigma_b, \sigma_y \sim \mathrm{HalfCauchy}(5), \; b_i \sim_{iid} N(\mu_b, \sigma_b), \\
    r_j &\sim N(a f_{c(j)} + b_{c(j)}, \sigma_y).
\end{align*}
Here, $c(j)$ is the county associated with the data point at index $j$.

\subsubsection{Radon (varying slopes)}
Given a parameter vector $(\mu_a, \widetilde{\sigma}_a, \widetilde{\sigma}_y, a, b)$ with $\mu_a, \widetilde{\sigma}_a, \widetilde{\sigma}_y, b \in \mathbb{R}, a \in \mathbb{R}^{85}$ and measurements $(r_j, f_j)$ with $r_j \in \mathbb{R}$, $f_j \in \{0, 1\}$, the 89D radon model with varying intercepts is defined as:
\begin{align*}
    \sigma_a &= \log (1 + \exp (\widetilde{\sigma}_a)),\; \sigma_y = \log (1 + \exp (\widetilde{\sigma}_y)) \\
    \mu_a, b &\sim N(0, 10^5), \; \sigma_a, \sigma_y \sim \mathrm{HalfCauchy}(5), \; a_i \sim_{iid} N(\mu_a, \sigma_a), \\
    r_j &\sim N(a_{c(j)} f_{c(j)} + b, \sigma_y).
\end{align*}
Here, $c(j)$ is the county associated with the data point at index $j$.

\subsubsection{Radon (varying intercepts and slopes)}
Given a parameter vector $(\mu_a, \widetilde{\sigma}_a, \widetilde{\sigma}_y, a, b)$ with $\mu_a, \widetilde{\sigma}_a, \widetilde{\sigma}_y, a, b \in \mathbb{R}^{85}$ and measurements $(r_j, f_j)$ with $r_j \in \mathbb{R}$, $f_j \in \{0, 1\}$, the 175D radon model with varying intercepts and slopes is defined as:
\begin{align*}
    \sigma_a &= \log (1 + \exp (\widetilde{\sigma}_a)), \; \sigma_b = \log (1 + \exp (\widetilde{\sigma}_b)),\; \sigma_y = \log (1 + \exp (\widetilde{\sigma}_y)) \\
    \mu_a, \mu_b &\sim N(0, 10^5), \; \sigma_a, \sigma_b, \sigma_y \sim \mathrm{HalfCauchy}(5), \; a_i \sim_{iid} N(\mu_a, \sigma_a), b_i \sim_{iid} N(\mu_b, \sigma_b), \\
    r_j &\sim N(a_{c(j)} f_{c(j)} + b_{c(j)}, \sigma_y).
\end{align*}
Here, $c(j)$ is the county associated with the data point at index $j$.

\subsubsection{Synthetic item response theory}
Given a parameter vector $(\alpha, \beta, \delta)$ with $\alpha \in \mathbb{R}^{400}, \beta \in \mathbb{R}^{100}, \delta \in \mathbb{R}$ and measurements $y_k \in \{0, 1\}$, the 501D synthetic item response theory model is defined as:
\begin{align*}
    \delta &\sim N(0.75, 1), \; \alpha_i, \beta_j \sim_{iid} N(0, 1) \\
    y_k &\sim \mathrm{Bernoulli}(\sigma(\alpha_{s(k)} - \beta_{r(k)} + \delta))
\end{align*}
Here, $s(k)$ is the student, and $r(k)$ is the response associated with the data point at index $k$. Bernoulli parameters are computed with the sigmoid function $\sigma$.

\subsubsection{Stochastic volatility}
Given a parameter vector $(z, \widetilde{\sigma}, \widetilde{\mu}, \widetilde{\phi}^\prime)$ with $z \in \mathbb{R}^{3000}$; $\widetilde{\sigma}, \widetilde{\mu}, \widetilde{\phi}^\prime \in \mathbb{R}$ and measurements $y_i \in \mathbb{R}$, the 3003D stochastic volatility model is defined as:
\begin{align*}
    \sigma &= \log (1 + \exp(\widetilde{\sigma})), \; \mu = \log (1 + \exp(\widetilde{\mu})),\; \phi^\prime = 1 / (1 + \exp(-\widetilde{\phi}^\prime)),\; \phi = 2 \phi^\prime - 1, \\
    h_1 &= \mu + \sigma z_1 / \sqrt{1 - \phi^2}, \; h_i = \mu + \sigma z_i + \phi (h_{i-1} - \mu) \textrm{ for } i > 1, \\
    z_i &\sim_{iid} N(0, 1),\; \sigma \sim \mathrm{HalfCauchy}(2),\; \mu \sim \mathrm{Exp}(1), \; \phi^\prime \sim \mathrm{Beta}(20, 1.5), \\
    y_i &\sim N(0, \exp(h_i / 2)).
\end{align*}

\section{Comparison metric definitions}\label{app:sec:comparison-metrics}
In this section, we define the squared bias of the second moment and the standardized rank.
\subsection{Squared bias of the second moment}
Let $X$ be the random variable corresponding to a target distribution, and $X_d $ its $d$-th dimension.
Let $\mathbb{E}[X_d^2]$ be the true $d$-th marginal second moment and $\mathrm{Var}[X_d]$ the true $d$-th marginal variance of the target distribution.
Let $\widetilde{\mathbb{E}}[X_d^2]$ be the estimated second moment of $X_d$, obtained using MCMC samples $x^{(i,j)}$ from $n$ MCMC steps and $m$ independent chains:
\begin{align*}
    \widetilde{\mathbb{E}}[X_d^2] = \frac{1}{nm} \sum_{i=1}^n \sum_{j=1}^m \left(x_d^{(i,j)}\right)^2.
\end{align*}
We measure the error in estimating a distribution's moments with the squared bias of the second moment:
\begin{align*}
    b^2 = \max_{d} \frac{\left(\widetilde{\mathbb{E}}[X_d^2] - \mathbb{E}[X_d^2]\right)^2}{\mathrm{Var}[X_d]}.
\end{align*}
The minimum possible value of $b^2$ is zero when the estimated second moment exactly matches the true second moment.
In practice, we observe non-negative values of $b^2$.
If one sampler attains lower $b^2$ than another on a target distribution, we deem it better for second moment estimation on that distribution.

Ignoring the scaling via the true variance and summarization with the maximum function, $b^2$ relates to the classical bias that arises as a component in the bias-variance decomposition of the mean squared error (MSE):
\begin{align}
    \mathrm{MSE}(f(X), f(\hat{X})) &= \mathbb{E}[(f(X) - f(\hat{X}))^2] \\ 
    &= \underbrace{\mathrm{Var}[f(\hat{X})]}_{\text{Estimator variance}} + \underbrace{(\mathrm{E}[f(X)] - \mathrm{E}[f(\hat{X})])^2}_\text{Squared bias} + \underbrace{\mathrm{Var}[f(X)],}_\text{Irreducible error}
\end{align}
where $f$ is a statistical functional, $X$ is the target distribution, and $\hat{X}$ is its approximation, often an empirical distribution based on MCMC samples.
The irreducible error term can be ignored within comparisons.
Estimator variance decays as $\bigo{1/n}$ under the Markov Chain central limit theorem~\citep[Theorem 3]{chan_discussion_1994}, while bias may not vanish as quickly and may thus dominate MSE.
For large sample sizes, as in our experiments, we thus expect squared bias to distinguish different methods well.

\subsection{Standardized rank}
Suppose we rank sampling methods $m_1, \dots, m_K$ on a single target according to $b^2$, obtaining ranks $r_1, \dots, r_K$. Sampling methods include MCMC samplers, NFs, or both.
We obtain standardized ranks by subtracting the empirical mean and dividing by the standard deviation:
\begin{align*}
    r_{s,i} = \frac{r_i - \widetilde{\mu}}{\widetilde{\sigma}}, \textrm{ where } \widetilde{\mu} = \frac{1}{K} \sum_{i=1}^K r_i \textrm{ and } \widetilde{\sigma} = \sqrt{\frac{1}{K - 1} \sum_{i=1}^K \left(r_i - \widetilde{\mu}\right)^2}.
\end{align*}

If we compute $r_{s,i}$ for different targets $j = 1, \dots, B$, we can observe their empirical distribution. We can also estimate the mean and the standard error of the mean:
\begin{align*}
    \overline{r_{s,i}} = \frac{1}{B} \sum_{j = 1}^B r_{s,i}^{(j)} \textrm{ and } \hat{\sigma}_{s,i} = \frac{\sigma_{s,i}}{\sqrt{B}}, \textrm{ where } \sigma_{s,i} &= \sqrt{\frac{1}{B - 1} \sum_{j=1}^B \left( r_{s,i}^{(j)} - \overline{r_{s,i}}\right)^2}.
\end{align*}
We construct a confidence interval for $\overline{r_{s,i}}$ as $\left(\overline{r_{s,i}} - \hat{\sigma}_{s,i}, \overline{r_{s,i}} + \hat{\sigma}_{s,i}\right)$.
This interval defines uncertainty in estimating $\overline{r_{s,i}}$. 
The smaller the interval, the more confident we are in our estimate of $\overline{r_{s,i}}$.

\section{Memory-efficient moment estimation}\label{app:sec:eficient-running-moments}
Estimating moments of high-dimensional targets by averaging all acquired samples is computationally inefficient and causes out-of-memory errors on longer runs.
Instead, we implemented a running-average approach for moment estimation.
Suppose we are at step $m$ of an MCMC run.
We have already used samples $x_1, \dots, x_m$ to compute the current running average, and we wish to use samples $x_{m+1}, \dots, x_n$ to update the running average.
We derive the empirical running average $\mathbb{E}_{1:n}[f(x)]$ of a statistical functional $f$ with transformed data point $f_i = f(x_i)$ as follows:
\begin{align*}
    \mathbb{E}_{1:n}[f(x)] &= \frac{1}{n} \sum_{i=1}^n f_i = \frac{1}{n} \sum_{i=1}^m f_i + \frac{1}{n} \sum_{i=m+1}^n f_i \\
     &= \frac{m}{n} \frac{1}{m} \sum_{i=1}^m f_i + \frac{n-m}{n}\frac{1}{n-m} \sum_{i=m+1}^n f_i \\
     &= \frac{m}{n} \mathbb{E}_{1:m}[f(x)] + \frac{n-m}{m}\mathbb{E}_{m+1:n}[f(x)].
\end{align*}
We thus weigh the previous average and the average of the incoming batch of samples.
The space complexity of the estimate is bounded by the size of the sample batch, which is usually just one data point of size equal to the target dimensionality for each chain.
We estimate the first moment with $f(x) = x$ and the second moment with $f(x) = x^2$.

\subsection{Efficient NeuTra moments}
In its original formulation, \neutramcmc samples all points in the latent space and then transforms them back to the original space once sampling has finished.
To avoid excessive memory usage, we instead reuse the described running moment estimation approach and transform data points with the inverse NF transformation.
The functionals thus become $f(x) = \texttt{inverse}(x)$ for the first moment and $f(x) = \texttt{inverse}(x)^2$ for the second moment.
Note that the inverse only has to be called once.
This approach applies the same number of inverse calls as classic \neutramcmc but requires constant memory, whereas space requirements otherwise grow with the number of MCMC steps.

\end{appendices}

\end{document}